\documentclass[sigconf,screen]{acmart}
\usepackage{xspace,balance,tabularx,multirow}
\usepackage{flushend}
\usepackage{tikz}
\usepackage{pgfplots}
\pgfplotsset{compat=1.16}
\usetikzlibrary{patterns}
\usepackage{subfig}
\usepackage[ruled, vlined, linesnumbered]{algorithm2e}
\usepackage{xcolor}
\usepackage{colortbl}
\usepackage{bbold}
\SetKwComment{Comment}{$\triangleright$\ }{}
\usepackage{enumitem}
\usepackage{tablefootnote}
\usepackage{upgreek,textgreek}
\usepackage{pifont}%
\usepackage[noabbrev]{cleveref}
\usepackage{titlecaps}
\usepackage{lipsum}
\usepackage{makecell}

\pgfplotsset{every tick label/.append style={font=\tiny}}

\newlength{\starsize}
\newlength{\starspread}
\tikzset{starsize/.code={\setlength{\starsize}{#1}},
         starspread/.code={\setlength{\starspread}{#1}}}
\tikzset{starsize=1mm,
         starspread=3mm}
\pgfdeclarepatternformonly[\starspread,\starsize]%
  {my fivepointed stars}%
  {\pgfpointorigin}%
  {\pgfqpoint{\starspread}{\starspread}}%
  {\pgfqpoint{\starspread}{\starspread}}%
  {%
   \pgftransformshift{\pgfqpoint{\starsize}{\starsize}}
   \pgfpathmoveto{\pgfqpointpolar{18}{\starsize}}
   \pgfpathlineto{\pgfqpointpolar{162}{\starsize}}
   \pgfpathlineto{\pgfqpointpolar{306}{\starsize}}
   \pgfpathlineto{\pgfqpointpolar{90}{\starsize}}
   \pgfpathlineto{\pgfqpointpolar{234}{\starsize}}
   \pgfpathclose%
   \pgfusepath{fill}
  }

\newcommand{\renchi}[1]{{\color{red}{[Renchi: #1]}}}

\newcommand{\haoran}[1]{{\color{blue}{[Haoran: #1]}}}

\makeatletter
\newcommand*\bigcdot{\mathpalette\bigcdot@{.5}}
\newcommand*\bigcdot@[2]{\mathbin{\vcenter{\hbox{\scalebox{#2}{$\m@th#1\bullet$}}}}}
\makeatother

\newcommand{\stitle}[1]{\vspace*{0.5em}\noindent{\underline{\bf #1.\/}}}

\newcommand{\V}{\mathcal{V}\xspace}
\newcommand{\G}{\mathcal{G}\xspace}
\newcommand{\N}{\mathcal{N}\xspace}
\newcommand{\Y}{\mathcal{Y}\xspace}
\newcommand{\EDG}{\mathcal{E}\xspace}

\newcommand{\WM}{\mathbf{W}\xspace}
\newcommand{\AM}{\mathbf{A}\xspace}
\newcommand{\NAM}{\hat{\mathbf{A}}\xspace}
\newcommand{\NLM}{\hat{\mathbf{L}}\xspace}

\newcommand{\DM}{\mathbf{D}\xspace}
\newcommand{\IM}{\mathbf{I}\xspace}

\newcommand{\MM}{\mathbf{M}\xspace}
\newcommand{\PM}{\mathbf{P}\xspace}

\newcommand{\YM}{\mathbf{Y}\xspace}
\newcommand{\XM}{\mathbf{X}\xspace}
\newcommand{\LM}{\mathbf{L}\xspace}
\newcommand{\UM}{\mathbf{U}\xspace}

\newcommand{\HM}{\mathbf{H}\xspace}

\newcommand{\ZM}{\mathbf{Z}\xspace}

\newcommand{\eat}[1]{}

\newcommand{\QM}{\mathbf{Q}\xspace}
\newcommand{\LABM}{\boldsymbol{\Lambda}\xspace}

\newcommand{\alg}{\texttt{DDSM}\xspace}
\newcommand{\algp}{\texttt{DDSM} (PRDD)\xspace}
\newcommand{\algv}{\texttt{DDSM} (VDD)\xspace}
\newcommand{\algh}{\texttt{DDSM} (HKDD)\xspace}

\newenvironment{customlegend}[1][]{%
    \begingroup
    \csname pgfplots@init@cleared@structures\endcsname
    \pgfplotsset{#1}%
}{%
    \csname pgfplots@createlegend\endcsname
    \endgroup
}%

\def\addlegendimage{\csname pgfplots@addlegendimage\endcsname}

\makeatletter
\newcommand\footnoteref[1]{\protected@xdef\@thefnmark{\ref{#1}}\@footnotemark}
\makeatother

\let\oldnl\nl%
\newcommand{\nonl}{\renewcommand{\nl}{\let\nl\oldnl}}%

\definecolor{myred}{HTML}{fd7f6f}
\definecolor{myred_new}{HTML}{D8D8D8}
\definecolor{myred_new2}{HTML}{D7191C}
\definecolor{myblue}{HTML}{7eb0d5}
\definecolor{mygreen}{HTML}{b2e061}
\definecolor{mypurple}{HTML}{bd7ebe}
\definecolor{myorange}{HTML}{ffb55a}
\definecolor{myyellow}{HTML}{ffee65}
\definecolor{mypurple2}{HTML}{beb9db}
\definecolor{mypink}{HTML}{fdcce5}
\definecolor{mycyan}{HTML}{8bd3c7}

\definecolor{myblue2}{HTML}{115f9a}
\definecolor{myred2}{HTML}{c23728}

\definecolor{PageRank-color}{HTML}{FC9871}
\definecolor{Vanilla-color}{HTML}{377483}
\definecolor{HeatKernel-color}{HTML}{4D4D9F}

\newcommand{\cmark}{\ding{51}}%
\newcommand{\xmark}{\ding{55}}%

\AtBeginDocument{%
  \providecommand\BibTeX{{%
    \normalfont B\kern-0.5em{\scshape i\kern-0.25em b}\kern-0.8em\TeX}}}

\setcopyright{acmcopyright}
\copyrightyear{2018}
\acmYear{2018}
\acmDOI{XXXXXXX.XXXXXXX}

\acmPrice{15.00}
\acmISBN{978-1-4503-XXXX-X/18/06}

\acmSubmissionID{982}

\begin{document}

\title{Rethinking Message Passing Neural Networks with Diffusion Distance-guided Stress Majorization}
\subtitle{Technical Report}

\author{Haoran Zheng}
\affiliation{%
  \institution{Hong Kong Baptist University}
  \country{Hong Kong SAR, China}
}
\email{cshrzheng@comp.hkbu.edu.hk}
\orcid{0009-0002-4769-3716}

\author{Renchi Yang}
\authornote{Corresponding Author}
\affiliation{%
  \institution{Hong Kong Baptist University}
  \country{Hong Kong SAR, China}
}
\email{renchi@hkbu.edu.hk}
\orcid{0000-0002-7284-3096}

\author{Yubo Zhou}
\authornote{Work done while at HKBU.}
\affiliation{%
  \institution{University of Michigan}
  \city{Ann Arbor, Michigan}
  \country{USA}
}
\email{yubozhou@umich.edu}

\author{Jianliang Xu}
\affiliation{%
  \institution{Hong Kong Baptist University}
  \country{Hong Kong SAR, China}
}
\email{xujl@comp.hkbu.edu.hk}
\orcid{0000-0001-9404-5848}

\settopmatter{printfolios=true}

\renewcommand{\shortauthors}{Zheng et al.}

\begin{abstract}
Message passing neural networks (MPNNs) have emerged as go-to models for learning on graph-structured data in the past decade.
Despite their effectiveness, most of such models still incur severe issues such as over-smoothing and -correlation, due to their underlying objective of minimizing the Dirichlet energy and the derived neighborhood aggregation operations.
In this paper, we propose the \alg, a new MPNN model built on an optimization framework that includes the {\em stress majorization} and {\em orthogonal regularization} for overcoming the above issues. Further, we introduce the {\em diffusion distances} for nodes into the framework to guide the new message passing operations and develop efficient algorithms for distance approximations, both backed by rigorous theoretical analyses.
Our comprehensive experiments showcase that \alg consistently and considerably outperforms 15 strong baselines on both homophilic and heterophilic graphs.
\end{abstract}



\maketitle

\section{Introduction}

\textit{Graph Neural Networks} (GNNs) are powerful models for learning representations on graph-structured data~\cite{wu2020comprehensive}.
Their applications span diverse domains, including  social networks~\cite{fan2019graph, zhang2022improving}, recommendation systems~\cite{wu2019session, ying2018graph}, and Bioinformatics~\cite{li2021dgllife}.
The remarkable success of most GNNs primarily relies on the message-passing mechanism therein, where nodes recursively aggregate and transform information from their neighbors along the edges.
By following this paradigm, such {\em message passing neural networks} (MPNNs)~\cite{gilmer2017neural} can effectively capture both local and global graph structures.

Recent studies~\cite{ma2021unified,zhu2021interpreting} have uncovered that the majority of existing MPNNs essentially minimize the {\em Dirichlet energy}~\cite{zhou2005regularization} of the node representations $\HM$ over the graph, which seek to reduce the discrepancy between the representations of adjacent nodes. Based thereon, our theoretical analyses pinpoint that this optimization objective will eventually engender both node- and feature-wise issues in $\HM$.
More concretely, from the node-wise perspective of $\HM$, as the number of layers increases, the message passing operations not only lead to the well-documented {\em global over-smoothing}~\cite{li2018deeper,oono2020graph}, 
but also will easily over-mix the features of nodes connected via noisy/heterophilic links, i.e., adjacent nodes with distinct labels, caused the {\em heterophilic over-smoothing} problem. In turn, node representations, particularly in high-degree~\cite{lai2024efficient} and heterophilic graphs, are rendered hard to distinguish.
On the other hand, the feature dimensions and spaces of node representations $\HM$ by MPNNs will be highly correlated in the limit of many layers~\cite{jin2022feature,sun2023feature}, resulting in severe feature deficiency and limited expressiveness.
Our empirical studies reveal that these problems engender conspicuous model performance degradation, even in MPNNs with a handful of layers.

In the literature, numerous workarounds have been explored to alleviate some of these issues. For global over-smoothing, common techniques include orthogonal weighting~\cite{zhou2021dirichlet}, feature normalization~\cite{zhao2020pairnorm,zhou2020towards}, edge masking~\cite{rong2020dropedge}, and skip connections~\cite{chen2020simple,xu2018representation,gasteiger2019combining,li2021training}. To handle heterophily, some researchers have developed specialized models~\cite{zheng2022graph,luan2024heterophilic}, though these often lack general applicability. For feature over-correlation, a few attempts have introduced explicit decorrelation losses~\cite{jin2022feature}. Concurrently, another promising direction leverages diffusion and physics-inspired principles. These approaches either reformulate the learning objective using concepts like stress functions or reaction-diffusion systems~\cite{cui2023mgnn, li2022tired, choi2023gread}, or use diffusion processes to directly rewire the graph for message passing~\cite{gasteiger2019diffusion, begga2023diffusion,huang2023node,xie2025diffusion}. However, these advanced methods often rely on learnable distances which can be unstable, or they use diffusion primarily to alter the message-passing topology rather than to define a global learning objective. The ideal choice of a principled distance metric and its role within a robust optimization framework remain underexplored. Thus, a unified solution that addresses the node-wise and feature-wise issues by tackling their fundamental causes is still needed.

Motivated by this, we present \underline{D}iffusion \underline{D}istance-guided \underline{S}tress \underline{M}ajorization
(\alg), a novel MPNN model that overcomes the aforementioned deficiencies through a new design of the optimization framework.
More specifically, unlike the Dirichlet energy-based objective in existing MPNNs that encourages node representations to be overly smoothed and correlated, \alg draws inspiration from the {\em stress majorization} adopted in {\em graph drawing}~\cite{cohen1997drawing}, whose idea is to impose a predefined distance between nodes to prevent their overlapping and overspacing in the visualization. In the same vein, \alg can circumvent {\em over-smoothing} and {\em over-sharpening}~\cite{di2023understanding} in $\HM$ with an appropriate distance measure for nodes.
Additionally, \alg incorporates an orthogonal regularization of $\HM$ into the optimization objective to decorrelate the features, leading to our new message passing operations.

Building on the above framework, we propose to employ the {\em diffusion distances}~\cite{nadler2005diffusion, de2008hierarchical} in \alg after a careful investigation of existing distance metrics in both theoretical and empirical aspects. In particular, competing metrics such as shortest path~\cite{dijkstra1959note}, Jaccard~\cite{kosub2019note}, resistance~\cite{Klein1993}, biharmonic~\cite{lipman2010biharmonic}, and learnable distances~\cite{cui2023mgnn} either are unbounded or fail to capture the global structure, and thus, are incompetent for the role in \alg.
By contrast, diffusion distances rely on node connectivity patterns via random walk distributions, offering multiple desired properties pertaining to the range, structural preservation, robustness to noise/perturbations, etc., as unveiled by our analysis.
On top of that, we develop a fast and theoretically-grounded approach for diffusion distance approximation via the truncated spectral decomposition.

Our experiments on 11 benchmark datasets demonstrate the consistent superiority of \alg over 15 baselines in terms of supervised node classification.

\section{Related Work}
\subsection{Classic Message Passing Neural Networks}
GNNs can be broadly categorized into spectral-based and spatial-based approaches. Spectral-based GNNs leverage graph signal processing techniques and spectral graph theory to define convolution operations in the Fourier domain. For example, \texttt{ChebNet}~\cite{defferrard2016convolutional} uses Chebyshev polynomials to approximate localized filters, while \texttt{GCN}~\cite{kipf2017semisupervised} simplifies these filters to a 1-hop neighborhood aggregation. \texttt{GWNN}~\cite{xu2018graph} introduces wavelet-based transformations for spectral graph convolution. On the other hand, spatial-based GNNs focus on feature aggregation directly over the graph topology. Models like \texttt{GCN}, \texttt{GAT}~\cite{veličković2018graph}, and \texttt{SAGE}~\cite{hamilton2017inductive} employ varying aggregation strategies such as attention mechanisms, pooling, and adaptive policies. Recent studies~\cite{bachmann2020constant, lim2021large, luan2022revisiting, li2022finding, xu2018how, chen2020simple,zhou2023slotgat,wang2023efficient,wang2024effective} also explore advanced topics, including GNNs in non-Euclidean spaces, heterogeneous graphs, and robust graph learning, among others.

\subsection{Graph Filtering in MPNNs}
Recent studies have linked MPNNs to concepts in graph signal processing, offering valuable insights into their propagation mechanisms and guiding their design. Several studies~\cite{li2018deeper, wu2019simplifying, nt2019revisiting,wang2025gegennet} demonstrate that graph convolution operations, as used in GNNs, function as forms of Laplacian smoothing or low-pass filters. Recent works~\cite{ma2021unified,zhu2021interpreting} have further unified various GNN propagation mechanisms into a single framework. Starting from the optimization function in Eq. (\ref{eq:obj}), \texttt{GNN-LF/HF}~\cite{ma2021unified} derives low-pass and high-pass filters grounded in the optimization objective. Subsequent studies, such as~\cite{luan2022revisiting, huang2024how}, extend this idea by leveraging the optimization function to design propagation mechanisms that combine low-pass and high-pass filters or employ polynomial filters. Nevertheless, these approaches do not focus on refining the Dirichlet energy term in the optimization objective. A recent advancement, \texttt{MGNN}~\cite{cui2023mgnn}, incorporates a learnable distance metric through edge attention to refine the Dirichlet energy term during training. Nonetheless, the efficacy of edge attention might be constrained, as it overlooks the graph topology and fails to ensure a proper range.

\subsection{Diffusion-improved Graph Learning}
Recent works leverage diffusion principles to address key GNN challenges like over-smoothing. One line of work directly integrates diffusion into message passing, such as \texttt{GDC}~\cite{gasteiger2019diffusion} using generalized graph diffusion and \texttt{TIDE}~\cite{behmanesh2023tide} employing learnable time-derivative diffusion, and \texttt{HiD-Net}~\cite{li2024generalized}, which proposes a generalized framework incorporating high-order diffusion. Others design novel architectures inspired by diffusion, like \texttt{BuNNs}~\cite{bamberger2024bundle} and \texttt{DIFFormer}~\cite{wu2024neural}. Another popular approach, inspired by physics and graph drawing, uses a stress-like objective with attractive/repulsive forces to maintain ideal node distances and prevent feature collapse, as seen in \texttt{StressGNN}~\cite{li2022tired}, \texttt{MGNN}~\cite{cui2023mgnn}, and \texttt{ACMP}~\cite{wang2022acmp}. Similarly, models like \texttt{GREAD}~\cite{choi2023gread} leverage reaction-diffusion systems to balance feature blurring and sharpening across diverse graph types. Finally, methods like \texttt{DRew}~\cite{gutteridge2023drew} and {Diffusion-Jump GNNs}~\cite{begga2023diffusion} focus on adaptively rewiring the graph, either through multi-hop edges or shortcuts based on diffusion distances.

Our proposed \alg advances these efforts via a new design of the optimization objective based on the stress majorization, which overcomes the deficiencies of existing Dirichlet energy-based and stress-like objectives~\cite{li2022tired, cui2023mgnn} by innovatively integrating the diffusion distance and orthogonal regularization into the framework.

\section{Background}
\stitle{Graph Nomenclature} Let $\G=(\V,\EDG)$ be a graph, where $\V$ and $\EDG$ are a set of $n=|\V|$ nodes and $m=|\EDG|$ edges, respectively. For each edge $e_{i,j}\in \EDG$, we say $v_i$ and $v_j$ are neighbors to each other, and use $\N(v_i)$ to symbolize the neighbors of $v_i$, where the degree of $v_i$ is $d_i=|\N(v_i)|$. 
$\AM$ and $\DM$ denote the adjacency and degree matrices of $\G$. $\PM=\DM^{-1}\AM$ (resp. $\NAM=\DM^{-1/2}\AM\DM^{-1/2}$) is defined as the transition (resp. normalized adjacency) matrix of $\G$. Accordingly, $\PM^t_{i,j}$ stands for the probability of a simple random walk starting from $v_i$ visiting $v_j$ at the $t$-th hop.
$\LM=\DM-\AM$ and $\NLM=\IM-\NAM$ denote the Laplacian and normalized Laplacian of $\G$, respectively.
For weighted graphs where each edge $(v_i,v_j)\in \EDG$ is associated with a weight $w_{i,j}$, the definitions of the above matrices naturally follow by letting \(\AM_{i,j}=w_{i,j}\) and \( \DM_{i,i} = \sum_{v_j\in \N(v_i)} w_{i,j}\).

\stitle{Dirichlet Engry} The {\em Dirichlet energy}~\cite{zhou2005regularization} of matrix $\XM\in \mathbb{R}^{n\times d}$ over the graph $\G$ is defined by
\begin{equation}\label{eq:DE}
\mathcal{D}(\XM):= \frac{1}{2}\sum_{(v_i,v_j)\in \EDG}{\left\|\frac{\XM_i}{\sqrt{d_i}}-\frac{\XM_j}{\sqrt{d_j}}\right\|^2_2},
\end{equation}
which measures the {\em smoothness} of $\XM$ over $\G$, indicating whether signal values in $\XM$ are similar across adjacent nodes.

\stitle{Homophily Ratio} 
Given a set of classes $\mathcal{Y}$ and the class label $y_i\in \mathcal{Y}$ for each node $v_i\in \V$, the {\em homophily ratio} (HR) of $\G$ is calculated by \(h(\G)=\frac{|\{(v_i,v_j)\in \EDG: y_i=y_j\}|}{m}\), which quantifies the fraction of homophilic edges that connect nodes of the same classes~\cite{zhu2020beyond}.

\subsection{Diffusion Distance}\label{sec:diffusion-distance}
{\em Diffusion distance} is a metric for measuring the distance between two nodes based on their connectivity patterns via random walks over the graph. 
By the random walk models adopted, it can be categorized into the following three types.

The {\em vanilla diffusion distance} (VDD)~\cite{nadler2005diffusion, coifman2006diffusion} of any node pair $v_i,v_j$ essentially quantifies the probability distributions of length-$t$ random walks originating from $v_i$ and $v_j$. Mathematically,
\begin{align}\label{eq:DD}
\Delta_{\text{vanilla}}(v_i, v_j) 
&= \|(\PM^t\DM^{-1/2})_i-(\PM^t\DM^{-1/2})_j\|_2.
\end{align}
By substituting heat kernel-based~\cite{chung2007heat} and PageRank-based~\cite{jeh2003scaling} random walk models for the above length-$t$ random walks, the {\em heat kernel diffusion distance} (HKDD)~\cite{de2008hierarchical} and {\em PageRank diffusion distance} (PRDD) are defined as
\begin{equation}\label{eq:hkdd}
\begin{gathered}
\Delta_{\text{HK}}(v_i, v_j) = \left\|\sum_{t=0}^{\infty}{\frac{e^{-\gamma}\gamma^t}{t!}\PM^t_i}-\sum_{t=0}^{\infty}{\frac{e^{-\gamma}\gamma^t}{t!}\PM^t_j}\right\|_2,\\
\Delta_{\text{PR}}(v_i, v_j) 
= \left\|\sum_{t=0}^{\infty}{\gamma^t(\PM^t\DM^{-\frac{1}{2}})_i}-{\gamma^t(\PM^t\DM^{-\frac{1}{2}})_j}\right\|_2,
\end{gathered}
\end{equation}
where \( \gamma \) stands for the heat constant and teleportation factor in both definitions, respectively. The diffusion distances are bounded as in the following lemma. All missing proofs appear in Appendix~\ref{sec:proofs}.
\begin{lemma}\label{lem:dd-range}
Let $d_{\min}=\min_{v_k\in\V}{d_k}$. Then, \(\forall{v_i,v_j\in \V}\), \(\Delta_{\textnormal{vanilla}} \allowbreak(v_i, v_j)\in[0,\frac{\sqrt{2}}{d_{\min}}], \Delta_{\textnormal{PR}}(v_i, v_j)\in[0,\frac{\sqrt{2}}{(1-\gamma) d_{\min}}] \) and \(\Delta_{\textnormal{HK}}(v_i, v_j)\in[0,\sqrt{2}]\).
\end{lemma}

\subsection{Message Passing Neural Networks (MPNNs)}
MPNNs are a class of GNNs that operate under the {\em message passing} mechanism, wherein nodes exchange messages (i.e., features) along the edges to refine their representations.
Assume that the graph $\G$ is accompanied by node attributes $\XM\in \mathbb{R}^{n\times d^{(0)}}$, where $\XM_i$ symbolizes the attribute vector of node $v_i\in \V$. Denote by $\HM^{(k)}_i$ as the node representation of $v_i$ at the $k$-th ($k\geq 0$) layer with $\HM^{(0)}=f(\XM)$, where $f(\cdot)$ can be an identity mapping, linear transformation, or MLP.
The output node representation at $(k+1)$-th layer in a generic MPNN can be formulated using the following recursion~\cite{gilmer2017neural}:
\begin{equation}\label{eq:mpnn-update-rule}
\HM^{(k+1)}_i = \phi_{k}\Bigg(\HM^{(k)}_i, \psi_k\Big(\sum_{v_j\in \N(v_i)}\NAM_{i,j} \HM^{(k)}_j\Big) \Bigg),
\end{equation}
where $\psi_{k}: \mathbb{R}^{d_k}\times\mathbb{R}^{d_k}\rightarrow \mathbb{R}^{d^\prime_k}$ and $\phi_k: \mathbb{R}^{d_k}\times\mathbb{R}^{d^\prime_k}\rightarrow \mathbb{R}^{d_{k+1}}$ are learnable message and update functions (a.k.a. aggregation and transformation) that are typically implemented as MLPs. 
Well-known MPNNs include \texttt{GCN}~\cite{kipf2017semisupervised}, \texttt{SAGE}~\cite{hamilton2017inductive}, \texttt{APPNP}~\cite{gasteiger2019combining}, \texttt{GCNII}~\cite{chen2020simple}, etc.

\section{Limitation Analyses of MPNNs}
This section begins with unifying existing MPNNs into a framework optimizing the Dirichlet energy, followed by analyzing the deficiencies of node representations from MPNNs from the perspective of nodes and features, respectively.

\subsection{Dirichlet Energy Minimization Perspective}
As demystified in recent studies~\cite{ma2021unified,zhu2021interpreting}, 
the learning process of most MPNNs can be unified into a framework optimizing
\begin{equation}\label{eq:GNN-obj}
\arg\min_{\HM} \omega\cdot\mathcal{D}(\HM) + \xi,
\end{equation}
where the first term is the Dirichlet energy of the target node representations $\HM$, $\omega$ is a scaling parameter, and $\xi$ stands for an additional term, which is neglected in GCN/SGC~\cite{kipf2017semisupervised,wu2019simplifying} or can be the $L_2$-penalty in other MPNNs.
Accordingly, by setting the derivative of Eq.~\eqref{eq:GNN-obj} w.r.t. $\HM$ to zero, the closed-form solution for $\HM$ can be obtained.
Table~\ref{tbl:GNNs} in Appendix~\ref{sec:add-tables} summarizes the choices of $\omega$, $\xi$ and their corresponding formulations of $\HM$ in various MPNNs.
Corollary 3.4.1 in \cite{cai2020note} reveals that the Dirichlet energy $\mathcal{D}(\HM)$ decreases monotonically as the model layers increase.

\subsection{Node-wise Issues}
Next, we discuss the issues of node representations $\HM$ by MPNNs from the perspective of nodes.

\stitle{Global Over-smoothing} As remarked above, MPNNs essentially minimize the Dirichlet energy of $\HM$, which encourages embeddings of adjacent nodes to be similar.  Consequently, the neighborhood aggregation operations render features of all nodes indistinguishable in the limit of many layers, i.e., \(\mathcal{D}(\HM)\rightarrow 0\) and \(\left\|\frac{\HM_i}{\sqrt{d_i}}-\frac{\HM_j}{\sqrt{d_j}}\right\|^2_2 \rightarrow 0\ \forall{v_i,v_j\in \EDG}\), which is referred to as {\em over-smoothing}~\cite{li2018deeper,oono2020graph} in the literature. 
\begin{lemma}\label{lem:oversmooth}
Let $\boldsymbol{\pi} \in \mathbb{R}^n$ be the vector in which $\boldsymbol{\pi}_i =  \sqrt{\frac{d_i}{2m}}\ \forall{v_i\in \V}$. Then \(\lim_{k \to \infty} {\NAM}^k = \boldsymbol{\pi} \boldsymbol{\pi}^\top\).
\end{lemma}
For instance, by Lemma~\ref{lem:oversmooth}, the eventual embedding $\HM_i$ of each node $v_i\in \V$ by GCN/SGC in the limit of many layers can be represented by
\(\HM_i = \sqrt{d_i}\sum_{v_j\in \V}{\frac{\sqrt{d_j}}{2m}\cdot\HM^{(0)}_j}\),
which is the same for all nodes with merely a different scaling factor $\sqrt{d_i}$, leading to poor model performance in downstream tasks. This phenomenon is exacerbated in graphs with high node degrees~\cite{lai2024efficient}.

\begin{lemma}\label{lem:homophily-DE}
\(h(\G) = 1-\frac{\mathcal{D}(\DM^{1/2}\YM)}{m}\).
\end{lemma}

\stitle{Heterophilic Over-smoothing}
Let $\YM\in \mathbb{R}^{n\times |\Y|}$ contain the node labels of $\G$, where $\YM_{i,k}=1$ if $y_i=k$ and $0$ otherwise. Our Lemma~\ref{lem:homophily-DE} establishes the relation between the homophily ratio of $\G$ and the Dirichlet energy of node labels, meaning that the optimization goal of MPNNs in Eq.~\eqref{eq:GNN-obj} is equivalent to inferring node labels maximizing the homophily ratio over $\G$.

Intuitively, this learning paradigm exhibits high efficacy on {\em homophilic graphs}, i.e., graphs with large homophily ratios, but largely fails for {\em heterophilic graphs} with low homophily or nodes with many cross-class connections (hereinafter {\em heterophilic links}) in both.
For example, 
for the real-world Wikipedia network {\em Chameleon}~\cite{rozemberczki2021multi}, the homophily ratio is merely 0.23, meaning that around $77\%$ edges connect nodes with different labels.
In turn, minimizing Eq.~\eqref{eq:GNN-obj} renders the representations of most adjacent nodes with distinct labels overly mixed and hard to distinguish, and hence, dwarfs the abilities of classifiers.
This also leads to compromised performance of MPNNs on homophilic graphs due to the existence of heterophilic links.

\subsection{Feature-wise Issues}\label{sec:feat-wise}
Aside from node-wise issues, $\HM$ also incurs feature-wise problems, e.g., {\em feature dimension over-correlation}~\cite{jin2022feature} and {\em feature space over-correlation\allowbreak}~\cite{sun2023feature}.

\begin{lemma}
\label{thm:overcorrelation}
\(\lim_{k \to \infty} ({\NAM}^k \HM^{(0)})^\top ({\NAM}^k \HM^{(0)}) = ({\HM^{(0)}}^\top \boldsymbol{\pi})\cdot(\boldsymbol{\pi}^\top \HM^{(0)})\), which is a rank-one matrix.
\end{lemma}

Specifically, feature dimension over-correlation refers to the fact that the feature dimensions of $\HM$ become highly correlated to each other as the number of layers in the MPNNs increases. 
As pinpointed by our theoretical analysis in Lemma~\ref{thm:overcorrelation}, 
$\HM^\top\HM$ will evolve into a rank-one matrix in the limit of many layers, implying that it deviates from being a diagonal matrix, and thus, columns in $\HM$ are overly correlated and are linearly dependent.
This issue causes feature redundancy among the learned representations, meaning that only a paucity of feature dimensions in $\HM$ are informative, while others are duplicative, which in turn degrades the representation effectiveness.

\begin{lemma}[\cite{sun2023feature}]\label{lem:featspace-corr}
There exist weights $\WM$ s.t. \(\|\NAM^k\HM^{(0)}\WM-\NAM^{k+\tau}\allowbreak\HM^{(0)}\|_F^2 \to 0\) as $\tau \in \mathbb{Z}$ increases.
\end{lemma}

Denote by $\NAM^k\HM^{(0)}$ a feature space. As in Table~\ref{tbl:GNNs}, $\HM$ can be expressed as a set of feature spaces, i.e., \(\{\NAM^k\HM^{(0)}|k\in [0,K]\}\).
Lemma~\ref{lem:featspace-corr} indicates that feature spaces will be {\em linearly correlated} to each other, i.e., a feature space can be represented as a linear multiple of others, due to the iterative neighborhood aggregations over the graph~\cite{sun2023feature}. As a aftermath, this feature space over-correlation issue leads to information redundancy and limited expressiveness in $\HM$.

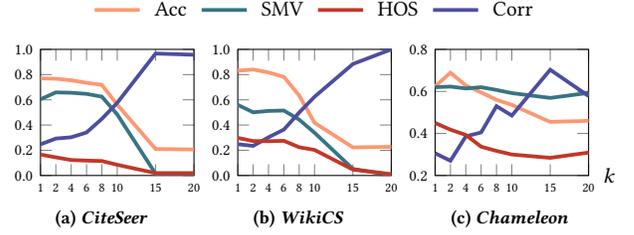
\begin{figure}[!t]
\centering
\begin{small}
\begin{tikzpicture}
    \begin{customlegend}[legend columns=4,
        legend entries={Acc, SMV, HOS, Corr},
        legend style={at={(0.45,1.35)},anchor=north,draw=none,font=\small,column sep=0.1cm}]
    \addlegendimage{line width=0.5mm,mark size=1pt,mark=dot, color=PageRank-color}
    \addlegendimage{line width=0.5mm,mark size=1pt,mark=dot, color=Vanilla-color}
    \addlegendimage{line width=0.5mm,mark size=1pt,mark=dot, color=myred2}
    \addlegendimage{line width=0.5mm,mark size=1pt,mark=dot, color=HeatKernel-color}
    \end{customlegend}
\end{tikzpicture}
\\[-\lineskip]
\vspace{-2ex}
\subfloat[{\em CiteSeer}]{
\begin{tikzpicture}[scale=1,every mark/.append style={mark size=1pt}]
    \begin{axis}[
        height=\columnwidth/2.6,
        width=\columnwidth/2.34,
        xmin=0, xmax=20,
        ymin=0, ymax=1,
        xtick={0,2,4,6,8,10,15,20},
        ytick={0.0,0.2,0.4,0.6,0.8,1.0},
        xticklabel style = {font=\tiny},
        yticklabel style = {font=\scriptsize},
        xticklabels={1,2,4,6,8,10,15,20},
        yticklabels={0.0,0.2,0.4,0.6,0.8,1.0},
    ]
    \addplot[line width=0.5mm, mark=dot, color=PageRank-color]
        plot coordinates {
(0  ,	0.770	)
(2  ,	0.767	)
(4  ,	0.755	)
(6  ,	0.735	)
(8  ,	0.718	)
(10 ,	0.562	)
(15 ,	0.210	)
(20 ,	0.206	)
    };

    \addplot[line width=0.5mm, mark=dot, color=Vanilla-color]
        plot coordinates {
(0  ,	0.604	)
(2  ,	0.659	)
(4  ,	0.656	)
(6  ,	0.647	)
(8  ,	0.625	)
(10 ,	0.484	)
(15 ,	0.014	)
(20 ,	0.014	)
    };

    \addplot[line width=0.5mm, mark=dot, color=HeatKernel-color]
        plot coordinates {
(0  ,	0.247	)
(2  ,	0.293	)
(4  ,	0.303	)
(6  ,	0.341	)
(8  ,	0.449	)
(10 ,	0.576	)
(15 ,	0.967	)
(20 ,	0.957	)
    };
    
    \addplot[line width=0.5mm, mark=dot, color=myred2]
        plot coordinates {
(0  ,	0.166	)
(2  ,	0.144	)
(4  ,	0.123	)
(6  ,	0.118	)
(8  ,	0.115	)
(10 ,	0.085	)
(15 ,	0.019	)
(20 ,	0.019	)
    };

    \end{axis}
\end{tikzpicture}\hspace{-1mm}\label{fig:over-smoothing-citeseer}%
}
\subfloat[{\em WikiCS}]{
\begin{tikzpicture}[scale=1,every mark/.append style={mark size=1pt}]
    \begin{axis}[
        height=\columnwidth/2.6,
        width=\columnwidth/2.34,
        xmin=0, xmax=20,
        ymin=0, ymax=1,
        xtick={0,2,4,6,8,10,15,20},
        ytick={0.0,0.2,0.4,0.6,0.8,1.0},
        xticklabel style = {font=\tiny},
        yticklabel style = {font=\scriptsize},
        xticklabels={1,2,4,6,8,10,15,20},
        yticklabels={0.0,0.2,0.4,0.6,0.8,1.0},
    ]
    \addplot[line width=0.5mm, mark=dot, color=PageRank-color]
        plot coordinates {
(0  ,	0.831	)
(2  ,	0.840	)
(4  ,	0.817	)
(6  ,	0.782	)
(8  ,	0.634	)
(10 ,	0.418	)
(15 ,	0.223	)
(20 ,	0.227	)
    };

    \addplot[line width=0.5mm, mark=dot, color=Vanilla-color]
        plot coordinates {
(0  ,	0.560	)
(2  ,	0.501	)
(4  ,	0.512	)
(6  ,	0.515	)
(8  ,	0.443	)
(10 ,	0.339	)
(15 ,	0.052	)
(20 ,	0.004	)
    };

    \addplot[line width=0.5mm, mark=dot, color=HeatKernel-color]
        plot coordinates {
(0  ,	0.248	)
(2  ,	0.235	)
(4  ,	0.303	)
(6  ,	0.364	)
(8  ,	0.499	)
(10 ,	0.624	)
(15 ,	0.884	)
(20 ,	1.000	)
    };
    
    \addplot[line width=0.5mm, mark=dot, color=myred2]
        plot coordinates {
(0  ,	0.298	)
(2  ,	0.273	)
(4  ,	0.272	)
(6  ,	0.275	)
(8  ,	0.224	)
(10 ,	0.203	)
(15 ,	0.047	)
(20 ,	0.011	)
    };

    \end{axis}
\end{tikzpicture}\hspace{-1mm}\label{fig:over-smoothing-wikics}%
}
\subfloat[{\em Chameleon}]{
\begin{tikzpicture}[scale=1,every mark/.append style={mark size=1pt}]
    \begin{axis}[
        height=\columnwidth/2.6,
        width=\columnwidth/2.34,
        xlabel={$k$},
        xmin=0, xmax=20,
        ymin=0.2, ymax=0.8,
        xtick={0,2,4,6,8,10,15,20},
        ytick={0.0,0.2,0.4,0.6,0.8,1.0},
        xticklabel style = {font=\tiny},
        yticklabel style = {font=\scriptsize},
        xticklabels={1,2,4,6,8,10,15,20},
        yticklabels={0.0,0.2,0.4,0.6,0.8,1.0},
        every axis x label/.style={
        at={(ticklabel* cs:1.05)},
        anchor=west,
        },
    ]
    \addplot[line width=0.5mm, mark=dot, color=PageRank-color]
        plot coordinates {
(0  ,	0.624	)
(2  ,	0.689	)
(4  ,	0.628	)
(6  ,	0.595	)
(8  ,	0.560	)
(10 ,	0.536	)
(15 ,	0.455	)
(20 ,	0.460	)
    };

    \addplot[line width=0.5mm, mark=dot, color=Vanilla-color]
        plot coordinates {
(0  ,	0.620	)
(2  ,	0.623	)
(4  ,	0.614	)
(6  ,	0.620	)
(8  ,	0.607	)
(10 ,	0.592	)
(15 ,	0.569	)
(20 ,	0.593	)
    };

    \addplot[line width=0.5mm, mark=dot, color=HeatKernel-color]
        plot coordinates {
(0  ,	0.306	)
(2  ,	0.270	)
(4  ,	0.386	)
(6  ,	0.405	)
(8  ,	0.530	)
(10 ,	0.485	)
(15 ,	0.703	)
(20 ,	0.578	)
    };
    
    \addplot[line width=0.5mm, mark=dot, color=myred2]
        plot coordinates {
(0  ,	0.450	)
(2  ,	0.418	)
(4  ,	0.393	)
(6  ,	0.337	)
(8  ,	0.317	)
(10 ,	0.300	)
(15 ,	0.284	)
(20 ,	0.309	)
    };

    \end{axis}
\end{tikzpicture}\label{fig:over-smoothing-chameleon}%
}
\end{small}
 \vspace{-2ex}
\caption{Acc, SMV, Corr, and HOS of $\HM^{(k)}$ in GCN.} \label{fig:over-smoothing}
\vspace{-3ex}
\end{figure}

\subsection{Empirical Studies}
Based on the above theoretical observations, we examine the node-wise and feature-wise issues on real homophilic and heterophilic graphs. 

We adopt the SMV (Eq.~\eqref{eq:SMV})~\cite{liu2020towards} and Corr (Eq.~\eqref{eq:Corr})~\cite{jin2022feature} for measuring the global over-smoothing and feature dimension over-correlation, respectively.
As for heterophilic over-smoothing, we define HOS: 
\begin{small}
\begin{equation*}
\text{HOS}(\HM^{(k)}) = \frac{1}{2|\EDG^{\prime}|} \sum_{(v_i,v_j)\in \EDG^{\prime}}\left\| {\HM^{(k)}_{i}}/{\|\HM^{(k)}_{i}\|_2}- {\HM^{(k)}_{j}}/{\|\HM^{(k)}_{j}\|_2}\right\|_2,
\end{equation*}
\end{small}
where $\EDG^{\prime}$ contains all heterophilic links in the graph.
Particularly, the smaller \(\text{SMV}(\HM^{(k)})\) (resp. \(\text{HOS}(\HM^{(k)})\)) is, the smoother the embeddings of all nodes (nodes involving heterophilic links) are. 
In contrast, a larger \(\text{Corr}(\HM^{(k)})\) indicates a higher correlation of feature dimensions of \(\HM^{(k)}\).

Fig.~\ref{fig:over-smoothing} plots the classification accuracy (Acc), SMV, HOS, and Corr values of \(\HM^{(k)}\) output at each $k$-th ($\in[1,20]$) layer in \texttt{GCN} on the homophilic graphs {\em CiteSeer}, {\em WikiCS}, and heterophilic graph {\em Chameleon}.
It can be observed that as layers increase, Acc, SMV, and HOS decrease while Corr grows, validating the adverse impacts of over-smoothing and over-correlation on classification performance. Compared to SMV, the changes in Acc are more consistent with those in HOS and Corr when $k$ is small (\(\in[1,6]\)).
In particular, the HOS is substantially lower than the SMV even after a few layers, indicating that adjacent nodes with different ground-truth labels will have closer embeddings and are very likely to be wrongly predicted as the same class. 

Such empirical observations imply that the over-correlation and heterophilic over-smoothing make critical bottlenecks of shallow GNNs on both homophilic and heterophilic graphs, while global over-smoothing is more pronounced in deeper models.
More empirical results are in Appendix~\ref{sec:exp-issues}.

\section{Methodology}\label{sec:methodology}

In this section, we propose \alg to address the aforementioned fundamental deficiencies of classic MPNNs from the perspective of optimizations. We first delineate the basic idea of \alg in Section~\ref{sec:overview}, followed by our diffusion distance-based optimization objective and derived a new message passing scheme in Section~\ref{sec:dist-MPNNs}. In Section~\ref{sec:distance-computation}, we elaborate on our spectral decomposition techniques for efficient computation of diffusion distances.

\begin{figure}[!t]
\centering
\includegraphics[width=0.9\linewidth]{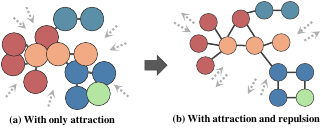}
\vspace{-2ex}
\caption{Illustration of the basic idea of \alg{}.}
\label{fig:viz-example}
\vspace{-2ex}
\end{figure}

\subsection{Design Overview}\label{sec:overview}%

As unveiled in the preceding section, the global and heterophilic over-smoothing issues inside MPNNs emanate from the objective of minimizing the Dirichlet energy of $\HM$ (see Eq.~\eqref{eq:GNN-obj}), which pulls the features of connected nodes as closer to each other as possible in the embedding space. 
By regarding node features as particles, such a Dirichlet energy-based objective solely imposes the force of attraction along the edges without repulsion.
As illustrated in {Fig.~\ref{fig:viz-example} (a)}, on a 2D layout, nodes are thus pulled to clutter with severe node overlap, leading to poor visualizations.
In the context of graph layout~\cite{cohen1997drawing}, a common treatment is to introduce the repulsive forces between nodes such that they remain properly distanced (see {Fig.~\ref{fig:viz-example} (b)). In the prominent {\em stress majorization}~\cite{gansner2005graph}, the layout is generated by optimizing the stress function as follows:
\begin{equation}\label{eq:stress}
\textstyle \sum_{v_i,v_j\in \V}w_{i,j}{\left(\|\QM_i-\QM_j\|_2-\delta_{i,j}\right)^2}.
\end{equation}
It induces the position embeddings $\QM_i$ and $\QM_j$ of every two data points $i,j$ by aligning their actual distance with the desired distance $\delta_{i,j}$, and assigns a weight $w_{i,j}$ to each pair.

Inspired by this, we propose to substitute the following loss for the Dirichlet energy term \(\mathcal{D}(\HM)\) in Eq.~\eqref{eq:GNN-obj}:
\begin{equation}\label{eq:distance-loss}
\mathcal{S} = \sum_{(v_i,v_j)\in \EDG}{\left(\left\|\frac{\HM_i}{\sqrt{d_i}}-\frac{\HM_j}{\sqrt{d_j}}\right\|_2 - \delta_{i,j}\right)^2}.
\end{equation}
In particular, \(\mathcal{S}\) combines the merits of both \(\mathcal{D}(\HM)\) (see Eq.~\eqref{eq:DE}) and the stress function in Eq.~\eqref{eq:stress} by injecting graph-theoretic distance \(\delta_{i,j}\ \forall{(v_i,v_j)\in \EDG}\) into \(\mathcal{D}(\HM)\).
With \(\delta_{i,j}\) considering the holistic topological connectivity/similarity between nodes $v_i$ and $v_j$, we can avert pushing \(\HM_i\) and \(\HM_j\) too close solely based on the single (especially heterophilic and noisy) edge connecting them, thereby alleviating both global and heterophilic over-smoothing problems.

To further mitigate the feature correlation issue in MPNNs, according to our analysis in Section~\ref{sec:feat-wise}, the straightforward way is to prevent $\HM^\top\HM$ being rank-one but a full-rank matrix such that feature dimensions are decorrelated. As such, we additionally incorporate an orthogonal regularization Eq.~\eqref{eq:decorrelation} into the objective of MPNNs.
\begin{equation}\label{eq:decorrelation}
\mathcal{C} = \frac{1}{2}\|\HM^{\top}\HM-\IM\|^2_F.
\end{equation}
This term explicitly enforces feature dimensions of \(\HM\) to be orthogonal and penalizes them from being overly correlated.

\begin{table}[!t]
\centering
\caption{Comparison with existing graph-theoretic distance metrics (for node pairs in $\EDG$) in terms of properties.}\label{tbl:distances-concise}
\vspace{-2ex}
\addtolength{\tabcolsep}{-0.25em}
\resizebox{\columnwidth}{!}{
\begin{footnotesize}
\begin{tabular}{l|cccccccccc}
\toprule
 {\bf Distance} &	\rotatebox{90}{Non-negativity} 	&	\rotatebox{90}{Finity} 	&	\rotatebox{90}{Symmetry} 	&	\rotatebox{90}{Triangle Inequality} 	&	\rotatebox{90}{Edge Weights} 	&	\rotatebox{90}{Local Structure} 	&	\rotatebox{90}{Global Structure} 	&	\rotatebox{90}{Structural Similarity} 	&	\rotatebox{90}{\makecell{Range\\ (Directed graphs)}} 	&	\rotatebox{90}{\makecell{Range\\ (Undirected graphs)}} \\	\midrule
SPD~\cite{dijkstra1959note}	&	 \cmark 	&	 \xmark 	&	 \cmark 	&	 \cmark 	&	 \cmark 	&	 \cmark 	&	 \xmark 	&	 \xmark 	&	 \(\{1,+\infty\}\) 	&	 \(\{1\}\) 	\\
Jaccard~\cite{kosub2019note}	&	  \cmark 	&	  \cmark 	&	  \cmark 	&	  \cmark 	&	  \xmark 	&	  \cmark 	&	 \xmark  	&	 \xmark  	&	 \([\frac{2}{n},1]\)  	&	 \([\frac{2}{n},1]\)  	\\
Resistance~\cite{Klein1993}	&	 \cmark 	&	 \xmark 	&	 \cmark 	&	 \cmark 	&	 \cmark 	&	 \cmark 	&	 \cmark 	&	 \xmark 	&	 \((0,+\infty]\) 	&	 \((0,1]\) 	\\
Biharmonic~\cite{lipman2010biharmonic}	&	 \cmark 	&	 \xmark 	&	 \cmark 	&	 \cmark 	&	 \cmark 	&	 \cmark 	&	 \cmark 	&	 \xmark 	&	 undefined 	&	 \([\frac{\sqrt{2}}{2},\sqrt{2}nD]\) 	\\
VDD~\cite{nadler2005diffusion}	&	 \cmark 	&	 \cmark 	&	 \cmark 	&	 \cmark 	&	 \cmark 	&	 \cmark 	&	 \cmark 	&	 \cmark 	&	 \([0,\frac{\sqrt{2}}{d_{\min}}]\) 	&	 \([0,\frac{\sqrt{2}}{d_{\min}}]\) 	\\
PRDD	&	 \cmark 	&	 \cmark 	&	 \cmark 	&	 \cmark 	&	 \cmark 	&	 \cmark 	&	 \cmark 	&	 \cmark 	&	 \([0,\frac{\sqrt{2}}{(1-\gamma)d_{\min}}]\) 	&	 \([0,\frac{\sqrt{2}}{(1-\gamma)d_{\min}}]\) 	\\
HKDD~\cite{de2008hierarchical}	&	 \cmark	&	 \cmark	&	 \cmark	&	 \cmark	&	 \cmark	&	 \cmark	&	 \cmark 	&	 \cmark	&	 \([0,\sqrt{2}]\)	&	 \([0,\sqrt{2}]\)	\\
\bottomrule
\end{tabular}
\end{footnotesize}
}

{\textsuperscript{*}\scriptsize{$D$ is the diameter of $\G$ and $d_{\min}:=\min_{v_i\in \V}{d_i}$.}\hfill}
\vspace{-1ex}
\end{table}

\subsection{Diffusion Distance as $\delta_{i,j}$}
The linchpin to realizing our foregoing idea for ameliorating MPNNs then lies in choosing an appropriate distance measure $\delta_{i,j}$ for edges ${(v_i,v_j)\in \EDG}$. 
Specifically, we propose to adopt the diffusion distances (VDD, PRDD, or HKDD) due to the following reasons.

Firstly, an ideal $\delta_{i,j}\ \forall{(v_i,v_j)\in \EDG}$ should vary in a finite range such that the node embeddings can be properly adjusted to reflect their nuance in graph topology. 
Compared to diffusion distances, classic measures like {\em shortest path distance} (SPD)~\cite{dijkstra1959note}, {\em resistance distance}~\cite{Klein1993}, {\em biharmonic distance}~\cite{lipman2010biharmonic}, are either invariant or unbounded (i.e., infinite) or undefined for node pairs that are edges in directed or undirected graphs, making them ill-suited. For instance, the SPDs of all node pairs that are adjacent (i.e., edges) in undirected graphs are 1.
Moreover, $\delta_{i,j}$ should be small if $(v_i,v_j)$ is a homophilic edge, otherwise large when it is a noisy/heterophilic link.
Intuitively, two nodes are more likely to be homophilic when they are highly connected and share common structural characteristics.
Thus, {\em Jaccard distance}~\cite{kosub2019note} is also problematic as it is defined merely by the common neighbors of two nodes, which falls short of considering edge weights and global paths pertaining to nodes.

\begin{theorem}\label{thm:robustness}
Let \( {\DM}^\prime \) and \( {\NAM}^\prime \) be the degree and normalized adjacency matrices of $\G$ after a small perturbation s.t. \( \| \NAM - {\NAM}^\prime \|_F \leq \epsilon_a \) and \( \max(| d_i - {d}^\prime_i |, | d_j - {d}^\prime_j |) \leq \epsilon_d \).
Then, \(|\Delta(v_i,v_j) - \widetilde{\Delta}(v_i,v_j)| \leq \frac{c\epsilon_d}{\tilde{d}^{3/2}} + \frac{2C \epsilon_a}{\tilde{d}^{1/2}}\).
$\widetilde{\Delta}(v_i,v_j)$ is the diffusion distance of $v_i,v_j$ on the perturbated graph, \( \tilde{d} = \min(d_i, d_j, {d}^\prime_i, {d}^\prime_j) \), $c=1$ for VDD and HKDD, and $c=\frac{1}{1 - \gamma}$ for PRDD, and $C$ is a constant.
\end{theorem}

In comparison, diffusion distances are defined on the holistic connectivity patterns summarized using random walk distributions, not only yielding reasonable ranges stated in Lemma~\ref{lem:dd-range} but also capturing global structures.
Theorem~\ref{thm:robustness} further indicates their stability under small graph perturbations, connoting that \(\delta_{i,j}\) will not be significantly affected by noise and minor changes in graph structures, especially those far from the vicinity of $v_i$ and $v_j$.
In Table~\ref{tbl:distances-concise}, we summarize the merits of diffusion distances and deficiencies of other graph-theoretic distance measures in terms of ten important topological and metric properties, such as preservation of local/global structures, capturing of structural similarities, support of symmetry, finiteness, and valid ranges.
Detailed theoretical analyses and experimental comparisons are deferred to Appendix~\ref{sec:distance-properties} and Section~\ref{eq:ablation}, respectively.

\subsection{Diffusion Distance-based Message Passing Scheme}\label{sec:dist-MPNNs}

Plugging the distance-based loss \(\mathcal{S}\) (Eq.~\eqref{eq:distance-loss}) and orthogonality regularization term \(\mathcal{C}\) (Eq.~\eqref{eq:decorrelation}) into the optimization objective of MPNNs leads to
\begin{align}\label{eq:obj}
\min_{\HM}(1 - \alpha - \beta) \cdot \mathcal{S} + \beta \cdot \mathcal{C} + \alpha \cdot \|\HM-\HM^{(0)}\|^2_F,
\end{align}
where $\alpha$ and $\beta$ are the coefficients for the fitting term \(\|\HM-\HM^{(0)}\|^2_F\) and orthogonality regularization term \(\mathcal{C}\), respectively. In $\mathcal{S}$, we set $\delta_{i,j}$ to \(\eta\cdot\Delta(v_i,v_j)\), where $\eta$ is a scaling factor and \(\Delta(v_i,v_j)\) can be any of the diffusion distances \(\Delta_{\text{vanilla}}(v_i, v_j)\), \(\Delta_{\text{HK}}(v_i, v_j)\), and \(\Delta_{\text{PR}}(v_i, v_j)\).

By taking the derivative of Eq. (\ref{eq:obj}) w.r.t. $\HM_i$ and setting it to zero, we can obtain the new message passing (or aggregation) operations for updating the node representation $\HM_i$ of the node $v_i$ at layer $k+1$ in \alg as follows (see Appendix~\ref{sec:ours-details} for detailed steps):
\begin{align}\label{eq:propagation-rule}
\HM_i^{(k+1)} =&  (1 - \alpha - \beta) \sum_{v_j \in \N(v_i)} \frac{{\HM}_j^{(k)}}{\sqrt{d_i d_j}}  \nonumber \\
 + &\eta (1 - \alpha - \beta) \sum_{v_j \in \N(v_i)} \frac{\Delta(v_i,v_j)\cdot \left( \frac{\HM_i^{(k)}}{{d_i}} - \frac{{\HM}_j^{(k)}}{\sqrt{d_i d_j} }\right)}{\left\| \frac{\HM_i^{(k)}}{\sqrt{d_i}} - \frac{\HM_j^{(k)}}{\sqrt{d_j}} \right\|_2} \nonumber\\
 + &\beta\cdot (\HM^{(k)}{\HM^{(k) \top}}\HM^{(k)})_i + \alpha \HM^{(0)}_i.
\end{align}
Specifically, the first term is the same as in previous MPNNs, which aggregates features from the neighborhood, while our second term assimilates the feature difference of the node $v_i$ and each of its neighbors $v_j$ as per their diffusion and embedding distances. The latter essentially pushes adjacent nodes away from each other when their diffusion distance is too large or the embedding distance is too small, thereby alleviating the over-smoothing, particularly in heterophilic graphs.
In the third term, entries in \(\HM^{(k)}{\HM^{(k) \top}}\) can be exceedingly large or small, which tend to overwhelm or be overwhelmed by other terms and engender vanishing/exploding gradients.
As a remedy, we replace it by \(\beta\cdot (\widehat{\HM}^{(k)}{\widehat{\HM}^{(k) \top}}\HM^{(k)})\) with a manageable range, where \(\widehat{\HM}^{(k)}_{\cdot,\ell} = {{\HM}^{(k)}_{\cdot,\ell}}/{\|{\HM}^{(k)}_{\cdot,\ell}\|_2}\).
Lastly, the term \(\alpha \HM^{(0)}\) acts as a residual connection~\cite{he2016deep}.

\subsection{Distance Computation via Spectral Decomposition}\label{sec:distance-computation}

According to Eq.~\eqref{eq:DD} and~\eqref{eq:hkdd}, the exact computations of \(\Delta_{\text{vanilla}}(v_i, v_j)\), \(\Delta_{\text{HK}}(v_i, v_j)\), or \(\Delta_{\text{PR}}(v_i, v_j)\ \forall{(v_i,v_j)\in \EDG}\) require materializing the $n\times n$ dense matrix $\PM^t$ and even involve summing up an infinite series, which is prohibitive or infeasible for medium-sized/large graphs.
Thus, we resort to an approximate solution by their spectral properties stated in the theorem below.
\begin{theorem}\label{thm:dd-computation}
Let $\UM\LABM\UM^{\top}$ be the eigendecomposition of $\NAM$.
Then, \[
\Delta_{\textnormal{vanilla}}(v_i, v_j) = \left\|\frac{\UM_i\LABM^t}{\sqrt{d_i}}-\frac{\UM_j\LABM^t}{\sqrt{d_j}}\right\|_2,
\] 
\[\Delta_{\textnormal{PR}}(v_i,v_j) = \left\|\frac{\UM_i}{\sqrt{d_i}(1-\gamma\LABM)}\allowbreak-\allowbreak\frac{\UM_j}{\sqrt{d_j}(1-\gamma\LABM)} \right\|_2.\]
If $\UM\LABM\UM^{\top}$ is the eigendecomposition of $\NLM$,
\[\Delta_{\textnormal{HK}}(v_i,v_j) = \left\|\frac{\UM_i e^{-\gamma\LABM}}{\sqrt{d_i}}-\frac{\UM_j e^{-\gamma\LABM}}{\sqrt{d_j}}\right\|_2.\]
\end{theorem}

More concretely, the VDD, PRDD, and HKDD of each edge $(v_i,v_j)$ can be unified into the Euclidean distance $\Delta(v_i,v_j)$ of length-$n$ embeddings $\ZM_i$ and $\ZM_j$:
\begin{equation}
\Delta(v_i,v_j) = \|\ZM_i-\ZM_j\|_2,\ \ZM_i = \frac{\UM_i}{\sqrt{d_i}}\cdot f(\boldsymbol{\Lambda}),
\end{equation}
where $f(\cdot)$ is an elementwise function applied over the eigenvalues on the diagonals of \(\boldsymbol{\Lambda}\), i.e., 
\begin{equation*}
f(\boldsymbol{\Lambda}) = \begin{cases}
\boldsymbol{\Lambda}^t & \text{VDD},\\
\frac{1}{1-\gamma\boldsymbol{\Lambda}} & \text{PRDD},\\
e^{-\gamma\boldsymbol{\Lambda}} & \text{HKDD}.
\end{cases}
\end{equation*}

As such, instead of calculating the exact $\ZM$ via a full eigendecomposition of $\NAM$ or $\NLM$ that needs a computational cost of $O(n^3)$ and $O(n^2)$ space consumption, we employ the {\em truncated eigendecomposition} to compute the \(\kappa\)-largest ($\kappa\ll n$) eigenvalues $\boldsymbol{\Lambda}^\prime$ and the corresponding eigenvectors $\UM^\prime$. Based thereon, we can construct the approximate distance embeddings $\ZM^\prime =  \DM^{-\frac{1}{2}}{\UM^\prime}\cdot f(\boldsymbol{\Lambda}^\prime) \in \mathbb{R}^{n\times \kappa}$ and estimate the diffusion distance \(\Delta(v_i,v_j)\) by
\begin{equation}\label{eq:approx-dist}
\Delta^\prime(v_i,v_j) = \|\ZM^\prime_i-\ZM^\prime_j\|_2. 
\end{equation}

\begin{theorem}\label{prop:dd-unified}
$\forall (v_i,v_j)\in \EDG$, \(\Delta^2(v_i, v_j) - \frac{2\epsilon\cdot (1 - \mathbb{1}_{i=j})}{\min(d_i,d_j)} \leq \Delta^{\prime2}(v_i, v_j) \allowbreak\leq \Delta^2(v_i, v_j)\),
where $\epsilon=\lambda_{\kappa}^{2t}$, ${1}/{(1-\gamma\lambda_{\kappa})^2}$, and $e^{-2\gamma\lambda_{\kappa}}$ when the VDD, PRDD, and HKDD are adopted, respectively. $\lambda_k$ is the $k$-th largest eigenvalue of $\NAM$ in absolute and algebraic values for VDD and PRDD, respectively, and the $k$-th smallest algebraic eigenvalue of $\NLM$ for HKDD.
\end{theorem}

Theorem~\ref{prop:dd-unified} establishes the approximation guarantee for our estimated diffusion distance defined in Eq.~\eqref{eq:approx-dist}.

\subsection{Complexity Analysis}\label{sec:time-complexity}
In light of the sparsity and symmetry of $\G$, the $\kappa$-eigendecomposition of $\NAM$ and $\NLM$ can be done in $O(\kappa m\cdot \ell)$ time, where $\ell$ is the number of iterations (typically $7$). This procedure is efficient in practice due to randomized solvers~\cite{halko2011finding} and highly-optimized parallel libraries (LAPACK and BLAS). 
Given the partial eigenvectors, the diffusion distances of all edges can be easily pre-calculated in \( O((n+m)\kappa) \) time.
During training, the model consists of three main phases: initial embedding, propagation, and classification. In the initial embedding phase, a linear layer is used to reduce dimensions, with a forward time complexity of $O(nd^{(0)}d)$. The propagation phase includes aggregating information from neighboring nodes, which costs $O(md)$, and performing linear transformations in the convolution module, resulting in a per-layer complexity of $O(nd^2)$. In the classification phase, another linear layer is used, with a forward time complexity of $O(nd|\Y|)$.
We use $K$ to denote the number of the convolution layers. The overall training complexity per epoch is $O(nd(d^{(0)} +dK + |\Y|) + mdK)$.

\begin{table}[t]
\centering
\caption{Dataset statistics.}\label{tbl:data-statistics}
\vspace{-2ex}
\renewcommand{\arraystretch}{0.8}
\begin{small}    
\begin{tabular}{lrrrrrcl}
\toprule
{\bf Dataset}        & $n$ & $m$ & $d^{(0)}$ & $|\Y|$ & $h(\G)$ & \multicolumn{1}{c}{{\bf Type}} \\
\midrule
\textit{Cora}           & 2,708   & 10,556        & 1,433      & 7         & 0.81   & \multirow{8}{*}{Homophilic} \\
\textit{CiteSeer}       & 3,327   & 9,104         & 3,703      & 6         & 0.74   & \\
\textit{PubMed}         & 19,717  & 88,648        & 500        & 3         & 0.80   & \\
\textit{CoraFull  }     & 19,793  & 126,842       & 8,710      & 70        & 0.57   & \\
\textit{CS}             & 18,333  & 163,788       & 6,805      & 15        & 0.81   & \\
\textit{Physics}        & 34,493  & 495,924       & 8,415      & 5         & 0.93   & \\
\textit{WikiCS}         & 11,701  & 431,726       & 300        & 10        & 0.65   & \\
\midrule
\textit{Cornell}        & 183     & 557           & 1,703      & 5         & 0.13   & \multirow{4}{*}{Heterophilic} \\
\textit{Texas}          & 183     & 574           & 1,703      & 5         & 0.09   & \\
\textit{Wisconsin}      & 251     & 916           & 1,703      & 5         & 0.19   & \\
\textit{Chameleon}      & 2,277   & 62,792        & 2,325      & 5         & 0.23   & \\
\bottomrule
\end{tabular}
\end{small}
\vspace{-1ex}
\end{table}
\begin{table*}[t]
\centering
\renewcommand{\arraystretch}{0.9}
\caption{Node classification accuracy with the standard deviation. 
The best baseline is \underline{underlined}.
}\vspace{-2ex}
\begin{small}
\addtolength{\tabcolsep}{-0.25em}
\begin{tabular}{c|c | c | c| c| c | c| c| c| c | c| c}
\toprule
& {\bf \em Cora} & {\bf \em CiteSeer} & {\bf \em PubMed} & {\bf \em CoraFull} & {\bf \em CS} & {\bf \em Physics} & {\bf \em Cornell} & {\bf \em Texas} & {\bf \em Wisconsin} & {\bf \em Chameleon} & {\bf \em WikiCS} \\
\midrule
\texttt{Linear}	&	{76.01{\scriptsize$\pm$1.55}}	&	{71.31{\scriptsize$\pm$1.74}}	&	{87.08{\scriptsize$\pm$0.64}}	&	{60.55{\scriptsize$\pm$0.80}}	&	{94.48{\scriptsize$\pm$0.31}}	&	{95.93{\scriptsize$\pm$0.15}}	&	{78.65{\scriptsize$\pm$2.82}}	&	{81.35{\scriptsize$\pm$4.90}}	&	{86.20{\scriptsize$\pm$3.63}}	&	{50.77{\scriptsize$\pm$2.07}}	&	{78.76{\scriptsize$\pm$0.55}} \\
\texttt{MLP}	&	{75.87{\scriptsize$\pm$1.40}}	&	{73.20{\scriptsize$\pm$1.07}}	&	{88.22{\scriptsize$\pm$0.40}}	&	{61.54{\scriptsize$\pm$0.52}}	&	{94.87{\scriptsize$\pm$0.28}}	&	{96.03{\scriptsize$\pm$0.25}}	&	{74.05{\scriptsize$\pm$4.55}}	&	{85.14{\scriptsize$\pm$6.97}}	&	{88.40{\scriptsize$\pm$2.33}}	&	{50.57{\scriptsize$\pm$1.84}}	&	{80.09{\scriptsize$\pm$0.64}} \\
\texttt{GCN}	&	{88.25{\scriptsize$\pm$1.09}}	&	{77.05{\scriptsize$\pm$1.14}}	&	{89.05{\scriptsize$\pm$0.46}}	&	{71.83{\scriptsize$\pm$0.75}}	&	{93.76{\scriptsize$\pm$0.42}}	&	{96.45{\scriptsize$\pm$0.22}}	&	{50.27{\scriptsize$\pm$7.66}}	&	{59.73{\scriptsize$\pm$9.40}}	&	{57.00{\scriptsize$\pm$4.75}}	&	{68.88{\scriptsize$\pm$2.07}}	&	{83.89{\scriptsize$\pm$0.79}} \\
\texttt{PointNet}	&	{84.39{\scriptsize$\pm$1.82}}	&	{73.37{\scriptsize$\pm$1.37}}	&	{88.97{\scriptsize$\pm$0.51}}	&	{62.84{\scriptsize$\pm$1.27}}	&	{93.34{\scriptsize$\pm$0.44}}	&	{96.34{\scriptsize$\pm$0.22}}	&	{68.11{\scriptsize$\pm$8.44}}	&	{77.57{\scriptsize$\pm$7.46}}	&	{81.80{\scriptsize$\pm$5.02}}	&	{64.88{\scriptsize$\pm$2.30}}	&	{84.26{\scriptsize$\pm$0.76}} \\
\texttt{GAT}	&	{88.51{\scriptsize$\pm$0.80}}	&	{76.00{\scriptsize$\pm$1.25}}	&	{87.86{\scriptsize$\pm$0.70}}	&	{70.64{\scriptsize$\pm$0.87}}	&	{93.24{\scriptsize$\pm$0.37}}	&	{96.45{\scriptsize$\pm$0.24}}	&	{50.27{\scriptsize$\pm$6.96}}	&	{59.73{\scriptsize$\pm$3.07}}	&	{55.20{\scriptsize$\pm$6.76}}	&	{66.53{\scriptsize$\pm$2.77}}	&	{83.66{\scriptsize$\pm$0.86}} \\
\texttt{SGC}	&	{88.34{\scriptsize$\pm$1.41}}	&	{77.22{\scriptsize$\pm$1.42}}	&	{87.59{\scriptsize$\pm$0.54}}	&	{71.86{\scriptsize$\pm$0.70}}	&	{94.04{\scriptsize$\pm$0.35}}	&	{96.47{\scriptsize$\pm$0.18}}	&	{53.51{\scriptsize$\pm$4.32}}	&	{56.22{\scriptsize$\pm$6.37}}	&	{58.20{\scriptsize$\pm$4.04}}	&	{67.23{\scriptsize$\pm$2.27}}	&	{83.49{\scriptsize$\pm$0.82}} \\
\texttt{GIN}	&	{85.77{\scriptsize$\pm$1.19}}	&	{72.93{\scriptsize$\pm$1.17}}	&	{87.24{\scriptsize$\pm$0.42}}	&	{67.00{\scriptsize$\pm$0.57}}	&	{91.74{\scriptsize$\pm$0.45}}	&	{95.39{\scriptsize$\pm$0.27}}	&	{48.65{\scriptsize$\pm$8.72}}	&	{62.97{\scriptsize$\pm$4.02}}	&	{54.80{\scriptsize$\pm$8.54}}	&	{45.58{\scriptsize$\pm$13.72}}	&	{54.85{\scriptsize$\pm$21.81}} \\
\texttt{APPNP}	&	{89.24{\scriptsize$\pm$1.49}}	&	{76.80{\scriptsize$\pm$1.10}}	&	{89.75{\scriptsize$\pm$0.58}}	&	{72.12{\scriptsize$\pm$0.67}}	&	\underline{95.90{\scriptsize$\pm$0.21}}	&	\underline{97.16{\scriptsize$\pm$0.18}}	&	{76.49{\scriptsize$\pm$8.38}}	&	{82.70{\scriptsize$\pm$3.01}}	&	{86.00{\scriptsize$\pm$3.10}}	&	{67.19{\scriptsize$\pm$2.40}}	&	{85.08{\scriptsize$\pm$0.51}} \\
\texttt{GCNII}	&	{88.39{\scriptsize$\pm$1.30}}	&	{76.36{\scriptsize$\pm$0.82}}	&	{90.15{\scriptsize$\pm$0.63}}	&	{71.03{\scriptsize$\pm$0.62}}	&	{95.87{\scriptsize$\pm$0.24}}	&	{97.07{\scriptsize$\pm$0.22}}	&	{75.14{\scriptsize$\pm$8.00}}	&	{87.57{\scriptsize$\pm$4.71}}	&	{88.00{\scriptsize$\pm$4.29}}	&	{68.02{\scriptsize$\pm$1.20}}	&	\underline{85.33{\scriptsize$\pm$0.58}} \\
\texttt{LINKX}	&	{82.14{\scriptsize$\pm$2.08}}	&	{71.37{\scriptsize$\pm$1.97}}	&	{86.66{\scriptsize$\pm$0.64}}	&	{63.50{\scriptsize$\pm$0.48}}	&	{94.96{\scriptsize$\pm$0.25}}	&	{96.88{\scriptsize$\pm$0.17}}	&	{69.46{\scriptsize$\pm$6.84}}	&	{82.43{\scriptsize$\pm$6.07}}	&	{83.00{\scriptsize$\pm$4.12}}	&	{67.43{\scriptsize$\pm$1.61}}	&	{83.83{\scriptsize$\pm$0.55}} \\
\texttt{pGNN}	&	{88.84{\scriptsize$\pm$1.37}}	&	{77.28{\scriptsize$\pm$0.86}}	&	{89.80{\scriptsize$\pm$0.38}}	&	{72.19{\scriptsize$\pm$0.63}}	&	{95.88{\scriptsize$\pm$0.22}}	&	{96.92{\scriptsize$\pm$0.11}}	&	{74.05{\scriptsize$\pm$9.22}}	&	{84.86{\scriptsize$\pm$4.86}}	&	{85.20{\scriptsize$\pm$4.40}}	&	{69.43{\scriptsize$\pm$1.29}}	&	{84.55{\scriptsize$\pm$0.77}} \\ 
\texttt{ACM-GCN}	& \underline{89.63{\scriptsize$\pm$0.26}} & {75.38{\scriptsize$\pm$0.49}} & {90.18{\scriptsize$\pm$0.20}} & {71.94{\scriptsize$\pm$0.40}} & {95.28{\scriptsize$\pm$0.13}} & {97.12{\scriptsize$\pm$0.09}} & {76.76{\scriptsize$\pm$1.79}} & {81.08{\scriptsize$\pm$1.71}} & \underline{91.60{\scriptsize$\pm$1.50}} & \underline{72.64{\scriptsize$\pm$1.01}} & {83.13{\scriptsize$\pm$0.54}} \\ 
\texttt{GloGNN}	& {86.93{\scriptsize$\pm$0.82}} & {77.52{\scriptsize$\pm$0.03}} & {88.93{\scriptsize$\pm$0.38}} & {63.42{\scriptsize$\pm$0.47}} & {94.17{\scriptsize$\pm$0.41}} & {95.99{\scriptsize$\pm$0.23}} & {74.59{\scriptsize$\pm$6.77}} & {82.30{\scriptsize$\pm$4.45}} & {86.75{\scriptsize$\pm$4.78}} & {69.76{\scriptsize$\pm$1.57}} & {80.56{\scriptsize$\pm$0.75}} \\ 
\texttt{MGNN}	&	{88.67{\scriptsize$\pm$0.96}}	&	77.98{\scriptsize$\pm$1.50}	&	\underline{90.39{\scriptsize$\pm$0.48}}	&	\underline{72.28{\scriptsize$\pm$0.74}}	&	{95.89{\scriptsize$\pm$0.13}}	&	{97.00{\scriptsize$\pm$0.22}}	&	{77.84{\scriptsize$\pm$6.92}}	&	\underline{88.65{\scriptsize$\pm$3.38}}	&	{89.00{\scriptsize$\pm$3.61}}	&	{72.11{\scriptsize$\pm$1.58}}	&	{84.92{\scriptsize$\pm$0.56}} \\ 
\texttt{UniFilter}	&	{86.22{\scriptsize$\pm$1.77}}	&	\underline{78.61{\scriptsize$\pm$1.29}}	&	{87.36{\scriptsize$\pm$0.05}}	&	{71.20{\scriptsize$\pm$0.43}}	&	{95.56{\scriptsize$\pm$0.20}}	&	{97.16{\scriptsize$\pm$0.24}}	&	\underline{79.57{\scriptsize$\pm$6.67}}	&	{86.23{\scriptsize$\pm$3.13}}	&	{86.00{\scriptsize$\pm$3.86}}	&	{67.64{\scriptsize$\pm$2.38}}	&	{80.06{\scriptsize$\pm$0.34}} \\ \midrule
\algp       	&	{92.10{\scriptsize$\pm$0.34}}	&	{80.86{\scriptsize$\pm$0.61}}	&	{90.96{\scriptsize$\pm$0.12}}	&	{73.65{\scriptsize$\pm$0.12}}	&	{96.22{\scriptsize$\pm$0.17}}	&	{97.45{\scriptsize$\pm$0.08}}	&	{85.14{\scriptsize$\pm$2.40}}	&	{96.76{\scriptsize$\pm$1.08}}	& {92.80{\scriptsize$\pm$1.04}}   &	{75.23{\scriptsize$\pm$0.48}}	&	{86.45{\scriptsize$\pm$0.29}}	\\
Improv.       	&	{+2.47}	&	{+2.25}	&	{+0.57}	&	{+1.37}	&	{+0.32}	&	{+0.29}	&	{+5.57}	&	{+8.11}	& {+1.20}   &	{+2.59}	&	{+1.12}	\\
\algv      	&	{92.69{\scriptsize$\pm$0.33}}	&	{81.01{\scriptsize$\pm$0.69}}	&	{90.98{\scriptsize$\pm$0.15}}	&	{73.68{\scriptsize$\pm$0.13}}	&	{96.10{\scriptsize$\pm$0.12}}	&	{97.46{\scriptsize$\pm$0.15}}	&	{86.22{\scriptsize$\pm$3.51}}	&	{93.24{\scriptsize$\pm$1.81}}	& {92.40{\scriptsize$\pm$0.80}}   &	{75.10{\scriptsize$\pm$0.30}}	&	{86.67{\scriptsize$\pm$0.10}}	\\
Improv.       	&	{+3.06}	&	{+2.40}	&	{+0.59}	&	{+1.40}	&	{+0.20}	&	{+0.30}	&	{+6.65}	&	{+4.59}	& {+0.80}   &	{+2.46}	&	{+1.34}	\\
\algh        	&	{92.16{\scriptsize$\pm$0.32}}	&	{80.96{\scriptsize$\pm$0.33}}	&	{91.05{\scriptsize$\pm$0.19}}	&	{73.96{\scriptsize$\pm$0.20}}	&	{96.08{\scriptsize$\pm$0.09}}	&	{97.45{\scriptsize$\pm$0.07}}	&	{85.41{\scriptsize$\pm$1.79}}	&	{97.03{\scriptsize$\pm$0.81}}	& {92.80{\scriptsize$\pm$1.34}}   &	{75.01{\scriptsize$\pm$0.49}}	&	{86.61{\scriptsize$\pm$0.09}}	\\
Improv.	&	{+2.53}	&	{+2.35}	&	{+0.66}	&	{+1.68}	&	{+0.18}	&	{+0.29}	&	{+5.84}	&	{+8.38}	&	{+1.20}	&	{+2.37}	&	{+1.28}	\\

\bottomrule
\end{tabular}
\end{small}
\label{tbl:node-classification}
\end{table*}

\begin{table*}[t]
\centering
\renewcommand{\arraystretch}{0.9}
\caption{Ablation study on orthogonal regularization.}\vspace{-2ex}
\begin{small}
\addtolength{\tabcolsep}{-0.25em}
\begin{tabular}{c|c | c | c| c| c | c| c| c| c | c| c}
\toprule
& {\bf \em Cora} & {\bf \em CiteSeer} & {\bf \em PubMed} & {\bf \em CoraFull} & {\bf \em CS} & {\bf \em Physics} & {\bf \em Cornell} & {\bf \em Texas} & {\bf \em Wisconsin} & {\bf \em Chameleon} & {\bf \em WikiCS} \\
\midrule
\algv      	&	{92.69{\scriptsize$\pm$0.33}}	&	{81.01{\scriptsize$\pm$0.69}}	&	{90.98{\scriptsize$\pm$0.15}}	&	{73.68{\scriptsize$\pm$0.13}}	&	{96.10{\scriptsize$\pm$0.12}}	&	{97.46{\scriptsize$\pm$0.15}}	&	{86.22{\scriptsize$\pm$3.51}}	&	{93.24{\scriptsize$\pm$1.81}}	& {92.40{\scriptsize$\pm$0.80}}   &	{75.10{\scriptsize$\pm$0.30}}	&	{86.67{\scriptsize$\pm$0.10}}	\\
Vanilla (w/o Orth.)        	&	{92.51{\scriptsize$\pm$0.24}}	&	{80.45{\scriptsize$\pm$0.20}}	&	{90.77{\scriptsize$\pm$0.10}}	&	{73.43{\scriptsize$\pm$0.10}}	&	{96.06{\scriptsize$\pm$0.09}}	&	{97.44{\scriptsize$\pm$0.09}}	&	{84.32{\scriptsize$\pm$2.65}}	&	{92.16{\scriptsize$\pm$0.81}}	& {90.60{\scriptsize$\pm$2.69}}   &	{74.86{\scriptsize$\pm$0.85}}	&	{86.19{\scriptsize$\pm$0.21}}	\\ \midrule
\algp       	&	{92.10{\scriptsize$\pm$0.34}}	&	{80.86{\scriptsize$\pm$0.61}}	&	{90.96{\scriptsize$\pm$0.12}}	&	{73.65{\scriptsize$\pm$0.12}}	&	{96.22{\scriptsize$\pm$0.17}}	&	{97.45{\scriptsize$\pm$0.08}}	&	{85.14{\scriptsize$\pm$2.40}}	&	{96.76{\scriptsize$\pm$1.08}}	& {92.80{\scriptsize$\pm$1.04}}   &	{75.23{\scriptsize$\pm$0.48}}	&	{86.45{\scriptsize$\pm$0.29}}	\\
PRDD (w/o Orth.)        	&	{91.79{\scriptsize$\pm$0.36}}	&	{80.42{\scriptsize$\pm$0.30}}	&	{90.81{\scriptsize$\pm$0.11}}	&	{73.60{\scriptsize$\pm$0.13}}	&	{96.14{\scriptsize$\pm$0.16}}	&	{97.39{\scriptsize$\pm$0.10}}	&	{85.95{\scriptsize$\pm$2.65}}	&	{92.43{\scriptsize$\pm$1.08}}	& {92.40{\scriptsize$\pm$1.96}}   &	{74.22{\scriptsize$\pm$0.38}}	&	{86.05{\scriptsize$\pm$0.36}}	\\ \midrule
\algh        	&	{92.16{\scriptsize$\pm$0.32}}	&	{80.96{\scriptsize$\pm$0.33}}	&	{91.05{\scriptsize$\pm$0.19}}	&	{73.96{\scriptsize$\pm$0.20}}	&	{96.08{\scriptsize$\pm$0.09}}	&	{97.45{\scriptsize$\pm$0.07}}	&	{85.41{\scriptsize$\pm$1.79}}	&	{97.03{\scriptsize$\pm$0.81}}	& {92.80{\scriptsize$\pm$1.34}}   &	{75.01{\scriptsize$\pm$0.49}}	&	{86.61{\scriptsize$\pm$0.09}}	\\
HKDD (w/o Orth.)        	&	{91.99{\scriptsize$\pm$0.41}}	&	{80.11{\scriptsize$\pm$0.53}}	&	{90.85{\scriptsize$\pm$0.08}}	&	{73.87{\scriptsize$\pm$0.14}}	&	{95.86{\scriptsize$\pm$0.08}}	&	{97.43{\scriptsize$\pm$0.06}}	&	{81.08{\scriptsize$\pm$2.09}}	&	{91.89{\scriptsize$\pm$1.21}}	& {92.00{\scriptsize$\pm$1.55}}   &	{74.73{\scriptsize$\pm$0.39}}	&	{86.13{\scriptsize$\pm$0.21}}	\\
\bottomrule
\end{tabular}
\end{small}
\label{tbl:ablation}
\end{table*}

\begin{table*}[t]
\centering
\renewcommand{\arraystretch}{0.9}
\caption{Ablation study on distance measures. (``-'' denotes incomputable)}\vspace{-2ex}
\begin{small}
\addtolength{\tabcolsep}{-0.25em}
\begin{tabular}{c|c | c | c| c| c | c| c| c| c | c| c}
\toprule
& {\bf \em Cora} & {\bf \em CiteSeer} & {\bf \em PubMed} & {\bf \em CoraFull} & {\bf \em CS} & {\bf \em Physics} & {\bf \em Cornell} & {\bf \em Texas} & {\bf \em Wisconsin} & {\bf \em Chameleon} & {\bf \em WikiCS} \\
\midrule
w/o distance        	&	{91.88{\scriptsize$\pm$0.37}}	&	{79.89{\scriptsize$\pm$0.50}}	&	{90.52{\scriptsize$\pm$0.14}}	&	{73.30{\scriptsize$\pm$0.27}}	&	{96.02{\scriptsize$\pm$0.12}}	&	{97.46{\scriptsize$\pm$0.06}}	&	{83.24{\scriptsize$\pm$2.91}}	&	{91.89{\scriptsize$\pm$0.00}}	& {89.80{\scriptsize$\pm$1.66}}   &	{73.45{\scriptsize$\pm$0.43}}	&	{86.08{\scriptsize$\pm$0.26}}	\\ 
SPD      	&	{90.61{\scriptsize$\pm$0.49}}	&	{78.54{\scriptsize$\pm$0.26}}	&	{90.56{\scriptsize$\pm$0.21}}	&	{73.60{\scriptsize$\pm$0.19}}	&	{95.72{\scriptsize$\pm$0.06}}	&	{{97.41\scriptsize$\pm$0.09}}	&	{84.59{\scriptsize$\pm$2.11}}	&	{91.89{\scriptsize$\pm$1.71}}	&	{92.20{\scriptsize$\pm$1.20}}	&	{74.18{\scriptsize$\pm$0.50}}	&	{86.06{\scriptsize$\pm$0.25}}	\\ 
Jaccard       	&	{91.77{\scriptsize$\pm$0.19}}	&	{80.98{\scriptsize$\pm$0.72}}	&	{90.49{\scriptsize$\pm$0.19}}	&	{73.53{\scriptsize$\pm$0.12}}	&	{95.66{\scriptsize$\pm$0.04}}	&	{97.41{\scriptsize$\pm$0.10}}	&	{84.59{\scriptsize$\pm$2.72}}	&	{92.16{\scriptsize$\pm$1.46}}	&	{92.20{\scriptsize$\pm$1.08}}	&	{73.89{\scriptsize$\pm$0.76}}	&	{86.26{\scriptsize$\pm$0.21}}	\\ 
Resistance       	&	{91.83{\scriptsize$\pm$0.23}}	&	{80.45{\scriptsize$\pm$0.40}}	&	{81.15{\scriptsize$\pm$0.39}}	&	{73.60{\scriptsize$\pm$0.13}}	&	{95.98{\scriptsize$\pm$0.14}}	&	{-}	&	{84.22{\scriptsize$\pm$2.42}}	&	{93.14{\scriptsize$\pm$2.02}}	&	{92.00{\scriptsize$\pm$1.55}}	&	{73.89{\scriptsize$\pm$0.67}}	&	{86.06{\scriptsize$\pm$0.35}}	\\ 
Biharmonic       	&	{91.77{\scriptsize$\pm$0.56}}	&	{80.23{\scriptsize$\pm$0.35}}	&	{82.72{\scriptsize$\pm$0.39}}	&	{73.64{\scriptsize$\pm$0.15}}	&	{95.88{\scriptsize$\pm$0.19}}	&	{-}	&	{84.59{\scriptsize$\pm$1.73}}	&	{91.89{\scriptsize$\pm$0.00}}	&	{92.40{\scriptsize$\pm$1.50}}	&	{74.18{\scriptsize$\pm$0.61}}	&	{84.57{\scriptsize$\pm$0.33}}	\\ \midrule
PRDD       	&	{92.10{\scriptsize$\pm$0.34}}	&	{80.86{\scriptsize$\pm$0.61}}	&	{90.96{\scriptsize$\pm$0.12}}	&	{73.65{\scriptsize$\pm$0.12}}	&	{96.22{\scriptsize$\pm$0.17}}	&	{97.45{\scriptsize$\pm$0.08}}	&	{85.14{\scriptsize$\pm$2.40}}	&	{96.76{\scriptsize$\pm$1.08}}	& {92.80{\scriptsize$\pm$1.04}}   &	{75.23{\scriptsize$\pm$0.48}}	&	{86.45{\scriptsize$\pm$0.29}}	\\
VDD      	&	{92.69{\scriptsize$\pm$0.33}}	&	{81.01{\scriptsize$\pm$0.69}}	&	{90.98{\scriptsize$\pm$0.15}}	&	{73.68{\scriptsize$\pm$0.13}}	&	{96.10{\scriptsize$\pm$0.12}}	&	{97.46{\scriptsize$\pm$0.15}}	&	{86.22{\scriptsize$\pm$3.51}}	&	{93.24{\scriptsize$\pm$1.81}}	& {92.40{\scriptsize$\pm$0.80}}   &	{75.10{\scriptsize$\pm$0.30}}	&	{86.67{\scriptsize$\pm$0.10}}	\\
HKDD        	&	{92.16{\scriptsize$\pm$0.32}}	&	{80.96{\scriptsize$\pm$0.33}}	&	{91.05{\scriptsize$\pm$0.19}}	&	{73.96{\scriptsize$\pm$0.20}}	&	{96.08{\scriptsize$\pm$0.09}}	&	{97.45{\scriptsize$\pm$0.07}}	&	{85.41{\scriptsize$\pm$1.79}}	&	{97.03{\scriptsize$\pm$0.81}}	& {92.80{\scriptsize$\pm$1.34}}   &	{75.01{\scriptsize$\pm$0.49}}	&	{86.61{\scriptsize$\pm$0.09}}	\\
\bottomrule
\end{tabular}
\end{small}
\label{tbl:ablation-dd}
\end{table*}

\section{Experiments}\label{sec:exp}
This section experimentally evaluates our proposed \alg model against 15 baselines on 11 real-world datasets for the node classification task.
All experiments are conducted on a Linux machine with an Intel Xeon(R) Gold 6438Y 2.00GHz CPU and an NVIDIA GeForce RTX 4090 GPU.
The source code is available at \url{https://github.com/HaoranZ99/DDSM}.

\stitle{Datasets and Baselines} 
We utilize a collection of well-known benchmark graph datasets, including the citation networks {\em Cora}, {\em CiteSeer}, and {\em PubMed} from~\cite{yang2016revisiting}; the {\em CoraFull} dataset from~\cite{bojchevski2018deep}; the coauthor networks {\em CS} and {\em Physics} from~\cite{shchur2018pitfalls}; the WebKB networks {\em Cornell}, {\em Texas}, and {\em Wisconsin} from~\cite{pei2020GeomGCN}; the {\em Chameleon} graph from Wikipedia~\cite{rozemberczki2021multi}; the {\em WikiCS} dataset from~\cite{mernyei2020wiki}. All datasets were acquired and preprocessed using the PyTorch Geometric Library~\cite{fey2019pyg}.
Table~\ref{tbl:data-statistics} provides an overview of the datasets’ fundamental characteristics, including the number of nodes, edges, features, and classes, as well as the homophily ratio, $h(\G)$ as defined in~\cite{zhu2020beyond}. A graph is referred to as homophilic if its homophily ratio exceeds 0.5; otherwise, it is classified as heterophilic.

As for baselines, we consider two non-graph models that are composed of a one-layer linear model (with bias) or a two-layer MLP, and 13 classic GNN models,
including \texttt{GCN}~\cite{kipf2017semisupervised}, \texttt{PointNet}~\cite{charles2017pointnet}, \texttt{GAT}~\cite{veličković2018graph}, \texttt{SGC}~\cite{wu2019simplifying}, \texttt{GIN}~\cite{xu2018how}, \texttt{APPNP}~\cite{gasteiger2019combining}, \texttt{GCNII}~\cite{chen2020simple}, \texttt{LINKX}~\cite{lim2021large}, \texttt{pGNN}~\cite{fu2022plaplacian}, \texttt{ACM-GCN}~\cite{luan2022revisiting}, \texttt{GloGNN}~\cite{li2022finding}, \texttt{MGNN}~\cite{cui2023mgnn}, and \texttt{UniFilter}~\cite{huang2024how}. 
For each baseline method, we tune its hyperparameters by following the search strategy and space recommended in the official implementation. We report the results from the best-performing configuration to ensure a fair comparison.
The details of the dataset and our hyperparameter settings are provided in Appendix~\ref{sec:exp-detail}. Additional experiments on efficiency analysis, scalability analysis, and ablation studies can be found in Appendix~\ref{sec:add-exp}.

\subsection{Node Classification Performance}
Table~\ref{tbl:node-classification} presents the results of the supervised node classification experiments. Our proposed methods (\algp, \algv, \algh) demonstrate superior performance across all datasets, achieving significant improvements over baseline models. In homophilic datasets such as \textit{Cora}, \textit{CiteSeer}, \textit{PubMed}, and \textit{CS}, our methods consistently outperform previous models by considerable margins. For instance, \algp achieves an accuracy of 92.10\% on Cora, surpassing the best model by 2.47\%. Similarly, \algv and \algh show steady improvements on \textit{CiteSeer} (2.40\% and 2.35\%, respectively) and \textit{PubMed} (0.59\% and 0.66\%). This robust performance is attributed to the effective integration of our key components: a diffusion distance term, which alleviates the global over-smoothing issue, and an orthogonal regularization term, which mitigates the feature correlation issue.
On heterophilic datasets including \textit{Cornell}, \textit{Texas}, \textit{Wisconsin}, and \textit{Chameleon}, the improvements are even more pronounced. 
The gains highlight the adaptability of our framework to diverse graph structures, leveraging our key components, especially the diffusion distance term, which alleviates the heterophilic over-smoothing issue.

\subsection{Ablation Study}\label{eq:ablation}
We further study the empirical effectiveness of 
our injected loss terms $\mathcal{S}$ (Eq.~\eqref{eq:distance-loss}) and $\mathcal{C}$ (Eq.~\eqref{eq:decorrelation}) in enhancing MPNNs.
First, we ablate the orthogonal regularization term $\beta\cdot\mathcal{C}$ in Eq.~\eqref{eq:obj} across all datasets. 
From Table~\ref{tbl:ablation}, it can be observed that when $\beta\cdot\mathcal{C}$ is removed, the ablated counterparts of \algv, \algp, and \algh all incur conspicuous performance degradation on almost all datasets.
For instance, on homophilic graphs {\em CiteSeer} and {\em WikiCS}, a performance improvement of at least $0.4\%$ can be attained after $\beta\cdot\mathcal{C}$ is added.
On heterophilic datasets {\em Cornell} and {\em Texas}, significant drops of $1.9\%$ and $1.08\%$ in classification accuracy can be observed for \algv, and $4.33\%$ and $5.14\%$ for \algh, respectively.

We next substitute other distance measures including SPD, Jaccard, resistance, and biharmonic distances, for the VDD, PRDD, or HKDD used as $\delta_{i,j}\ \forall{(v_i,v_j)\in \EDG}$ in \alg, respectively. We additionally include a baseline that sets all distance values to 0, dubbed as w/o distance.
As reported in Table~\ref{tbl:ablation-dd}, the node classification performance achieved by the variants of \alg w/o distances or other distances is all considerably inferior to \alg{}.
These advancements underscore the superiority of diffusion distances over other measures.

\subsection{Parameter Analysis}
This section analyzes the impact of coefficients $\alpha$, $\beta$, and $\eta$ in message functions (Eq.~\eqref{eq:propagation-rule}) of \alg and dimension $\kappa$ for distance approximation
on \textit{CiteSeer} and \textit{WikiCS}.
As shown in Fig.~\ref{fig:vary-alpha} and~\ref{fig:vary-alpha-wikics}, on both datasets, the node classification accuracies by \algv, \algp, and \algh first undergo an increase as $\alpha$ is increased from $0.1$ to 0.2 or $0.3$, followed by a consistent decline when $\alpha$ continues to grow. Hence, a weight of $0.2$ or $0.3$ for residual connection is favorable.
Fig.~\ref{fig:vary-beta} and~\ref{fig:vary-beta-wikics} depict the accuracy scores of \alg when varying \( \beta \) from $0.1$ to $0.9$.
It can be observed that increasing \( \beta \) to $0.3$ will significantly reduce accuracy on \textit{CiteSeer}.
On \textit{WikiCS}, \alg first experiences a performance decrease when $\beta$ is increased to $0.7$ or $0.8$, and then sees a notable improvement afterwards, manifesting the importance of feature decorrelation on this dataset.
In Fig.~\ref{fig:vary-eta} and~\ref{fig:vary-eta-wikics}, we display the accuracies when varying \( \eta \) from $0.1$ to $0.9$.
On \textit{CiteSeer}, there is an upward trend in the accuracy as \( \eta \) increases, indicating that larger values are generally beneficial. In contrast, on \textit{WikiCS}, the performance of \alg is relatively less sensitive to $\eta$.
We report the performance of \alg when varying the dimension \( \kappa \) from $8$ to $n$.
From Fig.~\ref{fig:vary-k} and~\ref{fig:vary-k-wikics}, we can observe that, on both datasets, a small \( \kappa \) of $32$, $64$, or $128$ is sufficient for approximating VDD, HKDD, and PRDD and learning predictive embeddings in \alg.
\begin{figure}[t]
\centering
\begin{small}
\begin{tikzpicture}
    \begin{customlegend}[legend columns=3,
        legend entries={PRDD, Vanilla, HKDD},
        legend style={at={(0.45,1.35)},anchor=north,draw=none,font=\small,column sep=0.1cm}]
    \addlegendimage{line width=0.3mm,mark size=3pt,mark=triangle, color=PageRank-color}
    \addlegendimage{line width=0.3mm,mark size=3pt,mark=diamond, color=Vanilla-color}
    \addlegendimage{line width=0.3mm,mark size=3pt,mark=o, color=HeatKernel-color}
    \end{customlegend}
\end{tikzpicture}
\\[-\lineskip]
\vspace{-4mm}
\subfloat[Varying $\alpha$ on {\em CiteSeer}]{
\begin{tikzpicture}[scale=1,every mark/.append style={mark size=2pt}]
    \begin{axis}[
        height=\columnwidth/2.3,
        width=\columnwidth/1.9,
        ylabel={\it Accuracy},
        xmin=0.5, xmax=9.5,
        ymin=74, ymax=82,
        xtick={1,2,3,4,5,6,7,8,9},
        ytick={74,76,78,80,82},
        xticklabel style = {font=\tiny},
        yticklabel style = {font=\scriptsize},
        xticklabels={0.1,,0.3,,0.5,,0.7,,0.9},
        yticklabels={74,76,78,80,82},
    ]
    \addplot[line width=0.3mm, mark=triangle, color=PageRank-color]  %
        plot coordinates {
(1,	76.29	)
(2,	79.53	)
(3,	80.23	)
(4,	79.25	)
(5,	79.04	)
(6,	78.33	)
(7,	78.02	)
(8,	76.87	)
(9,	75.79	)
    };

    \addplot[line width=0.3mm, mark=diamond, color=Vanilla-color]  %
        plot coordinates {
(1,	76.09	)
(2,	79.32	)
(3,	80.92	)
(4,	80.05	)
(5,	79.02	)
(6,	78.32	)
(7,	77.85	)
(8,	76.98	)
(9,	76.18	)
    };

    \addplot[line width=0.3mm, mark=o, color=HeatKernel-color]  %
        plot coordinates {
(1,	75.83	)
(2,	79.56	)
(3,	80.24	)
(4,	79.16	)
(5,	79.04	)
(6,	78.5	)
(7,	78	)
(8,	77.07	)
(9,	76.06	)
    };

    \end{axis}
\end{tikzpicture}\hspace{1mm}\label{fig:vary-alpha}%
}
\subfloat[Varying $\alpha$ on {\em WikiCS}]{
\begin{tikzpicture}[scale=1,every mark/.append style={mark size=2pt}]
    \begin{axis}[
        height=\columnwidth/2.3,
        width=\columnwidth/1.9,
        ylabel={\it Accuracy},
        xmin=0.5, xmax=9.5,
        ymin=81, ymax=87,
        xtick={1,2,3,4,5,6,7,8,9},
        ytick={81,83,85,87},
        xticklabel style = {font=\tiny},
        yticklabel style = {font=\scriptsize},
        xticklabels={0.1,,0.3,,0.5,,0.7,,0.9},
        yticklabels={81,83,85,87},
    ]
    \addplot[line width=0.3mm, mark=triangle, color=PageRank-color]  %
        plot coordinates {
(1,	85.94	)
(2,	86.41	)
(3,	86.21	)
(4,	86.08	)
(5,	85.5	)
(6,	85.25	)
(7,	84.91	)
(8,	84.23	)
(9,	83.71	)
    };

    \addplot[line width=0.3mm, mark=diamond, color=Vanilla-color]  %
        plot coordinates {
(1,	86.04	)
(2,	86.26	)
(3,	86.06	)
(4,	85.81	)
(5,	85.26	)
(6,	84.06	)
(7,	83.88	)
(8,	83.15	)
(9,	82.47	)
    };

    \addplot[line width=0.3mm, mark=o, color=HeatKernel-color]  %
        plot coordinates {
(1,	85.53	)
(2,	86.27	)
(3,	86.54	)
(4,	86.41	)
(5,	86.37	)
(6,	86	)
(7,	85.81	)
(8,	86.14	)
(9,	85.5	)
    };

    \end{axis}
\end{tikzpicture}\hspace{1mm}\label{fig:vary-alpha-wikics}%
}
\vspace{-4mm}
\subfloat[Varying $\beta$ on {\em CiteSeer}]{
\begin{tikzpicture}[scale=1,every mark/.append style={mark size=2pt}]
    \begin{axis}[
        height=\columnwidth/2.3,
        width=\columnwidth/1.9,
        ylabel={\it Accuracy},
        xmin=0.5, xmax=9.5,
        ymin=50, ymax=82,
        xtick={1,2,3,4,5,6,7,8,9},
        ytick={50,58,66,74,82},
        xticklabel style = {font=\tiny},
        yticklabel style = {font=\scriptsize},
        xticklabels={0.1,,0.3,,0.5,,0.7,,0.9},
        yticklabels={50,58,66,74,82},
    ]
    \addplot[line width=0.3mm, mark=triangle, color=PageRank-color]  %
        plot coordinates {
(1,	79.86	)
(2,	68.09	)
(3,	55.16	)
(4,	53.19	)
(5,	55.14	)
(6,	55.92	)
(7,	55.55	)
(8,	55.26	)
(9,	55.7	)
    };

    \addplot[line width=0.3mm, mark=diamond, color=Vanilla-color]  %
        plot coordinates {
(1,	80.83	)
(2,	75.43	)
(3,	63.44	)
(4,	60.89	)
(5,	58.17	)
(6,	55.11	)
(7,	55.23	)
(8,	54.71	)
(9,	57.76	)
    };

    \addplot[line width=0.3mm, mark=o, color=HeatKernel-color]  %
        plot coordinates {
(1,	79.94	)
(2,	71.61	)
(3,	59.71	)
(4,	58.65	)
(5,	61.28	)
(6,	60.51	)
(7,	59.44	)
(8,	57.62	)
(9,	57.91	)
    };

    \end{axis}
\end{tikzpicture}\hspace{1mm}\label{fig:vary-beta}%
}
\subfloat[Varying $\beta$ on {\em WikiCS}]{
\begin{tikzpicture}[scale=1,every mark/.append style={mark size=2pt}]
    \begin{axis}[
        height=\columnwidth/2.3,
        width=\columnwidth/1.9,
        ylabel={\it Accuracy},
        xmin=0.5, xmax=9.5,
        ymin=85, ymax=87,
        xtick={1,2,3,4,5,6,7,8,9},
        ytick={85,85.5,86,86.6,87},
        xticklabel style = {font=\tiny},
        yticklabel style = {font=\scriptsize},
        xticklabels={0.1,,0.3,,0.5,,0.7,,0.9},
        yticklabels={85,,86,,87},
    ]
    \addplot[line width=0.3mm, mark=triangle, color=PageRank-color]  %
        plot coordinates {
(1,	85.97	)
(2,	86.17	)
(3,	86.25	)
(4,	86.33	)
(5,	86.01	)
(6,	86.23	)
(7,	86.22	)
(8,	85.79	)
(9,	85.96	)
    };

    \addplot[line width=0.3mm, mark=diamond, color=Vanilla-color]  %
        plot coordinates {
(1,	86.01	)
(2,	86.26	)
(3,	86.31	)
(4,	85.93	)
(5,	85.94	)
(6,	85.72	)
(7,	85.62	)
(8,	85.91	)
(9,	86.18	)
    };

    \addplot[line width=0.3mm, mark=o, color=HeatKernel-color]  %
        plot coordinates {
(1,	85.86	)
(2,	86.21	)
(3,	86.19	)
(4,	86.14	)
(5,	86.09	)
(6,	86.06	)
(7,	86.06	)
(8,	86.29	)
(9,	86.43	)
    };

    \end{axis}
\end{tikzpicture}\hspace{1mm}\label{fig:vary-beta-wikics}%
}
\vspace{-4mm}
\subfloat[Varying $\eta$ on {\em CiteSeer}]{
\begin{tikzpicture}[scale=1,every mark/.append style={mark size=2pt}]
    \begin{axis}[
        height=\columnwidth/2.3,
        width=\columnwidth/1.9,
        ylabel={\it Accuracy},
        xmin=0.5, xmax=9.5,
        ymin=80, ymax=82,
        xtick={1,2,3,4,5,6,7,8,9},
        ytick={80,80.5,81,81.5,82},
        xticklabel style = {font=\tiny},
        yticklabel style = {font=\scriptsize},
        xticklabels={0.1,,0.3,,0.5,,0.7,,0.9},
        yticklabels={80,,81,,82},
    ]
    \addplot[line width=0.3mm, mark=triangle, color=PageRank-color]  %
        plot coordinates {
(1,	80.69	)
(2,	80.83	)
(3,	80.68	)
(4,	80.74	)
(5,	81.01	)
(6,	80.69	)
(7,	80.8	)
(8,	80.78	)
(9,	80.75	)
    };

    \addplot[line width=0.3mm, mark=diamond, color=Vanilla-color]  %
        plot coordinates {
(1,	80.35	)
(2,	80.53	)
(3,	80.74	)
(4,	80.68	)
(5,	80.8	)
(6,	80.83	)
(7,	80.93	)
(8,	81.01	)
(9,	80.95	)
    };

    \addplot[line width=0.3mm, mark=o, color=HeatKernel-color]  %
        plot coordinates {
(1,	80.53	)
(2,	80.6	)
(3,	80.56	)
(4,	80.74	)
(5,	80.77	)
(6,	80.62	)
(7,	80.93	)
(8,	80.92	)
(9,	80.89	)
    };

    \end{axis}
\end{tikzpicture}\hspace{1mm}\label{fig:vary-eta}%
}
\subfloat[Varying $\eta$ on {\em WikiCS}]{
\begin{tikzpicture}[scale=1,every mark/.append style={mark size=2pt}]
    \begin{axis}[
        height=\columnwidth/2.3,
        width=\columnwidth/1.9,
        ylabel={\it Accuracy},
        xmin=0.5, xmax=9.5,
        ymin=85, ymax=87.5,
        xtick={1,2,3,4,5,6,7,8,9},
        ytick={85,85.5,86,86.5,87,87.5},
        xticklabel style = {font=\tiny},
        yticklabel style = {font=\scriptsize},
        xticklabels={0.1,,0.3,,0.5,,0.7,,0.9},
        yticklabels={85,,86,,87},
    ]
    \addplot[line width=0.3mm, mark=triangle, color=PageRank-color]  %
        plot coordinates {
(1,	86.23	)
(2,	86.24	)
(3,	86.15	)
(4,	86.08	)
(5,	86.16	)
(6,	86.04	)
(7,	86.06	)
(8,	86.15	)
(9,	86.35	)
    };

    \addplot[line width=0.3mm, mark=diamond, color=Vanilla-color]  %
        plot coordinates {
(1,	86.25	)
(2,	86.14	)
(3,	86.4	)
(4,	86.21	)
(5,	86.39	)
(6,	86.48	)
(7,	86.35	)
(8,	86.47	)
(9,	86.13	)
    };

    \addplot[line width=0.3mm, mark=o, color=HeatKernel-color]  %
        plot coordinates {
(1,	86.52	)
(2,	86.52	)
(3,	86.5	)
(4,	86.44	)
(5,	86.61	)
(6,	86.56	)
(7,	86.53	)
(8,	86.56	)
(9,	86.56	)
    };

    \end{axis}
\end{tikzpicture}\hspace{1mm}\label{fig:vary-eta-wikics}%
}
\vspace{-4mm}
\subfloat[Varying $\kappa$ on {\em CiteSeer}]{
\begin{tikzpicture}[scale=1,every mark/.append style={mark size=2pt}]
    \begin{axis}[
        height=\columnwidth/2.3,
        width=\columnwidth/1.9,
        ylabel={\it Accuracy},
        xmin=0.5, xmax=7.5,
        ymin=78, ymax=82,
        xtick={1,2,3,4,5,6,7},
        ytick={78,79,80,81,82},
        xticklabel style = {font=\tiny},
        yticklabel style = {font=\scriptsize},
        xticklabels={8,16,32,64,128,256,$n$},
        yticklabels={78,79,80,81,82},
    ]
    \addplot[line width=0.3mm, mark=triangle, color=PageRank-color]  %
        plot coordinates {
(1,	80.45	)
(2,	80.77	)
(3,	80.86	)
(4,	80.74	)
(5,	80.35	)
(6,	80.33	)
(7,	78.05	)
    };

    \addplot[line width=0.3mm, mark=diamond, color=Vanilla-color]  %
        plot coordinates {
(1,	80.65	)
(2,	80.62	)
(3,	80.77	)
(4,	81.01	)
(5,	80.77	)
(6,	80.9	)
(7,	80.27	)
    };

    \addplot[line width=0.3mm, mark=o, color=HeatKernel-color]  %
        plot coordinates {
(1, 80.51	)
(2, 80.54	)
(3, 80.59	)
(4, 80.77	)
(5, 80.65	)
(6, 80.81	)
(7,	80.96	)

    };

    \end{axis}
\end{tikzpicture}\hspace{1mm}\label{fig:vary-k}%
}
\subfloat[Varying $\kappa$ on {\em WikiCS}]{
\begin{tikzpicture}[scale=1,every mark/.append style={mark size=2pt}]
    \begin{axis}[
        height=\columnwidth/2.3,
        width=\columnwidth/1.9,
        ylabel={\it Accuracy},
        xmin=0.5, xmax=7.5,
        ymin=85, ymax=87,
        xtick={1,2,3,4,5,6,7},
        ytick={85,85.5,86,86.5,87},
        xticklabel style = {font=\tiny},
        yticklabel style = {font=\scriptsize},
        xticklabels={8,16,32,64,128,256,$n$},
        yticklabels={85,,86,,87},
    ]
    \addplot[line width=0.3mm, mark=triangle, color=PageRank-color]  %
        plot coordinates {
(1,	86.21	)
(2,	86.01	)
(3,	86.25	)
(4,	86.45	)
(5,	86.08	)
(6,	86.2	)
(7,	85.37	)
    };

    \addplot[line width=0.3mm, mark=diamond, color=Vanilla-color]  %
        plot coordinates {
(1,	86.18	)
(2,	86.09	)
(3,	86.67	)
(4,	86.57	)
(5,	86.58	)
(6,	86.48	)
(7,	86.53	)
    };

    \addplot[line width=0.3mm, mark=o, color=HeatKernel-color]  %
        plot coordinates {
(1,	86.58	)
(2,	86.61	)
(3,	86.52	)
(4,	86.39	)
(5,	86.44	)
(6,	86.52	)
(7, 86.56   )

    };

    \end{axis}
\end{tikzpicture}\hspace{1mm}\label{fig:vary-k-wikics}%
}
\end{small}
 \vspace{-1ex}
\caption{Accuracy when varying parameters.} \label{fig:parameter}
\vspace{-1ex}
\end{figure}
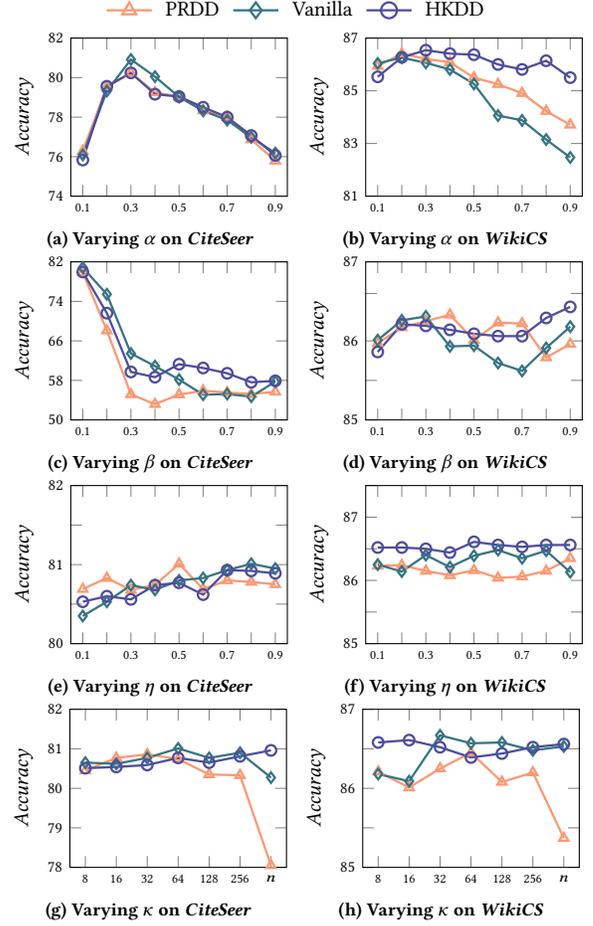

\section{Conclusion}\label{sec:conclusion}
\alg is a novel MPNN model based on an optimization framework with a diffusion distance-guided stress function and orthogonal regularization terms for overcoming the limitations of previous Dirichlet energy-based MPNNs.
We devise fast spectral decomposition approaches for diffusion distance approximations and provide non-trivial theoretical analyses.
Extensive experiments validate that our \alg consistently outperforms existing MPNNs across diverse benchmarking datasets.

While its relative efficiency is demonstrated in Appendix~\ref{sec:scalability}, directly applying \alg to massive graphs with billions of nodes and dynamic graphs remains challenging.
In our future work, we will focus on devising incremental methods to cope with frequent updates in dynamic graphs and algorithmic optimizations for higher practical scalability.
We also plan to extend \alg to other graph learning tasks, including link prediction and graph regression.

\begin{acks}
This work is partially supported by the National Natural Science Foundation of China (No. 62302414), the Hong Kong RGC ECS grant (No. 22202623) and YCRG (No. C2003-23Y), Guangdong Basic and Applied Basic Research Foundation (Project No. 2023B1515130002), the Huawei Gift Fund, and Guangdong and Hong Kong Universities ``1+1+1'' Joint Research Collaboration Scheme, project No.: 2025A0505000002.
\end{acks}

\balance

\bibliographystyle{ACM-Reference-Format}
\bibliography{sample-base}

\newpage
\onecolumn
\appendix

\section{Notations and Tables}\label{sec:add-tables}

\begin{table}[hb]
\centering
\caption{Frequently used notations.} \label{tbl:notations}
\vspace{-1ex}
\begin{small}
\begin{tabular}{l|p{0.7\columnwidth}}
    \toprule
    {\bf Symbol}  &  {\bf Description}\\
    \midrule
    $\V,\EDG$   & The node set and edge set, respectively.\\ 
    $n, m$  & The size of the node set and edge set, respectively. \\ 
    $\XM, \HM$  & The initial node attribute matrix and the learned node representation.\\ 
    $d, d^{(0)}$  & The node attribute's hidden dimensionality and initial dimensionality, respectively.\\ 
    $\PM, \DM, d_i$  & The transition, degree matrix of the graph, and the degree of a specific node $i$.\\ 
    $\AM, \NAM, \LM, \NLM$  & The adjacency, normalized adjacency, Laplacian and normalized Laplacian matrix of the graph.\\ 
    $\mathcal{D}(\HM)$  & The Dirichlet energy of the target node representations $\HM$.\\ 
    $\YM, |\Y|$  & The node label matrix and the size of the label set.\\ 
    $\Delta, \delta_{i,j}$  & The distance matrix and a distance between two nodes $i$ and $j$.\\ 
    $\alpha, \beta, \gamma$   & The coefficients in message functions (Eq. (\ref{eq:propagation-rule})).\\ 
    $\kappa$  & The truncated dimensionality.\\ 
    
    \bottomrule
    
\end{tabular}%
\end{small}
\vspace{-1ex}
\end{table}

\begin{table*}[h!]
\centering
\caption{Overview of \(\omega\), \(\xi\) configurations, and node representation formulations in MPNNs.}\label{tbl:GNNs}
\vspace{-1ex}
\begin{small}
\begin{tabular}{llccc}
\toprule
Model & \(\HM\) & \(\omega\) & \(\xi\) & \(\HM^{(0)}\)\\
\midrule
GCN/SGC~\cite{kipf2017semisupervised} & \(\HM=\NAM^K\HM^{(0)}\) & \(1\) &  - & \(\XM\WM\)\\
PPNP~\cite{gasteiger2019combining} & \(\HM=\alpha\left(\IM -(1-\alpha)\NAM \right)^{-1}\HM^{(0)}\) & \(\frac{1}{\alpha}-1\) &  \(\|\HM-\HM^{(0)}\|_F^2\) &\(\textsf{MLP}(\XM)\)  \\
APPNP~\cite{gasteiger2019combining} & \(\HM=\alpha\sum_{k=0}^K{(1-\alpha)^k\NAM^k\HM^{(0)}}\) & \(\frac{1}{\alpha}-1\) &  \(\|\HM-\HM^{(0)}\|_F^2\) &\(\textsf{MLP}(\XM)\)  \\
JKNet~\cite{xu2018representation} & \(\HM=\sum_{k=1}^K{\frac{\alpha^{k-1}}{(1+\alpha)^{k}}\NAM^k\HM^{(0)}}\) & \(\alpha\) & \(\|\NAM\HM-\HM^{(0)}\|^2_F\) & \(\XM\WM\) \\
DAGNN~\cite{liu2020towards} & \(\HM=\sum_{k=0}^K{\frac{\alpha^{k}}{(1+\alpha)^{k+1}}\NAM^k\HM^{(0)}}\) & \(\alpha\) & \(\|\HM-\HM^{(0)}\|_F^2\) & \(\textsf{MLP}(\XM)\) \\
GNN-LF~\cite{zhu2021interpreting} & \(\HM=\{ \left( \mu + \frac{1}{\alpha} - 1 \right) \IM + \left( 2 - \mu - \frac{1}{\alpha} \right) \NAM \}^{-1} \{ \mu \IM + (1 - \mu) \NAM \}\HM^{(0)}\) & \(\frac{1}{\alpha}-1\) & \(\|\{ \mu \IM + (1 - \mu) \NAM \}^{\frac{1}{2}}(\HM-\HM^{(0)})\|_F^2\) & \(\textsf{MLP}(\XM)\) \\
GNN-HF~\cite{zhu2021interpreting} & \(\HM=\{ \left( \beta + \frac{1}{\alpha} \right) \IM + \left( 1 - \beta - \frac{1}{\alpha} \right) \NAM \}^{-1} \{\IM + \beta \NLM \}\HM^{(0)}\) & \(\frac{1}{\alpha}-1\) & \(\|\{\IM + \beta \NLM \}^{\frac{1}{2}}(\HM-\HM^{(0)})\|_F^2\) & \(\textsf{MLP}(\XM)\) \\

\bottomrule
\end{tabular}
\end{small}
\vspace{-1ex}
\end{table*}

\section{Empirical Studies of Over-smoothing and Over-correlation in MPNNs}\label{sec:exp-issues}

SMV~\cite{liu2020towards} calculates the average normalized Euclidean
distance between any two nodes in $\HM^{(k)}$:
\begin{equation}\label{eq:SMV}
\text{SMV}(\HM^{(k)}) = \frac{1}{2n(n-1)} \sum_{v_i \neq v_j} \left\| \frac{\HM^{(k)}_{i}}{\|\HM^{(k)}_{i}\|_2}- \frac{\HM^{(k)}_{j}}{\|\HM^{(k)}_{j}\|_2}\right\|_2.
\end{equation}

Corr~\cite{jin2022feature} is defined as the average {\em Pearson correlation coefficient}~\cite{cohen2009pearson} between any two feature dimensions in $\HM^{(k)}$:
\begin{equation}\label{eq:Corr}
\text{Corr}(\HM^{(k)}) = \frac{1}{d(d-1)} \sum_{i \neq j \in [1,d]}{\left|\frac{\text{cov}(\HM^{(k)}_{\cdot,i},\HM^{(k)}_{\cdot,j})}{\sigma(\HM^{(k)}_{\cdot,i})\cdot \sigma(\HM^{(k)}_{\cdot,j})}\right|},
\end{equation}
where \(\text{cov}(\cdot,\cdot)\) and \(\sigma(\cdot)\) denote the covariance and standard deviation, respectively.

\begin{figure}[!hb]
\centering
\begin{small}
\begin{tikzpicture}
    \begin{customlegend}[legend columns=4,
        legend entries={Acc, SMV, Corr, HOS},
        legend style={at={(0.45,1.35)},anchor=north,draw=none,font=\small,column sep=0.1cm}]
    \addlegendimage{line width=0.5mm,mark size=1pt,mark=dot, color=PageRank-color}
    \addlegendimage{line width=0.5mm,mark size=1pt,mark=dot, color=Vanilla-color}
    \addlegendimage{line width=0.5mm,mark size=1pt,mark=dot, color=HeatKernel-color}
    \addlegendimage{line width=0.5mm,mark size=1pt,mark=dot, color=myred2}
    
    \end{customlegend}
\end{tikzpicture}
\\[-\lineskip]
\vspace{-4mm}
\subfloat[{\em CiteSeer}]{
\begin{tikzpicture}[scale=1,every mark/.append style={mark size=1pt}]
    \begin{axis}[
        height=\columnwidth/5.4,
        width=\columnwidth/4.6,
        xmin=1, xmax=20,
        ymin=0, ymax=1,
        xtick={2,4,6,8,12,16,20},
        ytick={0.0,0.2,0.4,0.6,0.8,1.0},
        xticklabel style = {font=\tiny},
        yticklabel style = {font=\scriptsize},
        xticklabels={2,4,6,8,12,16,20},
        yticklabels={0.0,0.2,0.4,0.6,0.8,1.0},
    ]
    \addplot[line width=0.5mm, mark=dot, color=PageRank-color]
        plot coordinates {
(1  ,	0.754	)
(2  ,	0.760	)
(4  ,	0.746	)
(6  ,	0.741	)
(8  ,	0.617	)
(10 ,	0.205	)
(15 ,	0.206	)
(20 ,	0.204	)
    };

    \addplot[line width=0.5mm, mark=dot, color=Vanilla-color]
        plot coordinates {
(1  ,	0.641	)
(2  ,	0.633	)
(4  ,	0.655	)
(6  ,	0.637	)
(8  ,	0.550	)
(10 ,	0.000	)
(15 ,	0.000	)
(20 ,	0.000	)
    };

    \addplot[line width=0.5mm, mark=dot, color=HeatKernel-color]
        plot coordinates {
(1  ,	0.250	)
(2  ,	0.308	)
(4  ,	0.329	)
(6  ,	0.405	)
(8  ,	0.595	)
(10 ,	1.000	)
(15 ,	1.000	)
(20 ,	1.000	)
    };
    
    \addplot[line width=0.5mm, mark=dot, color=myred2]
        plot coordinates {
(1  ,	0.194	)
(2  ,	0.143	)
(4  ,	0.110	)
(6  ,	0.090	)
(8  ,	0.070	)
(10 ,	0.000	)
(15 ,	0.000	)
(20 ,	0.000	)
    };

    \end{axis}
\end{tikzpicture}\label{fig:over-smoothing-citeseer}%
}
\subfloat[{\em CS}]{
\begin{tikzpicture}[scale=1,every mark/.append style={mark size=1pt}]
    \begin{axis}[
        height=\columnwidth/5.4,
        width=\columnwidth/4.6,
        xmin=1, xmax=20,
        ymin=0, ymax=1,
        xtick={2,4,6,8,12,16,20},
        ytick={0.0,0.2,0.4,0.6,0.8,1.0},
        xticklabel style = {font=\tiny},
        yticklabel style = {font=\scriptsize},
        xticklabels={2,4,6,8,12,16,20},
        yticklabels={0.0,0.2,0.4,0.6,0.8,1.0},
    ]
    \addplot[line width=0.5mm, mark=dot, color=PageRank-color]
        plot coordinates {
(1  ,	0.923	)
(2  ,	0.930	)
(4  ,	0.919	)
(6  ,	0.913	)
(8  ,	0.908	)
(10 ,	0.788	)
(15 ,	0.568	)
(20 ,	0.198	)
    };

    \addplot[line width=0.5mm, mark=dot, color=Vanilla-color]
        plot coordinates {
(1  ,	0.504	)
(2  ,	0.577	)
(4  ,	0.566	)
(6  ,	0.559	)
(8  ,	0.558	)
(10 ,	0.480	)
(15 ,	0.394	)
(20 ,	0.004	)
    };

    \addplot[line width=0.5mm, mark=dot, color=HeatKernel-color]
        plot coordinates {
(1  ,	0.286	)
(2  ,	0.224	)
(4  ,	0.261	)
(6  ,	0.293	)
(8  ,	0.317	)
(10 ,	0.415	)
(15 ,	0.587	)
(20 ,	0.475	)
    };
    
    \addplot[line width=0.5mm, mark=dot, color=myred2]
        plot coordinates {
(1  ,	0.292	)
(2  ,	0.337	)
(4  ,	0.309	)
(6  ,	0.296	)
(8  ,	0.293	)
(10 ,	0.240	)
(15 ,	0.183	)
(20 ,	0.001	)
    };

    \end{axis}
\end{tikzpicture}\label{fig:over-smoothing-cs}%
}
\subfloat[{\em WikiCS}]{
\begin{tikzpicture}[scale=1,every mark/.append style={mark size=1pt}]
    \begin{axis}[
        height=\columnwidth/5.4,
        width=\columnwidth/4.6,
        xmin=1, xmax=20,
        ymin=0, ymax=1,
        xtick={2,4,6,8,12,16,20},
        ytick={0.0,0.2,0.4,0.6,0.8,1.0},
        xticklabel style = {font=\tiny},
        yticklabel style = {font=\scriptsize},
        xticklabels={2,4,6,8,12,16,20},
        yticklabels={0.0,0.2,0.4,0.6,0.8,1.0},
    ]
    \addplot[line width=0.5mm, mark=dot, color=PageRank-color]
        plot coordinates {
(1  ,	0.829	)
(2  ,	0.840	)
(4  ,	0.829	)
(6  ,	0.815	)
(8  ,	0.776	)
(10 ,	0.586	)
(15 ,	0.218	)
(20 ,	0.212	)
    };

    \addplot[line width=0.5mm, mark=dot, color=Vanilla-color]
        plot coordinates {
(1  ,	0.557	)
(2  ,	0.534	)
(4  ,	0.506	)
(6  ,	0.511	)
(8  ,	0.488	)
(10 ,	0.372	)
(15 ,	0.047	)
(20 ,	0.002	)
    };

    \addplot[line width=0.5mm, mark=dot, color=HeatKernel-color]
        plot coordinates {
(1  ,	0.273	)
(2  ,	0.275	)
(4  ,	0.320	)
(6  ,	0.366	)
(8  ,	0.359	)
(10 ,	0.514	)
(15 ,	0.857	)
(20 ,	0.895	)
    };
    
    \addplot[line width=0.5mm, mark=dot, color=myred2]
        plot coordinates {
(1  ,	0.301	)
(2  ,	0.286	)
(4  ,	0.257	)
(6  ,	0.259	)
(8  ,	0.242	)
(10 ,	0.188	)
(15 ,	0.022	)
(20 ,	0.000	)
    };

    \end{axis}
\end{tikzpicture}\label{fig:over-smoothing-wikics}%
}
\subfloat[{\em Cornell}]{
\begin{tikzpicture}[scale=1,every mark/.append style={mark size=1pt}]
    \begin{axis}[
        height=\columnwidth/5.4,
        width=\columnwidth/4.6,
        xmin=1, xmax=20,
        ymin=0, ymax=1,
        xtick={2,4,6,8,12,16,20},
        ytick={0.0,0.2,0.4,0.6,0.8,1.0},
        xticklabel style = {font=\tiny},
        yticklabel style = {font=\scriptsize},
        xticklabels={2,4,6,8,12,16,20},
        yticklabels={0.0,0.2,0.4,0.6,0.8,1.0},
    ]
    \addplot[line width=0.5mm, mark=dot, color=PageRank-color]
        plot coordinates {
(1  ,	0.427	)
(2  ,	0.432	)
(4  ,	0.411	)
(6  ,	0.397	)
(8  ,	0.405	)
(10 ,	0.389	)
(15 ,	0.449	)
(20 ,	0.419	)
    };

    \addplot[line width=0.5mm, mark=dot, color=Vanilla-color]
        plot coordinates {
(1  ,	0.532	)
(2  ,	0.558	)
(4  ,	0.519	)
(6  ,	0.469	)
(8  ,	0.368	)
(10 ,	0.344	)
(15 ,	0.096	)
(20 ,	0.000	)
    };

    \addplot[line width=0.5mm, mark=dot, color=HeatKernel-color]
        plot coordinates {
(1  ,	0.260	)
(2  ,	0.333	)
(4  ,	0.383	)
(6  ,	0.474	)
(8  ,	0.601	)
(10 ,	0.787	)
(15 ,	0.955	)
(20 ,	1.000	)
    };
    
    \addplot[line width=0.5mm, mark=dot, color=myred2]
        plot coordinates {
(1  ,	0.454	)
(2  ,	0.482	)
(4  ,	0.410	)
(6  ,	0.285	)
(8  ,	0.211	)
(10 ,	0.224	)
(15 ,	0.038	)
(20 ,	0.000	)
    };

    \end{axis}
\end{tikzpicture}\label{fig:over-smoothing-cornell}%
}
\subfloat[{\em Wisconsin}]{
\begin{tikzpicture}[scale=1,every mark/.append style={mark size=1pt}]
    \begin{axis}[
        height=\columnwidth/5.4,
        width=\columnwidth/4.6,
        xmin=1, xmax=20,
        ymin=0, ymax=1,
        xtick={2,4,6,8,12,16,20},
        ytick={0.0,0.2,0.4,0.6,0.8,1.0},
        xticklabel style = {font=\tiny},
        yticklabel style = {font=\scriptsize},
        xticklabels={2,4,6,8,12,16,20},
        yticklabels={0.0,0.2,0.4,0.6,0.8,1.0},
    ]
    \addplot[line width=0.5mm, mark=dot, color=PageRank-color]
        plot coordinates {
(1  ,	0.524	)
(2  ,	0.484	)
(4  ,	0.482	)
(6  ,	0.494	)
(8  ,	0.512	)
(10 ,	0.444	)
(15 ,	0.476	)
(20 ,	0.490	)
(25 ,	0.472	)
(30 ,	0.468	)
(35 ,	0.458	)
(40 ,	0.474	)
(45 ,	0.482	)
(50 ,	0.446	)
    };

    \addplot[line width=0.5mm, mark=dot, color=Vanilla-color]
        plot coordinates {
(1  ,	0.504	)
(2  ,	0.513	)
(4  ,	0.480	)
(6  ,	0.327	)
(8  ,	0.297	)
(10 ,	0.000	)
(15 ,	0.000	)
(20 ,	0.000	)
(25 ,	0.000	)
(30 ,	0.000	)
(35 ,	0.000	)
(40 ,	0.000	)
(45 ,	0.000	)
(50 ,	0.000	)
    };

    \addplot[line width=0.5mm, mark=dot, color=HeatKernel-color]
        plot coordinates {
(1  ,	0.269	)
(2  ,	0.386	)
(4  ,	0.543	)
(6  ,	0.882	)
(8  ,	0.992	)
(10 ,	1.000	)
(15 ,	1.000	)
(20 ,	1.000	)
(25 ,	1.000	)
(30 ,	1.000	)
(35 ,	1.000	)
(40 ,	1.000	)
(45 ,	1.000	)
(50 ,	1.000	)
    };
    
    \addplot[line width=0.5mm, mark=dot, color=myred2]
        plot coordinates {
(1  ,	0.442	)
(2  ,	0.413	)
(4  ,	0.342	)
(6  ,	0.139	)
(8  ,	0.114	)
(10 ,	0.000	)
(15 ,	0.000	)
(20 ,	0.000	)
(25 ,	0.000	)
(30 ,	0.000	)
(35 ,	0.000	)
(40 ,	0.000	)
(45 ,	0.000	)
(50 ,	0.000	)
    };

    \end{axis}
\end{tikzpicture}\label{fig:over-smoothing-wisconsin}%
}
\subfloat[{\em Chameleon}]{
\begin{tikzpicture}[scale=1,every mark/.append style={mark size=1pt}]
    \begin{axis}[
        height=\columnwidth/5.4,
        width=\columnwidth/4.6,
        xmin=1, xmax=20,
        ymin=0, ymax=1,
        xtick={2,4,6,8,12,16,20},
        ytick={0.0,0.2,0.4,0.6,0.8,1.0},
        xticklabel style = {font=\tiny},
        yticklabel style = {font=\scriptsize},
        xticklabels={2,4,6,8,12,16,20},
        yticklabels={0.0,0.2,0.4,0.6,0.8,1.0},
    ]
    \addplot[line width=0.5mm, mark=dot, color=PageRank-color]
        plot coordinates {
(1  ,	0.609	)
(2  ,	0.663	)
(4  ,	0.646	)
(6  ,	0.594	)
(8  ,	0.534	)
(10 ,	0.432	)
(15 ,	0.259	)
(20 ,	0.226	)
    };

    \addplot[line width=0.5mm, mark=dot, color=Vanilla-color]
        plot coordinates {
(1  ,	0.610	)
(2  ,	0.643	)
(4  ,	0.580	)
(6  ,	0.572	)
(8  ,	0.507	)
(10 ,	0.539	)
(15 ,	0.156	)
(20 ,	0.000	)
    };

    \addplot[line width=0.5mm, mark=dot, color=HeatKernel-color]
        plot coordinates {
(1  ,	0.233	)
(2  ,	0.253	)
(4  ,	0.402	)
(6  ,	0.407	)
(8  ,	0.481	)
(10 ,	0.617	)
(15 ,	0.923	)
(20 ,	1.000	)
    };
    
    \addplot[line width=0.5mm, mark=dot, color=myred2]
        plot coordinates {
(1  ,	0.537	)
(2  ,	0.532	)
(4  ,	0.425	)
(6  ,	0.308	)
(8  ,	0.212	)
(10 ,	0.149	)
(15 ,	0.043	)
(20 ,	0.000	)
    };

    \end{axis}
\end{tikzpicture}\label{fig:over-smoothing-chameleon}%
}
\end{small}
 \vspace{-2ex}
\caption{Acc, SMV, Corr, and HOS of $\HM^{(k)}$ in GAT.} \label{fig:over-smoothing-GAT}
\vspace{0ex}
\end{figure}
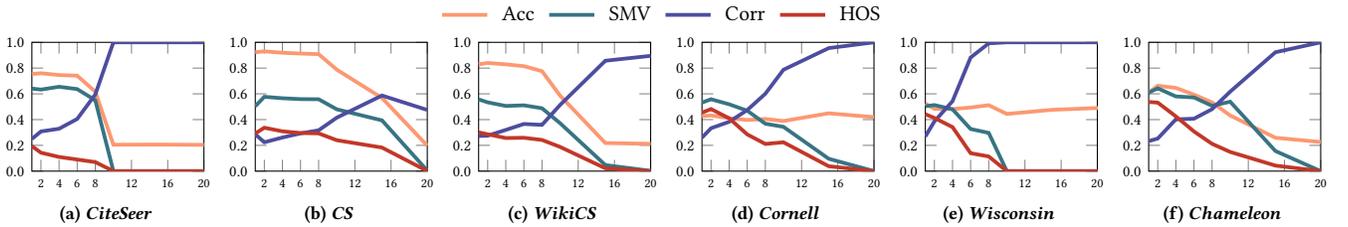

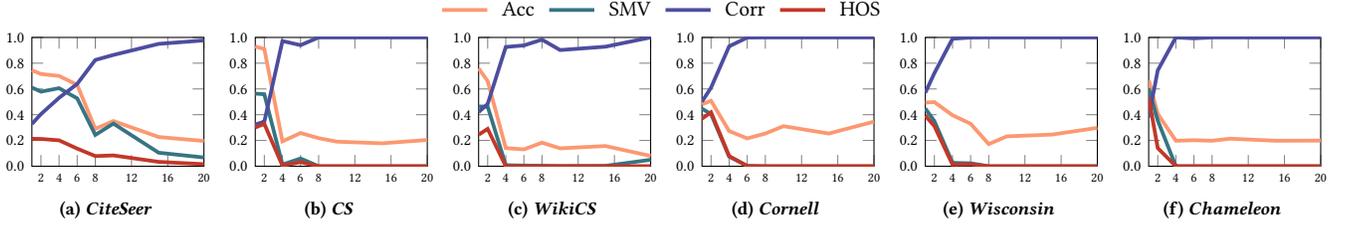
\begin{figure}[!h]
\centering
\begin{small}
\begin{tikzpicture}
    \begin{customlegend}[legend columns=4,
        legend entries={Acc, SMV, Corr, HOS},
        legend style={at={(0.45,1.35)},anchor=north,draw=none,font=\small,column sep=0.1cm}]
    \addlegendimage{line width=0.5mm,mark size=1pt,mark=dot, color=PageRank-color}
    \addlegendimage{line width=0.5mm,mark size=1pt,mark=dot, color=Vanilla-color}
    \addlegendimage{line width=0.5mm,mark size=1pt,mark=dot, color=HeatKernel-color}
    \addlegendimage{line width=0.5mm,mark size=1pt,mark=dot, color=myred2}
    
    \end{customlegend}
\end{tikzpicture}
\\[-\lineskip]
\vspace{-4mm}
\subfloat[{\em CiteSeer}]{
\begin{tikzpicture}[scale=1,every mark/.append style={mark size=1pt}]
    \begin{axis}[
        height=\columnwidth/5.4,
        width=\columnwidth/4.6,
        xmin=1, xmax=20,
        ymin=0, ymax=1,
        xtick={2,4,6,8,12,16,20},
        ytick={0.0,0.2,0.4,0.6,0.8,1.0},
        xticklabel style = {font=\tiny},
        yticklabel style = {font=\scriptsize},
        xticklabels={2,4,6,8,12,16,20},
        yticklabels={0.0,0.2,0.4,0.6,0.8,1.0},
    ]
    \addplot[line width=0.5mm, mark=dot, color=PageRank-color]
        plot coordinates {
(1  ,	0.745	)
(2  ,	0.716	)
(4  ,	0.701	)
(6  ,	0.632	)
(8  ,	0.290	)
(10 ,	0.351	)
(15 ,	0.227	)
(20 ,	0.196	)
    };

    \addplot[line width=0.5mm, mark=dot, color=Vanilla-color]
        plot coordinates {
(1  ,	0.611	)
(2  ,	0.579	)
(4  ,	0.606	)
(6  ,	0.527	)
(8  ,	0.243	)
(10 ,	0.331	)
(15 ,	0.105	)
(20 ,	0.068	)
    };

    \addplot[line width=0.5mm, mark=dot, color=HeatKernel-color]
        plot coordinates {
(1  ,	0.328	)
(2  ,	0.405	)
(4  ,	0.531	)
(6  ,	0.639	)
(8  ,	0.825	)
(10 ,	0.864	)
(15 ,	0.950	)
(20 ,	0.977	)
    };
    
    \addplot[line width=0.5mm, mark=dot, color=myred2]
        plot coordinates {
(1  ,	0.213	)
(2  ,	0.212	)
(4  ,	0.201	)
(6  ,	0.137	)
(8  ,	0.079	)
(10 ,	0.084	)
(15 ,	0.034	)
(20 ,	0.016	)
    };

    \end{axis}
\end{tikzpicture}\label{fig:over-smoothing-citeseer}%
}
\subfloat[{\em CS}]{
\begin{tikzpicture}[scale=1,every mark/.append style={mark size=1pt}]
    \begin{axis}[
        height=\columnwidth/5.4,
        width=\columnwidth/4.6,
        xmin=1, xmax=20,
        ymin=0, ymax=1,
        xtick={2,4,6,8,12,16,20},
        ytick={0.0,0.2,0.4,0.6,0.8,1.0},
        xticklabel style = {font=\tiny},
        yticklabel style = {font=\scriptsize},
        xticklabels={2,4,6,8,12,16,20},
        yticklabels={0.0,0.2,0.4,0.6,0.8,1.0},
    ]
    \addplot[line width=0.5mm, mark=dot, color=PageRank-color]
        plot coordinates {
(1  ,	0.930	)
(2  ,	0.910	)
(4  ,	0.193	)
(6  ,	0.258	)
(8  ,	0.218	)
(10 ,	0.191	)
(15 ,	0.178	)
(20 ,	0.204	)
    };

    \addplot[line width=0.5mm, mark=dot, color=Vanilla-color]
        plot coordinates {
(1  ,	0.564	)
(2  ,	0.559	)
(4  ,	0.013	)
(6  ,	0.060	)
(8  ,	0.000	)
(10 ,	0.000	)
(15 ,	0.000	)
(20 ,	0.000	)
    };

    \addplot[line width=0.5mm, mark=dot, color=HeatKernel-color]
        plot coordinates {
(1  ,	0.324	)
(2  ,	0.346	)
(4  ,	0.972	)
(6  ,	0.940	)
(8  ,	1.000	)
(10 ,	1.000	)
(15 ,	1.000	)
(20 ,	1.000	)
    };
    
    \addplot[line width=0.5mm, mark=dot, color=myred2]
        plot coordinates {
(1  ,	0.301	)
(2  ,	0.329	)
(4  ,	0.005	)
(6  ,	0.032	)
(8  ,	0.000	)
(10 ,	0.000	)
(15 ,	0.000	)
(20 ,	0.000	)
    };

    \end{axis}
\end{tikzpicture}\label{fig:over-smoothing-cs}%
}
\subfloat[{\em WikiCS}]{
\begin{tikzpicture}[scale=1,every mark/.append style={mark size=1pt}]
    \begin{axis}[
        height=\columnwidth/5.4,
        width=\columnwidth/4.6,
        xmin=1, xmax=20,
        ymin=0, ymax=1,
        xtick={2,4,6,8,12,16,20},
        ytick={0.0,0.2,0.4,0.6,0.8,1.0},
        xticklabel style = {font=\tiny},
        yticklabel style = {font=\scriptsize},
        xticklabels={2,4,6,8,12,16,20},
        yticklabels={0.0,0.2,0.4,0.6,0.8,1.0},
    ]
    \addplot[line width=0.5mm, mark=dot, color=PageRank-color]
        plot coordinates {
(1  ,	0.759	)
(2  ,	0.661	)
(4  ,	0.141	)
(6  ,	0.131	)
(8  ,	0.183	)
(10 ,	0.139	)
(15 ,	0.157	)
(20 ,	0.080	)
    };

    \addplot[line width=0.5mm, mark=dot, color=Vanilla-color]
        plot coordinates {
(1  ,	0.464	)
(2  ,	0.463	)
(4  ,	0.008	)
(6  ,	0.005	)
(8  ,	0.004	)
(10 ,	0.002	)
(15 ,	0.003	)
(20 ,	0.050	)
    };

    \addplot[line width=0.5mm, mark=dot, color=HeatKernel-color]
        plot coordinates {
(1  ,	0.420	)
(2  ,	0.484	)
(4  ,	0.926	)
(6  ,	0.938	)
(8  ,	0.983	)
(10 ,	0.902	)
(15 ,	0.928	)
(20 ,	1.000	)
    };
    
    \addplot[line width=0.5mm, mark=dot, color=myred2]
        plot coordinates {
(1  ,	0.245	)
(2  ,	0.290	)
(4  ,	0.000	)
(6  ,	0.000	)
(8  ,	0.000	)
(10 ,	0.000	)
(15 ,	0.000	)
(20 ,	0.001	)
    };

    \end{axis}
\end{tikzpicture}\label{fig:over-smoothing-wikics}%
}
\subfloat[{\em Cornell}]{
\begin{tikzpicture}[scale=1,every mark/.append style={mark size=1pt}]
    \begin{axis}[
        height=\columnwidth/5.4,
        width=\columnwidth/4.6,
        xmin=1, xmax=20,
        ymin=0, ymax=1,
        xtick={2,4,6,8,12,16,20},
        ytick={0.0,0.2,0.4,0.6,0.8,1.0},
        xticklabel style = {font=\tiny},
        yticklabel style = {font=\scriptsize},
        xticklabels={2,4,6,8,12,16,20},
        yticklabels={0.0,0.2,0.4,0.6,0.8,1.0},
    ]
    \addplot[line width=0.5mm, mark=dot, color=PageRank-color]
        plot coordinates {
(1  ,	0.481	)
(2  ,	0.508	)
(4  ,	0.273	)
(6  ,	0.216	)
(8  ,	0.254	)
(10 ,	0.311	)
(15 ,	0.254	)
(20 ,	0.346	)
    };

    \addplot[line width=0.5mm, mark=dot, color=Vanilla-color]
        plot coordinates {
(1  ,	0.447	)
(2  ,	0.405	)
(4  ,	0.075	)
(6  ,	0.003	)
(8  ,	0.000	)
(10 ,	0.000	)
(15 ,	0.000	)
(20 ,	0.000	)
    };

    \addplot[line width=0.5mm, mark=dot, color=HeatKernel-color]
        plot coordinates {
(1  ,	0.501	)
(2  ,	0.610	)
(4  ,	0.931	)
(6  ,	0.999	)
(8  ,	1.000	)
(10 ,	1.000	)
(15 ,	1.000	)
(20 ,	1.000	)
    };
    
    \addplot[line width=0.5mm, mark=dot, color=myred2]
        plot coordinates {
(1  ,	0.369	)
(2  ,	0.419	)
(4  ,	0.079	)
(6  ,	0.001	)
(8  ,	0.000	)
(10 ,	0.000	)
(15 ,	0.000	)
(20 ,	0.000	)
    };

    \end{axis}
\end{tikzpicture}\label{fig:over-smoothing-cornell}%
}
\subfloat[{\em Wisconsin}]{
\begin{tikzpicture}[scale=1,every mark/.append style={mark size=1pt}]
    \begin{axis}[
        height=\columnwidth/5.4,
        width=\columnwidth/4.6,
        xmin=1, xmax=20,
        ymin=0, ymax=1,
        xtick={2,4,6,8,12,16,20},
        ytick={0.0,0.2,0.4,0.6,0.8,1.0},
        xticklabel style = {font=\tiny},
        yticklabel style = {font=\scriptsize},
        xticklabels={2,4,6,8,12,16,20},
        yticklabels={0.0,0.2,0.4,0.6,0.8,1.0},
    ]
    \addplot[line width=0.5mm, mark=dot, color=PageRank-color]
        plot coordinates {
(1  ,	0.494	)
(2  ,	0.498	)
(4  ,	0.396	)
(6  ,	0.330	)
(8  ,	0.172	)
(10 ,	0.232	)
(15 ,	0.246	)
(20 ,	0.298	)
    };

    \addplot[line width=0.5mm, mark=dot, color=Vanilla-color]
        plot coordinates {
(1  ,	0.448	)
(2  ,	0.350	)
(4  ,	0.026	)
(6  ,	0.023	)
(8  ,	0.000	)
(10 ,	0.000	)
(15 ,	0.000	)
(20 ,	0.000	)
    };

    \addplot[line width=0.5mm, mark=dot, color=HeatKernel-color]
        plot coordinates {
(1  ,	0.571	)
(2  ,	0.723	)
(4  ,	0.990	)
(6  ,	0.999	)
(8  ,	1.000	)
(10 ,	1.000	)
(15 ,	1.000	)
(20 ,	1.000	)
    };
    
    \addplot[line width=0.5mm, mark=dot, color=myred2]
        plot coordinates {
(1  ,	0.393	)
(2  ,	0.311	)
(4  ,	0.013	)
(6  ,	0.015	)
(8  ,	0.000	)
(10 ,	0.000	)
(15 ,	0.000	)
(20 ,	0.000	)
    };

    \end{axis}
\end{tikzpicture}\label{fig:over-smoothing-wisconsin}%
}
\subfloat[{\em Chameleon}]{
\begin{tikzpicture}[scale=1,every mark/.append style={mark size=1pt}]
    \begin{axis}[
        height=\columnwidth/5.4,
        width=\columnwidth/4.6,
        xmin=1, xmax=20,
        ymin=0, ymax=1,
        xtick={2,4,6,8,12,16,20},
        ytick={0.0,0.2,0.4,0.6,0.8,1.0},
        xticklabel style = {font=\tiny},
        yticklabel style = {font=\scriptsize},
        xticklabels={2,4,6,8,12,16,20},
        yticklabels={0.0,0.2,0.4,0.6,0.8,1.0},
    ]
    \addplot[line width=0.5mm, mark=dot, color=PageRank-color]
        plot coordinates {
(1  ,	0.669	)
(2  ,	0.407	)
(4  ,	0.198	)
(6  ,	0.202	)
(8  ,	0.198	)
(10 ,	0.214	)
(15 ,	0.198	)
(20 ,	0.199	)
    };

    \addplot[line width=0.5mm, mark=dot, color=Vanilla-color]
        plot coordinates {
(1  ,	0.603	)
(2  ,	0.355	)
(4  ,	0.000	)
(6  ,	0.000	)
(8  ,	0.000	)
(10 ,	0.000	)
(15 ,	0.000	)
(20 ,	0.000	)
    };

    \addplot[line width=0.5mm, mark=dot, color=HeatKernel-color]
        plot coordinates {
(1  ,	0.380	)
(2  ,	0.744	)
(4  ,	1.000	)
(6  ,	0.993	)
(8  ,	1.000	)
(10 ,	1.000	)
(15 ,	1.000	)
(20 ,	1.000	)
    };
    
    \addplot[line width=0.5mm, mark=dot, color=myred2]
        plot coordinates {
(1  ,	0.534	)
(2  ,	0.141	)
(4  ,	0.000	)
(6  ,	0.000	)
(8  ,	0.000	)
(10 ,	0.000	)
(15 ,	0.000	)
(20 ,	0.000	)
    };

    \end{axis}
\end{tikzpicture}\label{fig:over-smoothing-chameleon}%
}
\end{small}
 \vspace{-2ex}
\caption{Acc, SMV, Corr, and HOS of $\HM^{(k)}$ in GIN.} \label{fig:over-smoothing-GIN}
\vspace{0ex}
\end{figure}

Similar to Fig.~\ref{fig:over-smoothing}, Fig.~\ref{fig:over-smoothing-GAT} and Fig.~\ref{fig:over-smoothing-GIN} illustrate the trends in classification accuracy (Acc), SMV, HOS, and Corr across layers for \texttt{GAT} and \texttt{GIN} on the homophilic graphs {\em CiteSeer}, {\em CS}, {\em WikiCS}, and heterophilic graph {\em Cornell}, {\em Wisconsin}, {\em Chameleon}.
The results are consistent with those observed for \texttt{GCN}: as the number of layers increases, Acc, SMV, and HOS decline, while Corr increases.
Notably, the gap between HOS and SMV is evident even in the early layers, emphasizing the challenge of distinguishing nodes with different labels when embeddings become overly similar. These observations suggest that the phenomena of over-smoothing and over-correlation are not unique to \texttt{GCN} but are shared across different GNN architectures like \texttt{GAT} and \texttt{GIN}.

\section{Properties of Diffusion Distances and Others}\label{sec:distance-properties}

\begin{table}[hb]
\centering
\caption{Comparison with existing graph-theoretic distance metrics (for node pairs in $\EDG$).}\label{tbl:distances}
\vspace{-1ex}
\begin{small}
\begin{tabular}{l|ccccccc}
\toprule
 & SPD & Jaccard & Resistance & Biharmonic & VDD & PRDD & HKDD\\
 & ~\cite{dijkstra1959note} & \cite{kosub2019note} & \cite{Klein1993} & \cite{lipman2010biharmonic} & \cite{nadler2005diffusion} & - & \cite{de2008hierarchical} \\
\midrule
Non-negativity & \cmark &  \cmark & \cmark & \cmark & \cmark & \cmark & \cmark \\
Finity & \xmark &  \cmark & \xmark & \xmark & \cmark & \cmark & \cmark \\
Symmetry & \cmark &  \cmark & \cmark & \cmark & \cmark & \cmark & \cmark \\
Triangle Inequality & \cmark &  \cmark & \cmark & \cmark & \cmark & \cmark & \cmark \\
Edge Weights & \cmark &  \xmark & \cmark & \cmark & \cmark & \cmark & \cmark \\
Local Structure & \cmark &  \cmark & \cmark & \cmark & \cmark & \cmark & \cmark \\
Global Structure & \xmark & \xmark  & \cmark & \cmark & \cmark & \cmark & \cmark \\ 
Structural Similarity & \xmark & \xmark  & \xmark & \xmark & \cmark & \cmark & \cmark \\ \midrule
Range (Directed graphs) & \(\{1,+\infty\}\) & \([\frac{2}{n},1]\)  & \((0,+\infty]\) & 
undefined & \([0,\frac{\sqrt{2}}{d_{\min}}]\) & \([0,\frac{\sqrt{2}}{(1-\gamma)d_{\min}}]\) & \([0,\sqrt{2}]\) \\
Range (Undirected graphs) & \(\{1\}\) & \([\frac{2}{n},1]\)  & \((0,1]\) & \([\frac{\sqrt{2}}{2},\sqrt{2}nD]
\) & \([0,\frac{\sqrt{2}}{d_{\min}}]\) & \([0,\frac{\sqrt{2}}{(1-\gamma)d_{\min}}]\) & \([0,\sqrt{2}]\)\\
\bottomrule
\end{tabular}
\end{small}

{\textsuperscript{*}\scriptsize{$D$ is the diameter of $\G$ and $d_{\min}:=\min_{v_i\in \V}{d_i}$.}\hfill}
\vspace{-1ex}
\end{table}

We evaluate graph-theoretic distance metrics based on the following properties:
\begin{itemize}[leftmargin=*]
\item \textbf{Non-negativity and finity}: \(\delta_{i,j} \in [0, +\infty),\ \forall v_i, v_j \in \V\);
\item \textbf{Symmetry}: \(\delta_{i,j} = \delta_{j,i},\ \forall v_i, v_j \in \V\);
\item \textbf{Triangle Inequality}: \( \delta_{i,j} \leq \delta_{i,k} + \delta_{k,j},\ \forall v_i, v_j, v_k \in \V\);
\item \textbf{Edge Weights}: the distance \(\delta_{i,j}\) between any two nodes $v_i,v_j$ needs to account for the importance weights of the edges or paths connecting them;
\item  \textbf{Local and Global Structure}: the local structure of a node $v_i$ refers to the direct neighborhood information of $v_i$, while the global structure of $v_i$ considers the diverse paths pertinent to $v_i$ from the perspective of the entire graph~\cite{lu2011link}.
\item  \textbf{Structural Similarity}: unlike connectivity-based proximity metrics, the structural similarity of two nodes measures if two nodes have similar roles in the graph structure, i.e., similar connectivity patterns with others, regardless of whether they are distant or not. Common ways leverage the random walk distributions over the graph to measure the structural similarity~\cite{donnat2018learning}.
\end{itemize}

The comparative analysis in Table~\ref{tbl:distances} shows that classic measures including shortest path distance (SPD), resistance distance, and biharmonic distance exhibit significant limitations, particularly in finiteness, edge weights, global structural integration, and structural similarity evaluation. However, diffusion distances including VDD, PRDD, and HKDD address these issues and meet all the outlined criteria.

\begin{proposition}
For any $(v_i,v_j)\in \EDG$, the Jaccard distance $\delta_{i,j}\in [\frac{2}{n},1]$ when $G$ is either directed or undirected.
\begin{proof}
The definition of Jaccard distance $\delta_{i,j}$ is defined as follows:
\begin{equation*}
\delta_{i,j} = 1-\frac{|\N(v_i)\cap\N(v_i)|}{|\N(v_i)\cup\N(v_i)|}.
\end{equation*}
In a directed graph, $\N(v_i)$ and $\N(v_j)$ usually refer to the outgoing neighbors of $v_i$ and $v_j$, respectively. In the worst case, $v_i$ and $v_j$ have no common out-neighbors in a directed graph or no common neighbors in an undirected graph, even if they form an edge, i.e., $|\N(v_i)\cap\N(v_i)|=0$. Thus, the upper bound for $\delta_{i,j}$ is $1$.

Note that the maximum number of common neighbors of $v_i$ and $v_j$, i.e., $|\N(v_i)\cap\N(v_i)|$, is $n-2$, in which $|\N(v_i)\cup\N(v_i)|=n$, which leads to the lower bound for $\delta_{i,j}$.
\end{proof}
\end{proposition}

\begin{proposition}
For any $(v_i,v_j)\in \EDG$, the resistance distance $\delta_{i,j}$ is in the ranges of $(0,+\infty]$ and $(0,1]$ when $G$ is directed and undirected, respectively.
\begin{proof}
According to the connection between resistance distance and commute time~\cite{chandra1989electrical}, the resistance distance $r(v_i,v_j)$ for each edge $(v_i,v_j)\in \EDG$ can be represented by $\frac{c(v_i,v_j)}{2m}$, where $c(v_i,v_j)$ denotes the commute time between $v_i$ and $v_j$. In a directed graph, the commute time can be up to $+\infty$ if there are no directed paths from $v_i$ coming back to $v_i$ through $v_j$, and thus, $r(v_i,v_j)\le +\infty$. In an undirected graph, the commute time of an edge $(v_i,v_j)$ is up to $2m$, leading to $r(v_i,v_j)\le 1$.
\end{proof}
\end{proposition}

\begin{proposition}
Assume that $\G$ is an undirected connected graph. The Biharmonic distance of any two nodes $v_i,v_j\in \V$ is bounded in $[\frac{\sqrt{2}}{2},\sqrt{2}nD]$, where $D$ is the diameter of $\G$.
\begin{proof}
Let $\Delta_{\text{Biharmonic}}(v_i,v_j)$ be the Biharmonic distance between nodes $v_i$ and $v_j$, $\lambda_n$ and $\lambda_2$ be the largest and second smallest eigenvalue of $\LM$. According to Theorem 3.2 in \cite{wei2021biharmonic}, 
\begin{equation*}
\frac{\sqrt{{2}}}{\lambda_n} \le \Delta_{\text{Biharmonic}}(v_i,v_j) \le \frac{\sqrt{{2}}}{\lambda_2}.
\end{equation*}
Since the eigenvalues of $\LM$ are bounded in $[0,2]$~\cite{chung1997spectral}, $\lambda_n=2$. The second smallest eigenvalue of $\LM$ is also known as the {\em algebraic connectivity} of $\G$, which is lower bounded by $\frac{1}{nD}$ \cite{gross2003handbook}. Then, the lemma follows.
\end{proof}
\end{proposition}

\section{Derivative Steps for Eq.~\eqref{eq:obj}}\label{sec:ours-details}

Denote the first term as $\mathcal{L}_{\textnormal{DD}} = \sum_{(v_i,v_j)\in \EDG} \left( \left( \left\| \frac{\HM_i}{\sqrt{d_i}} - \frac{\HM_j}{\sqrt{d_j}} \right\|_2 - \eta\cdot \Delta(v_i,v_j) \right)^2 \right)$. Similar to \cite{cui2023mgnn}, taking the derivative of $\mathcal{L}_{\textnormal{DD}}$ w.r.t. $\HM_i\ \forall{v_i\in \V}$ yields
\begin{align*}
&\frac{\partial\mathcal{L}_{\textnormal{DD}}}{\partial \HM_i} \\
&=\frac{\partial}{\partial \HM_i} \sum_{v_j\in \N(v_i)} \left( \left( \left\| \frac{\HM_i}{\sqrt{d_i}} - \frac{\HM_j}{\sqrt{d_j}} \right\|_2 - \eta\cdot \Delta(v_i,v_j) \right)^2 \right) \\
&= 2 \sum_{v_j\in \N(v_i)} \left( \left\| \frac{\HM_i}{\sqrt{d_i}} - \frac{\HM_j}{\sqrt{d_j}} \right\|_2 - \eta\cdot \Delta(v_i,v_j) \right)  \frac{1}{\left\|\frac{\HM_i}{\sqrt{d_i}} - \frac{\HM_j}{\sqrt{d_j}}\right\|_2} \left( \frac{\HM_i}{d_i} - \frac{\HM_j}{\sqrt{d_i} \sqrt{d_j}} \right) \\
&= 2 \sum_{v_j\in \N(v_i)} \left(1 - \frac{\eta\cdot \Delta(v_i,v_j)}{\left\|\frac{\HM_i}{\sqrt{d_i}} - \frac{\HM_j}{\sqrt{d_j}}\right\|_2}\right) \left( \frac{\HM_i}{d_i} - \frac{\HM_j}{\sqrt{d_i} \sqrt{d_j}} \right) \\
&= 2 \left(\HM_i -  \sum_{v_j\in \N(v_i)} \frac{\HM_j}{\sqrt{d_i} \sqrt{d_j}} -  \eta \sum_{v_j\in \N(v_i)}\frac{\Delta(v_i,v_j)}{\left\|\frac{\HM_i}{\sqrt{d_i}} - \frac{\HM_j}{\sqrt{d_j}}\right\|_2} \left( \frac{\HM_i}{d_i} - \frac{\HM_j}{\sqrt{d_i} \sqrt{d_j}} \right) \right).
\end{align*}

To get the optimal solution, we calculate the node-wise gradient of Eq.~\eqref{eq:obj} and obtain
\begin{equation}\label{eq:loss-gradient}
\frac{\partial \mathcal{L}}{\partial\HM_i} = 
(1 - \alpha - \beta)\frac{\partial\mathcal{L}_{\textnormal{DD}}}{\partial \HM_i}
+ 2 \alpha (\HM_i - \XM_i)
+ 4 (\frac{1}{2} \beta (\HM_i - (\HM\HM^{\top}\HM)_i))
\end{equation}
We set Eq.~\eqref{eq:loss-gradient} to zero and get the propagation rules as
\begin{align*}
\HM_i^{(k+1)} = & (1 - \alpha - \beta) \left(\sum_{v_j \in \N(v_i)} \frac{\HM_j^{(k)}}{\sqrt{d_i d_j}} \right) + \eta \sum_{v_j \in \N(v_i)} \frac{\Delta(v_i,v_j)}
{\left\| \frac{\HM_i^{(k)}}{\sqrt{d_i}} - \frac{\HM_j^{(k)}}{\sqrt{d_j}} \right\|_2}
\left( \frac{\HM_i^{(k)}}{d_i} - \frac{\HM_j^{(k)}}{\sqrt{d_i d_j} }\right)\\
& + \alpha \HM^{(0)}_i + \beta (\HM^{(k)}\HM^{^{(k)}\top}\HM^{(k)})_i.
\end{align*}

\section{Experimental Details}\label{sec:exp-detail}
\begin{table}[!b]
\centering
\caption{Hyperparameters for \algp.}
\vspace{-1ex}
\begin{small}
\begin{tabular}{c|ccccccccccc}
\toprule
 & {\em Cora} & {\em Citeseer} & {\em Pubmed} & {\em CoraFull} 
 & {\em CS} & {\em Physics} & {\em Cornell} & {\em Texas} & {\em Wisconsin}
 & {\em Chameleon} & {\em WikiCS}  \\
\midrule
lr              & 0.01 & 2e-4 & 0.05 & 5e-4 & 2e-4 & 0.005 & 0.02 & 0.01 & 0.01 & 0.002 & 0.01  \\
wd              & 0.001 & 0.01 & 5e-5 & 1e-4 & 1e-4 & 1e-5 & 1e-4 & 0.05 & 0.01 & 1e-4 & 1e-4  \\
dropout         & 0.7 & 0.5 & 0.4 & 0.3 & 0.7 & 0.2 & 0.6 & 0.5 & 0.7 & 0.2 & 0.3  \\
\# layers        & 7 & 9 & 9 & 4 & 10 & 8 & 8 & 2 & 4 & 1 & 2  \\
$d$             & 64 & 64 & 64 & 64 & 64 & 64 & 64 & 64 & 64 & 64 & 64  \\
$\alpha$        & 0.161 & 0.247 & 0.295 & 0.199 & 0.995 & 0.963 & 0.956 & 0.627 & 0.984 & 0.001 & 0.150  \\
$\beta$         & 0.001 & 0.019 & 0.018 & 0.003 & 0.472 & 0.787 & 0.001 & 0.121 & 0.001 & 0.038 & 0.431  \\
$\eta$          & 0.101 & 0.407 & 0.740 & 0.160 & 0.456 & 0.092 & 0.772 & 0.017 & 0.201 & 0.239 & 0.239  \\
$\kappa$        & 64 & 32 & 128 & 64 & 64 & 64 & 64 & 64 & 64 & 64 & 64  \\
$\gamma$        & 0.9 & 0.9 & 0.9 & 0.9 & 0.9 & 0.9 & 0.9 & 0.9 & 0.9 & 0.9 & 0.9  \\
\bottomrule
\end{tabular}
\end{small}
\label{tbl:combined_param_selection}
\end{table}

\begin{table}[!b]
\centering
\caption{Hyperparameters for \algv.}
\vspace{-1ex}
\begin{small}
\begin{tabular}{c|ccccccccccc}
\toprule
 & {\em Cora} & {\em Citeseer} & {\em Pubmed} & {\em CoraFull} 
 & {\em CS} & {\em Physics} & {\em Cornell} & {\em Texas} & {\em Wisconsin}
 & {\em Chameleon} & {\em WikiCS}  \\
\midrule
lr              & 0.002 & 1e-4 & 0.05 & 5e-4 & 2e-4 & 0.02 & 0.005 & 0.02 & 0.01 & 0.002 & 0.002  \\
wd              & 0.005 & 0.01 & 5e-5 & 1e-4 & 1e-4 & 5e-5 & 0.001 & 0.05 & 0.01 & 1e-4 & 5e-5  \\
dropout         & 0.6 & 0.6 & 0.2 & 0.3 & 0.6 & 0.2 & 0.7 & 0.2 & 0.6 & 0.5 & 0.2  \\
\# layers        & 7 & 9 & 7 & 4 & 10 & 8 & 8 & 2 & 8 & 1 & 2  \\
$d$     & 64 & 64 & 64 & 64 & 64 & 64 & 64 & 64 & 64 & 64 & 64  \\
$\alpha$        & 0.289 & 0.308 & 0.355 & 0.189 & 0.741 & 0.694 & 0.999 & 1.000 & 1.000 & 0.013 & 0.219  \\
$\beta$         & 0.036 & 0.048 & 0.001 & 0.002 & 0.193 & 0.642 & 0.001 & 0.155 & 0.037 & 0.001 & 0.179  \\
$\eta$          & 0.657 & 0.800 & 0.001 & 0.998 & 0.961 & 0.390 & 0.851 & 0.604 & 0.860 & 0.997 & 0.683  \\
$\kappa$        & 64 & 64 & 128 & 64 & 64 & 32 & 64 & 64 & 64 & 64 & 32  \\
$t$             & 10 & 10 & 10 & 10 & 10 & 10 & 10 & 10 & 10 & 10 & 10  \\
\bottomrule
\end{tabular}
\end{small}
\label{tbl:combined_param_selection-vanilla}
\end{table}

\begin{table}[!b]
\centering
\caption{Hyperparameters for \algh.}
\vspace{-1ex}
\begin{small}
\begin{tabular}{c|ccccccccccc}
\toprule
 & {\em Cora} & {\em Citeseer} & {\em Pubmed} & {\em CoraFull} 
 & {\em CS} & {\em Physics} & {\em Cornell} & {\em Texas} & {\em Wisconsin}
 & {\em Chameleon} & {\em WikiCS}  \\
\midrule
lr              & 0.005 & 2e-4 & 0.05 & 0.001 & 1e-4 & 0.01 & 0.005 & 0.01 & 0.02 & 0.01 & 0.002  \\
wd              & 0.001 & 0.01 & 1e-4 & 0.001 & 5e-4 & 5e-4 & 5e-4 & 0.05 & 0.01 & 5e-5 & 1e-4  \\
dropout         & 0.7 & 0.5 & 0.3 & 0.1 & 0.1 & 0.5 & 0.6 & 0.2 & 0.7 & 0.6 & 0.1  \\
\# layers        & 7 & 9 & 7 & 9 & 1 & 8 & 8 & 2 & 8 & 1 & 2  \\
$d$             & 64 & 64 & 64 & 64 & 64 & 64 & 64 & 64 & 64 & 64 & 64  \\
$\alpha$        & 0.146 & 0.246 & 0.381 & 0.449 & 0.749 & 0.715 & 0.999 & 0.591 & 0.968 & 0.020 & 0.282  \\
$\beta$         & 0.034 & 0.042 & 0.001 & 0.193 & 0.087 & 0.519 & 0.001 & 0.045 & 0.037 & 0.019 & 0.918  \\
$\eta$          & 0.433 & 0.945 & 1.000 & 0.962 & 0.865 & 0.415 & 0.126 & 0.305 & 0.335 & 0.913 & 0.500  \\
$\kappa$        & 64 & $n$ & 64 & 64 & 64 & 64 & 64 & $n$ & 64 & 64 & 16  \\
$\gamma$        & 10 & 10 & 10 & 10 & 10 & 10 & 10 & 10 & 10 & 10 & 10  \\
\bottomrule
\end{tabular}
\end{small}
\label{tbl:combined_param_selection-hk}
\end{table}

\subsection{Hyperparameter Settings}
The sets of hyperparameters used in \algp, \algv, and \algh are listed in Table~\ref{tbl:combined_param_selection},~\ref{tbl:combined_param_selection-vanilla} and~\ref{tbl:combined_param_selection-hk}, respectively.
For all methods and datasets, the hidden dimensionality $d$ is consistently set to 64. In \algp, the parameter $\gamma$ is set to $0.9$ across all datasets, while in \algv, $t$ is set to 10 for every dataset. Similarly, in \algh, $\gamma$ is set to 10. For the hyperparameters $\alpha$, $\beta$, and $\eta$, we run 100 trials for each method on each dataset using the Optuna package. The remaining hyperparameters are optimized through grid search to determine the best configuration, with their ranges specified below.
\begin{itemize}[leftmargin=*]
\item learning rate (lr): \{1e-1, 5e-2, 2e-2, 1e-2, 5e-3, 2e-3, 1e-3, 5e-4, 2e-4, 1e-4, 5e-5, 2e-5, 1e-5\}.
\item weight decay (wd): \{1e-1, 5e-2, 1e-2, 5e-3, 1e-3, 5e-4, 1e-4, 5e-5, 1e-5, 0\}.
\item dropout: \{0, 0.1, 0.2, 0.3, 0.4, 0.5, 0.6, 0.7, 0.8, 0.9\}.
\item number of layers: \{1, 2, 3, 4, 5, 6, 7, 8, 9, 10\}.
\item \( \alpha \): (0, 1).
\item \( \beta \): (0, 1).
\item \( \eta \): (0, 1).
\item \( \kappa \): \{16, 32, 68, 128, 256, \( n \)\}.

\end{itemize}

\section{Additional Experiments}\label{sec:add-exp}
\subsection{Efficiency Analysis of VDD Computation}\label{sec:time}
To provide a comprehensive overview of the computational costs associated with \algv, we present a detailed breakdown of the time required for both VDD computation and model training across a variety of datasets. As illustrated in Table~\ref{tab:computation_time_comparison}, the VDD computation is remarkably efficient across all benchmark datasets. For instance, on smaller graphs such as \textit{Cora} and \textit{CiteSeer}, the VDD computation is completed in a mere 0.1 seconds, which is negligible compared to the 9.3 and 12.6 seconds required for model training, respectively. Even for medium-sized graphs like \textit{Physics} and \textit{WikiCS}, the VDD computation time remains exceptionally low, clocking in at 1.3 and 0.3 seconds, respectively. This is substantially less than the corresponding model training times of 212.3 and 45.3 seconds. This consistent trend underscores that the VDD computation phase does not constitute a significant computational bottleneck.
\begin{table}[h]
\centering
\caption{Comparison of VDD Computation Time and Model Training Time}
\begin{small}
\begin{tabular}{lrr|rr}
\toprule
Dataset & \# Nodes & \# Edges & VDD Computation & Model Training \\
\midrule
\textit{Cora} & 2,708   & 10,556 & 0.1s & 9.3s \\
\textit{CiteSeer} & 3,327   & 9,104 & 0.1s & 12.6s \\
\textit{PubMed} & 19,717  & 88,648 & 0.4s & 29.6s \\
\textit{CoraFull} & 19,793  & 126,842 & 0.4s & 52.7s \\
\textit{CS} & 18,333  & 163,788 & 0.3s & 17.2s \\
\textit{Physics} & 34,493  & 495,924 & 1.3s & 212.3s \\
\textit{Cornell} & 183     & 557 & 0.04s & 7.9s \\
\textit{Texas} & 183     & 574 & 0.02s & 3.6s \\
\textit{Wisconsin} & 251     & 916 & 0.03s & 5.5s \\
\textit{Chameleon} & 2,277   & 62,792 & 0.07s & 4.2s \\
\textit{WikiCS} & 11,701  & 431,726 & 0.3s & 45.3s \\
\midrule
\textit{oogbn-arxiv} & 169,343 & 1,166,243 & 4.2s & 481.5s \\
\textit{opokec} & 1,632,803 & 30,622,564 & 76.4s & 8,667.3s \\
\textit{ogbn-products} & 2,449,029 & 61,859,140 & 231.8s & 25,975.7s \\
\bottomrule
\end{tabular}
\end{small}
\label{tab:computation_time_comparison}
\end{table}

\subsection{Scalability to Large Graphs}\label{sec:scalability}
We further evaluate the scalability of \algv on three large-scale datasets: ogbn-products and ogbn-arxiv from OGB~\cite{hu2020open}, along with the pokec dataset from~\cite{takac2012data}. As presented in Table~\ref{tab:computation_time_comparison}, the time costs for computing the diffusion distances are significantly lower than those required for model training. For instance, on ogbn-products with 2.4 million nodes, \alg can finish the VDD computation within merely 4 minutes, whereas training the model takes 433 minutes. This indicates that diffusion distance computation is not the major bottleneck in scaling \alg. On the contrary, our truncated spectral decomposition approach for distance approximation can be efficiently done on large graphs in practice due to the high sparsity of real graphs, the small dimension $\kappa$ (64 or 128) needed, and fast solvers (e.g., Lanczos and Arnoldi methods) for partial eigendecomposition.

As for the model training of \alg, similar to previous works, well-established sampling techniques such as neighbor sampling~\cite{hamilton2017inductive}, subgraph sampling~\cite{chiang2019cluster}, and mini-batching~\cite{lim2021large}, can be readily adopted to speed up the message passing in \alg. Since the focus of this work is not on the efficiency aspect, we will further scale \alg to extremely large graphs (e.g., with billions of nodes) and leave the exploration of faster and parallel algorithms for distance computation and graph sampling strategies for efficient training to future work.

 \begin{figure}[!t]
\centering
\begin{small}
\begin{tikzpicture}
    \begin{customlegend}[legend columns=3,
        legend entries={PRDD, Vanilla, HKDD},
        legend style={at={(0.45,1.35)},anchor=north,draw=none,font=\small,column sep=0.1cm}]
    \addlegendimage{line width=0.3mm,mark size=3pt,mark=triangle, color=PageRank-color}
    \addlegendimage{line width=0.3mm,mark size=3pt,mark=diamond, color=Vanilla-color}
    \addlegendimage{line width=0.3mm,mark size=3pt,mark=o, color=HeatKernel-color}
    \end{customlegend}
\end{tikzpicture}
\\[-\lineskip]
\vspace{-4mm}
\subfloat[Varying $\alpha$ on {\em Texas}]{
\begin{tikzpicture}[scale=1,every mark/.append style={mark size=2pt}]
    \begin{axis}[
        height=\columnwidth/4.8,
        width=\columnwidth/4.0,
        ylabel={\it Accuracy},
        xmin=0.5, xmax=9.5,
        ymin=74, ymax=98,
        xtick={1,2,3,4,5,6,7,8,9},
        ytick={74,80,86,92,98},
        xticklabel style = {font=\tiny},
        yticklabel style = {font=\scriptsize},
        xticklabels={0.1,,0.3,,0.5,,0.7,,0.9},
        yticklabels={74,80,86,92,98},
    ]
    \addplot[line width=0.3mm, mark=triangle, color=PageRank-color]  %
        plot coordinates {
(1,	75.14   )
(2,	78.92	)
(3,	83.78	)
(4,	89.73	)
(5,	93.78	)
(6,	95.14	)
(7,	95.14	)
(8,	90.54	)
(9,	91.89	)
    };

    \addplot[line width=0.3mm, mark=diamond, color=Vanilla-color]  %
        plot coordinates {
(1,	80.81	)
(2,	83.78	)
(3,	85.95	)
(4,	90.81	)
(5,	90.27	)
(6,	91.35	)
(7,	89.73	)
(8,	89.73	)
(9,	90.81	)
    };

    \addplot[line width=0.3mm, mark=o, color=HeatKernel-color]  %
        plot coordinates {
(1,	84.86	)
(2,	86.49	)
(3,	90.81	)
(4,	91.08	)
(5,	95.68	)
(6,	96.76   )
(7,	91.89   )
(8,	91.08	)
(9,	91.35	)
    };

    \end{axis}
\end{tikzpicture}\hspace{1mm}\label{fig:vary-alpha-texas}%
}
\subfloat[Varying $\beta$ on {\em Texas}]{
\begin{tikzpicture}[scale=1,every mark/.append style={mark size=2pt}]
    \begin{axis}[
        height=\columnwidth/4.8,
        width=\columnwidth/4.0,
        ylabel={\it Accuracy},
        xmin=0.5, xmax=9.5,
        ymin=74, ymax=98,
        xtick={1,2,3,4,5,6,7,8,9},
        ytick={74,80,86,92,98},
        xticklabel style = {font=\tiny},
        yticklabel style = {font=\scriptsize},
        xticklabels={0.1,,0.3,,0.5,,0.7,,0.9},
        yticklabels={74,80,86,92,98},
    ]
    \addplot[line width=0.3mm, mark=triangle, color=PageRank-color]  %
        plot coordinates {
(1,	94.59	)
(2,	94.32	)
(3,	91.08	)
(4,	87.57	)
(5,	82.16	)
(6,	83.51	)
(7,	82.43	)
(8,	81.89	)
(9,	84.32	)
    };

    \addplot[line width=0.3mm, mark=diamond, color=Vanilla-color]  %
        plot coordinates {
(1,	90	)
(2,	92.7	)
(3,	91.08	)
(4,	86.49	)
(5,	85.68	)
(6,	86.76	)
(7,	86.22	)
(8,	84.32	)
(9,	83.24	)
    };

    \addplot[line width=0.3mm, mark=o, color=HeatKernel-color]  %
        plot coordinates {
(1,	92.16	)
(2,	85.95	)
(3,	85.68	)
(4,	82.7	)
(5,	80.54	)
(6,	78.38	)
(7,	77.03	)
(8,	76.76	)
(9,	75.41	)
    };

    \end{axis}
\end{tikzpicture}\hspace{1mm}\label{fig:vary-beta-texas}%
}
\subfloat[Varying $\eta$ on {\em Texas}]{
\begin{tikzpicture}[scale=1,every mark/.append style={mark size=2pt}]
    \begin{axis}[
        height=\columnwidth/4.8,
        width=\columnwidth/4.0,
        ylabel={\it Accuracy},
        xmin=0.5, xmax=9.5,
        ymin=82, ymax=98,
        xtick={1,2,3,4,5,6,7,8,9},
        ytick={82,86,90,94,98},
        xticklabel style = {font=\tiny},
        yticklabel style = {font=\scriptsize},
        xticklabels={0.1,,0.3,,0.5,,0.7,,0.9},
        yticklabels={82,86,90,94,98},
    ]
    \addplot[line width=0.3mm, mark=triangle, color=PageRank-color]  %
        plot coordinates {
(1,	92.43	)
(2,	91.62	)
(3,	88.65	)
(4,	84.59	)
(5,	87.84	)
(6,	85.95	)
(7,	85.41	)
(8,	86.22	)
(9,	87.3	)
    };

    \addplot[line width=0.3mm, mark=diamond, color=Vanilla-color]  %
        plot coordinates {
(1,	90.81	)
(2,	91.08	)
(3,	91.35	)
(4,	91.35	)
(5,	91.89	)
(6,	91.08	)
(7,	91.35	)
(8,	91.89	)
(9,	91.62	)
    };

    \addplot[line width=0.3mm, mark=o, color=HeatKernel-color]  %
        plot coordinates {
(1,	96.76	)
(2,	96.49	)
(3,	96.76	)
(4,	97.03	)
(5,	96.76	)
(6,	96.22	)
(7,	95.95	)
(8,	94.86	)
(9,	97.03	)
    };

    \end{axis}
\end{tikzpicture}\hspace{1mm}\label{fig:vary-eta-texas}%
}
\subfloat[Varying $\kappa$ on {\em Texas}]{
\begin{tikzpicture}[scale=1,every mark/.append style={mark size=2pt}]
    \begin{axis}[
        height=\columnwidth/4.8,
        width=\columnwidth/4.0,
        ylabel={\it Accuracy},
        xmin=0.5, xmax=6.5,
        ymin=88, ymax=98,
        xtick={1,2,3,4,5,6},
        ytick={88,90,92,94,96,98},
        xticklabel style = {font=\tiny},
        yticklabel style = {font=\scriptsize},
        xticklabels={8,16,32,64,128,$n$},
        ytick={88,90,92,94,96,98},
    ]
    \addplot[line width=0.3mm, mark=triangle, color=PageRank-color]  %
        plot coordinates {
(1,	95.68	)
(2,	96.49	)
(3,	95.41	)
(4,	96.76	)
(5,	94.86	)
(6,	94.86	)
    };

    \addplot[line width=0.3mm, mark=diamond, color=Vanilla-color]  %
        plot coordinates {
(1,	92.7	)
(2,	90.81	)
(3,	90.81	)
(4,	93.24	)
(5,	92.16	)
(6,	92.16	)
    };

    \addplot[line width=0.3mm, mark=o, color=HeatKernel-color]  %
        plot coordinates {
(1, 97.03	)
(2, 96.49	)
(3, 97.03	)
(4, 96.76	)
(5, 95.14	)
(6, 97.03	)

    };

    \end{axis}
\end{tikzpicture}\hspace{1mm}\label{fig:vary-k-texas}%
}
\vspace{-4mm}
\subfloat[Varying $\alpha$ on {\em Chameleon}]{
\begin{tikzpicture}[scale=1,every mark/.append style={mark size=2pt}]
    \begin{axis}[
        height=\columnwidth/4.8,
        width=\columnwidth/4.0,
        ylabel={\it Accuracy},
        xmin=0.5, xmax=9.5,
        ymin=40, ymax=76,
        xtick={1,2,3,4,5,6,7,8,9},
        ytick={40,46,52,58,64,70,76},
        xticklabel style = {font=\tiny},
        yticklabel style = {font=\scriptsize},
        xticklabels={0.1,,0.3,,0.5,,0.7,,0.9},
        yticklabels={40,,52,,64,,76},
    ]
    \addplot[line width=0.3mm, mark=triangle, color=PageRank-color]  %
        plot coordinates {
(1,	64	)
(2,	50.75	)
(3,	45.65	)
(4,	44.75	)
(5,	42.97	)
(6,	43.19	)
(7,	43.8	)
(8,	44.55	)
(9,	45.03	)
    };

    \addplot[line width=0.3mm, mark=diamond, color=Vanilla-color]  %
        plot coordinates {
(1,	62.48	)
(2,	48.9	)
(3,	44.97	)
(4,	44.62	)
(5,	43.69	)
(6,	44.13	)
(7,	44.26	)
(8,	44.9	)
(9,	46.99	)
    };

    \addplot[line width=0.3mm, mark=o, color=HeatKernel-color]  %
        plot coordinates {
(1,	64.77	)
(2,	51.45	)
(3,	44.4	)
(4,	43.01	)
(5,	44.46	)
(6,	44.9	)
(7,	46.26	)
(8,	47.14	)
(9,	49.76	)
    };

    \end{axis}
\end{tikzpicture}\hspace{1mm}\label{fig:vary-alpha-chameleon}%
}
\subfloat[Varying $\beta$ on {\em Chameleon}]{
\begin{tikzpicture}[scale=1,every mark/.append style={mark size=2pt}]
    \begin{axis}[
        height=\columnwidth/4.8,
        width=\columnwidth/4.0,
        ylabel={\it Accuracy},
        xmin=0.5, xmax=9.5,
        ymin=40, ymax=76,
        xtick={1,2,3,4,5,6,7,8,9},
        ytick={40,46,52,58,64,70,76},
        xticklabel style = {font=\tiny},
        yticklabel style = {font=\scriptsize},
        xticklabels={0.1,,0.3,,0.5,,0.7,,0.9},
        yticklabels={40,,52,,64,,76},
    ]
    \addplot[line width=0.3mm, mark=triangle, color=PageRank-color]  %
        plot coordinates {
(1,	72.42	)
(2,	65.12	)
(3,	59.45	)
(4,	55.16	)
(5,	50.79	)
(6,	47.8	)
(7,	45.03	)
(8,	43.93	)
(9,	42.53	)
    };

    \addplot[line width=0.3mm, mark=diamond, color=Vanilla-color]  %
        plot coordinates {
(1,	72.11	)
(2,	63.78	)
(3,	57.54	)
(4,	52.62	)
(5,	50.97	)
(6,	47.47	)
(7,	44.97	)
(8,	44.46	)
(9,	43.71	)
    };

    \addplot[line width=0.3mm, mark=o, color=HeatKernel-color]  %
        plot coordinates {
(1,	73.76	)
(2,	71.01	)
(3,	65.93	)
(4,	61.93	)
(5,	58.22	)
(6,	56.37	)
(7,	55.01	)
(8,	53.1	)
(9,	52.59	)
    };

    \end{axis}
\end{tikzpicture}\hspace{1mm}\label{fig:vary-beta-chameleon}%
}
\subfloat[Varying $\eta$ on {\em Chameleon}]{
\begin{tikzpicture}[scale=1,every mark/.append style={mark size=2pt}]
    \begin{axis}[
        height=\columnwidth/4.8,
        width=\columnwidth/4.0,
        ylabel={\it Accuracy},
        xmin=0.5, xmax=9.5,
        ymin=72, ymax=76,
        xtick={1,2,3,4,5,6,7,8,9},
        ytick={72,73,74,75,76},
        xticklabel style = {font=\tiny},
        yticklabel style = {font=\scriptsize},
        xticklabels={0.1,,0.3,,0.5,,0.7,,0.9},
        yticklabels={72,73,74,75,76},
    ]
    \addplot[line width=0.3mm, mark=triangle, color=PageRank-color]  %
        plot coordinates {
(1,	74.31	)
(2,	75.16	)
(3,	75.03	)
(4,	74.35	)
(5,	73.87	)
(6,	73.27   )
(7,	72.99   )
(8,	73.21	)
(9,	72.95	)
    };

    \addplot[line width=0.3mm, mark=diamond, color=Vanilla-color]  %
        plot coordinates {
(1,	73.63	)
(2,	74.02	)
(3,	74.15	)
(4,	74.29	)
(5,	74.46	)
(6,	74.53	)
(7,	74.57	)
(8,	74.59	)
(9,	74.57	)
    };

    \addplot[line width=0.3mm, mark=o, color=HeatKernel-color]  %
        plot coordinates {
(1,	74.35	)
(2,	74.22	)
(3,	74.57	)
(4,	74.57	)
(5,	74.48	)
(6,	74.86	)
(7,	74.77	)
(8,	74.62	)
(9,	74.84	)
    };

    \end{axis}
\end{tikzpicture}\hspace{1mm}\label{fig:vary-eta-chameleon}%
}
\subfloat[Varying $\kappa$ on {\em Chameleon}]{
\begin{tikzpicture}[scale=1,every mark/.append style={mark size=2pt}]
    \begin{axis}[
        height=\columnwidth/4.8,
        width=\columnwidth/4.0,
        ylabel={\it Accuracy},
        xmin=0.5, xmax=7.5,
        ymin=71, ymax=76,
        xtick={1,2,3,4,5,6,7},
        ytick={71,72,73,74,75,76},
        xticklabel style = {font=\tiny},
        yticklabel style = {font=\scriptsize},
        xticklabels={8,16,32,64,128,256,$n$},
        yticklabels={71,72,73,74,75,76},
    ]
    \addplot[line width=0.3mm, mark=triangle, color=PageRank-color]  %
        plot coordinates {
(1,	73.14	)
(2,	73.19	)
(3,	75.03	)
(4,	75.23	)
(5,	74.07	)
(6,	72.9	)
(7,	71.89	)
    };

    \addplot[line width=0.3mm, mark=diamond, color=Vanilla-color]  %
        plot coordinates {
(1,	73.87	)
(2,	74.46	)
(3,	74.44	)
(4,	75.1	)
(5,	74.88	)
(6,	74.77	)
(7,	74.7	)
    };

    \addplot[line width=0.3mm, mark=o, color=HeatKernel-color]  %
        plot coordinates {
(1,	74.7	)
(2,	74.48	)
(3,	74.9	)
(4,	75.01	)
(5,	74.4	)
(6,	74.79	)
(7, 74.99   )

    };

    \end{axis}
\end{tikzpicture}\hspace{1mm}\label{fig:vary-k-chameleon}%
}
\end{small}
 \vspace{-2ex}
\caption{Accuracy when varying parameters.} \label{fig:parameter-appendix}
\vspace{-3ex}
\end{figure}
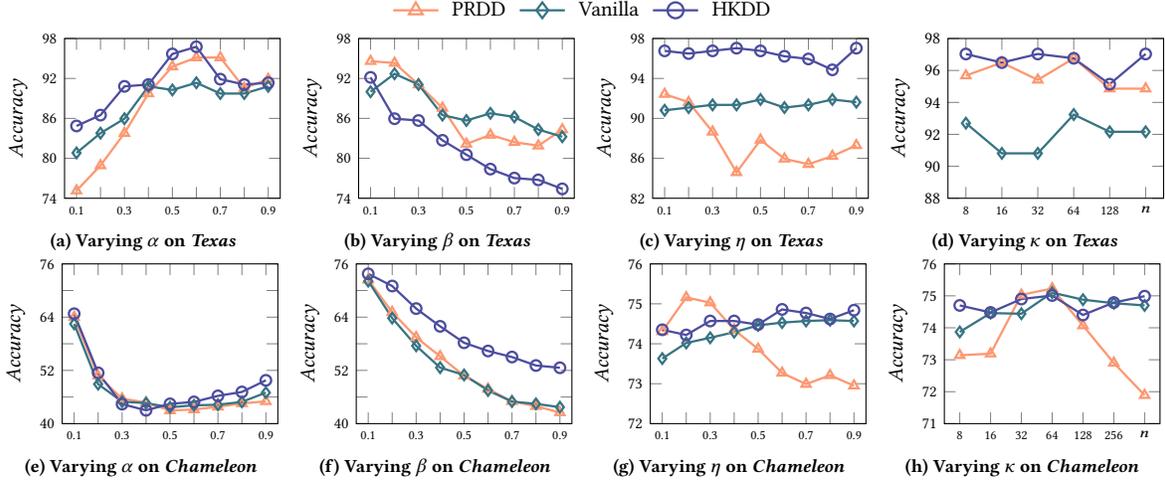

\subsection{Additional Ablation Study}
This section analyzes the effects of varying the coefficients $\alpha$, $\beta$, and $\eta$ in message functions (Eq.~\eqref{eq:propagation-rule}) of \alg, along with the dimension $\kappa$ used for diffusion distance approximation, on the \textit{Texas} and \textit{Chameleon} datasets.

Fig.~\ref{fig:vary-alpha-texas} and~\ref{fig:vary-alpha-chameleon} illustrate how the accuracy of \alg changes with \( \alpha \) ranging from 0.1 to 0.9. On the \textit{Texas} dataset, accuracy peaks for all methods at 0.6. Accuracy initially increases but declines for larger \( \alpha \), suggesting the optimal parameter range for residual connection lies around 0.6. On the \textit{Chameleon} dataset, accuracies drop sharply from 0.1 to 0.5, followed by slight recoveries, indicating that smaller \( \alpha \) values are more beneficial for residual connections in \textit{Chameleon}.

Fig.~\ref{fig:vary-beta-texas} and~\ref{fig:vary-beta-chameleon} depict the accuracy trends as \( \beta \) is adjusted. For both datasets, all \alg variants achieve their highest accuracy at 0.1 or 0.2, followed by a gradual decline. These results indicate that smaller \( \beta \) values are optimal for feature decorrelation in these settings.

Fig.~\ref{fig:vary-eta-texas} and~\ref{fig:vary-eta-chameleon} examine the performance of \alg for \( \eta \) values between 0.1 and 0.9. On both datasets, \algv and \algh maintain stable performance, with small variations, indicating robustness for this parameter range. \algp, however, shows a steady decline in performance.

Lastly, the impact of \( \kappa \) is shown in Fig.~\ref{fig:vary-k-texas} and~\ref{fig:vary-k-chameleon}. Across both datasets, smaller dimensions, such as $32$ or $64$, suffice for approximating VDD, HKDD, and PRDD while enabling \alg to learn effective predictive embeddings.

\section{Theoretical Proofs}\label{sec:proofs}

\eat{
\begin{proof}[\bf Proof of Lemma~\ref{lem:unified-propagation}]
By setting its derivative w.r.t. $\HM$ to zero, we obtain the optimal $\HM$ as:
\begin{align}
& \frac{\partial{\{\alpha\cdot\|\HM - f_{\boldsymbol{\theta}}(\XM)\|^2_F + (1-\alpha) \cdot trace(\HM^{\top}(\IM-\NAM)\HM)\}}}{\partial{\HM}}=0 \notag\\
& \Longrightarrow \alpha(\HM - f_{\boldsymbol{\theta}}(\XM)) + (1-\alpha)\cdot (\IM-\NAM)\HM = 0 \notag\\
& \Longrightarrow \HM = \alpha\cdot \left(\IM-(1-\alpha)\NAM\right)^{-1} f_{\boldsymbol{\theta}}(\XM). \label{eq:Z-derivative}
\end{align}
By the definition of Neumann series, i.e., $(\IM-\MM)^{-1}=\sum_{t=0}^{\infty}{\MM^t}$, we have 
$$(\IM - (1-\alpha) \NAM)^{-1} = \sum_{\ell=0}^{\infty}{(1-\alpha)^\ell \NAM^{\ell}}.$$
Hence, $\HM = \alpha\sum_{\ell=0}^{\infty}{(1-\alpha)^\ell\NAM^{\ell}} f_{\boldsymbol{\theta}}(\XM)$, which completes the proof.
\end{proof}
}

\begin{proof}[\bf Proof of Lemma~\ref{lem:dd-range}]
Recall that $\PM$ is row-stochastic. Then, $\PM^t$ in \(\Delta_{\text{vanilla}}(v_i, v_j)\) and \(\sum_{t=0}^{\infty}{\frac{e^{-\gamma}\gamma^t}{t!}\PM^t}\) in \(\Delta_{\text{HK}}(v_i, v_j)\) are both row-stochastic. Each row in \(\sum_{t=0}^{\infty}{\gamma^t\PM^t}\) in \(\Delta_{\text{PR}}(v_i, v_j)\) satisfies
\begin{equation*}
\sum_{v_j\in \V}\sum_{t=0}^{\infty}{\gamma^t\PM^t}_{i,j} = \sum_{t=0}^{\infty}{\gamma^t}\cdot \sum_{v_j\in \V}\PM^t_{i,j} = \sum_{t=0}^{\infty}{\gamma^t} = \frac{1}{1-\gamma}.
\end{equation*}

With slight abuse of notations, we use \(\pi_{i,\ell}\) represent $\PM^t_{i,\ell}$, \(\sum_{t=0}^{\infty}{\frac{e^{-\gamma}\gamma^t}{t!}\PM^t_{i,\ell}}\), and \(\sum_{t=0}^{\infty}{\gamma^t\PM^t_{i,\ell}}\) in VDD, HKDD, and PRDD, respectively. Accordingly,
\begin{align*}
\Delta_{\text{vanilla}}(v_i, v_j) &= \|(\PM^t\DM^{-1/2})_i-(\PM^t\DM^{-1/2})_j\|_2 = \sqrt{\sum_{v_\ell\in \V}{(\frac{\pi_{i,\ell}}{\sqrt{d_\ell}}-\frac{\pi_{j,\ell}}{\sqrt{d_\ell}})^2}} \\
& \le \frac{1}{d_{\min}}\cdot \sqrt{\sum_{v_\ell}{(\pi_{i,\ell}^2+\pi_{j,\ell}^2-2\pi_{i,\ell}\pi_{j,\ell})}} \\
& = \frac{1}{d_{\min}}\cdot \sqrt{\sum_{v_\ell}{\pi_{i,\ell}^2}+\sum_{v_\ell}\pi_{j,\ell}^2-2\sum_{v_\ell}\pi_{i,\ell}\pi_{j,\ell}}.
\end{align*}
In the same vein, we have
\begin{equation*}
\Delta_{\text{HK}}(v_i, v_j) \le \sqrt{\sum_{v_\ell}{\pi_{i,\ell}^2}+\sum_{v_\ell}\pi_{j,\ell}^2-2\sum_{v_\ell}\pi_{i,\ell}\pi_{j,\ell}},
\end{equation*}
and
\begin{equation*}
\Delta_{\text{PR}}(v_i, v_j) \le \frac{1}{d_{\min}}\cdot \sqrt{\sum_{v_\ell}{\pi_{i,\ell}^2}+\sum_{v_\ell}\pi_{j,\ell}^2-2\sum_{v_\ell}\pi_{i,\ell}\pi_{j,\ell}}.
\end{equation*}
Since $\sum_{v_\ell}{\pi_{i,\ell}}=\sum_{v_\ell}{\pi_{j,\ell}}=1$ in VDD and HKDD, and equal $1/(1-\gamma)$ in PRDD, we can derive $\sum_{v_\ell}{\pi_{i,\ell}^2}=\sum_{v_\ell}{\pi_{j,\ell}^2}\le 1$ in VDD and HKDD, and $\sum_{v_\ell}{\pi_{i,\ell}^2}=\sum_{v_\ell}{\pi_{j,\ell}^2}\le 1/(1-\gamma)$ in PRDD. The lemma then naturally follows.
\end{proof}

\begin{proof}[\bf Proof of Lemma~\ref{lem:oversmooth}]
We first apply spectral decomposition on ${\NAM}$. The normalized adjacency matrix ${\NAM}$ is a real symmetric matrix. Therefore, it admits an eigendecomposition:
\[
{\NAM} = \sum_{i=1}^n \lambda_i \mathbf{u}_i \mathbf{u}_i^\top,
\]
where $\{\mathbf{u}_i\}_{i=1}^n$ are the corresponding orthonormal eigenvectors. As $\G$ is non-bipartite, it follows that the Frobenius-Perron Theorem that $\lambda_1 = 1 > \lambda_2 \geq \lambda_3 \geq \dots \geq \lambda_n > -1$ and that $\mathbf{u}_{1i}=\sqrt{\frac{d_i}{2m}} = \boldsymbol{\pi}(i)$~\cite{lovasz1993random}.

Raising ${\NAM}$ to the $k$-th power yields:
\[
{\NAM}^k = \sum_{i=1}^n \lambda_i^k \mathbf{u}_i \mathbf{u}_i^\top.
\]
As $k \to \infty$, since $|\lambda_i| < 1$ for all $i \geq 2$, we have $\lambda_i^k \to 0$. Therefore,
\[
\lim_{k \to \infty} {\NAM}^k = \lambda_1^k \mathbf{u}_1 \mathbf{u}_1^\top = \mathbf{u}_1 \mathbf{u}_1^\top,
\]
since $\lambda_1 = 1$.

Substituting $\mathbf{u}_1$ into the limit, we get:
\[
\lim_{k \to \infty} {\NAM}^k = \boldsymbol{\pi} \boldsymbol{\pi}^\top.
\]
\end{proof}

\begin{proof}[\bf Proof of Lemma~\ref{lem:homophily-DE}]
Recall that the definition of homophily ratio over graph $\G$ is the fraction of edges whose endpoints are in the same class. Thus,
\begin{align*}
h(\G) & = \frac{\sum_{(v_i,v_j)\in \EDG}{\sum_{k=1}^{|\Y|}\YM_{i,k}\cdot \YM_{j,k}}}{m}\\
& = - \frac{\sum_{(v_i,v_j)\in \EDG}{\sum_{k=1}^{|\Y|}{\YM_{i,k}^2-\YM_{i,k}^2+\YM_{j,k}^2-\YM_{j,k}^2-2\YM_{i,k}\cdot \YM_{j,k}}}}{2m} \\
& = \frac{\sum_{(v_i,v_j)\in \EDG}{\sum_{k=1}^{|\Y|}{\YM_{i,k}^2+\YM_{j,k}^2}}}{2m} - \frac{\sum_{(v_i,v_j)\in \EDG}{\sum_{k=1}^{K}{\YM_{i,k}^2+\YM_{j,k}^2-2\YM_{i,k}\cdot \YM_{j,k}}}}{2m} \\
& = 1 - \frac{\sum_{(v_i,v_j)\in \EDG}{\sum_{k=1}^{|\Y|}{(\YM_{i,k}-\YM_{j,k})^2}}}{2m}\\
& = 1 - \frac{\sum_{(v_i,v_j)\in \EDG}{\|\YM_i-\YM_j\|_2^2}}{2m} = 1 - \frac{\sum_{(v_i,v_j)\in \EDG}{\left\|\frac{(\DM^{1/2}\YM)_i}{\sqrt{d_i}}-\frac{(\DM^{1/2}\YM)_j}{\sqrt{d_j}}\right\|_2^2}}{2m} \\
& = 1 - \frac{\mathcal{D}(\DM^{1/2}\YM)}{m},
\end{align*}
which finishes the proof.
\end{proof}

\begin{proof}[\bf Proof of Lemma~\ref{thm:overcorrelation}]
Consider the propagated features after $k$ layers:
\[
\mathbf{Y} = {\NAM}^k \XM.
\]
We are interested in the feature correlation matrix:
\[
\mathbf{C} = \mathbf{Y}^\top \mathbf{Y} = ({\NAM}^k \XM)^\top ({\NAM}^k \XM).
\]

Using Lemma~\ref{lem:oversmooth}, when $k \to \infty$:
\[
{\NAM}^k \to \boldsymbol{\pi} \boldsymbol{\pi}^\top.
\]
Therefore,
\[
\mathbf{Y} = {\NAM}^k \XM \to \boldsymbol{\pi} (\boldsymbol{\pi}^\top\XM) = \boldsymbol{\pi}\mathbf{a}^\top,
\]
where $\mathbf{a}^\top = \boldsymbol{\pi}^\top \XM \in \mathbb{R}^{1 \times d}$.

Then,
\[
\mathbf{C} = \mathbf{Y}^\top \mathbf{Y} \to (\boldsymbol{\pi} \mathbf{a}^\top)^\top (\boldsymbol{\pi} \mathbf{a}^\top) = (\mathbf{a} \boldsymbol{\pi}^\top) (\boldsymbol{\pi} \mathbf{a}^\top).
\]
Since $\boldsymbol{\pi}^\top \boldsymbol{\pi} = 1$, we have:
\[
\mathbf{C} = \mathbf{a}\mathbf{a}^\top.
\]
As $\mathbf{a} \mathbf{a}^\top \in \mathbb{R}^{d \times d}$ is a rank-one matrix (outer product of $\mathbf{a}$ with itself), 
$\mathbf{C}$ is also rank one.
\end{proof}

\begin{proof}[\bf Proof of Theorem~\ref{thm:robustness}]

\stitle{VDD}
We decompose the difference of node $i$ after a small perturbation into two terms:
\begin{align}\label{eq:uidi-ujdi-bound}
\left\|\frac{\UM_i\LABM^t}{\sqrt{d_i}}-\frac{\widetilde{\UM}_i\widetilde{\LABM}^t}{\sqrt{{d}^\prime_i}}\right\|_2 &= \left\|\frac{\UM_i\LABM^t}{\sqrt{d_i}}-\frac{\widetilde{\UM}_i\widetilde{\LABM}^t}{\sqrt{{d}_i}}+\frac{\widetilde{\UM}_i\widetilde{\LABM}^t}{\sqrt{{d}_i}}-\frac{\widetilde{\UM}_i\widetilde{\LABM}^t}{\sqrt{{d}^\prime_i}}\right\|_2 \nonumber \\ 
&\leq \left| \frac{1}{\sqrt{d_i}} - \frac{1}{\sqrt{{d}^\prime_i}} \right| \cdot \left\| \widetilde{\UM}_i\widetilde{\LABM}^t \right\|_2 + \frac{1}{\sqrt{{d}_i}} \left\| \UM_i \LABM^t - \widetilde{\UM}_i\widetilde{\LABM}^t \right\|_2.
\end{align}
We reference the following theorem, where $\Theta(\widetilde{\textbf{V}}, \textbf{V})$ denote the $d \times d$ diagonal matrix
defined by \( \Theta(\widetilde{\textbf{V}}, \textbf{V}) = \text{diag}(\theta_1, \dots, \theta_d) \) with being the $i$-th principal angle between the subspaces spanned by the columns of \( \widetilde{\textbf{V}} \) and \( \textbf{V} \),
and let $\sin \Theta(\widetilde{\textbf{V}}, \textbf{V})$ be defined entry-wise.
\begin{theorem}[Davis–Kahan \(\sin\Theta\) Theorem (Theorem V.3.6 in~\cite{stewart1990matrix})]
Let \(\textbf{S}, \widetilde{\textbf{S}} \in \mathbb{R}^{p \times p}\) be symmetric, with eigenvalues \(\lambda_1 \geq \dots \geq \lambda_p\) and \(\widetilde{\lambda}_1 \geq \dots \geq \widetilde{\lambda}_p\) respectively. Fix indices \(1 \leq r \leq s \leq p\), let \(d = s - r + 1\), and define \(\textbf{V} = (v_r, v_{r+1}, \dots, v_s) \in \mathbb{R}^{p \times d}\) and \(\widetilde{\textbf{V}} = (\widetilde{v}_r, \widetilde{v}_{r+1}, \dots, \widehat{v}_s) \in \mathbb{R}^{p \times d}\), each having orthonormal columns with \(\textbf{S} v_i = \lambda_i v_i\) and \(\widetilde{\textbf{S}} \widetilde{v}_i = \widetilde{\lambda}_i \widetilde{v}_i\) for \(i = r, r+1, \dots, s\). If
\[
\delta = \inf\{ |\widetilde{\lambda} - \lambda| : \lambda \in [\lambda_s, \lambda_r], \widetilde{\lambda} \in (-\infty, \widetilde{\lambda}_{s-1}] \cup [\widetilde{\lambda}_{r+1}, \infty) \} > 0,
\]
where \(\widetilde{\lambda}_0 = -\infty\) and \(\widetilde{\lambda}_{p+1} = \infty\), then
\[
\| \sin\Theta(\widetilde{\textbf{V}}, \textbf{V}) \|_F \leq \frac{\|\widetilde{\textbf{S}} - \textbf{S}\|_F}{\delta}.
\]
\end{theorem}
Based on this theorem, let $\UM$ and $\widetilde{\UM}$ denote the eigenvector matrices of $\NAM$ and ${\NAM}^\prime$, respectively. Provided that \( \delta > 0 \), the inequality below holds:
\[
\| \sin \Theta(\widetilde{\UM}, \UM) \|_F \leq \frac{\|{\NAM}^\prime - \NAM\|_F}{\delta} = \frac{\epsilon_a}{\delta},
\]

Next,
\begin{align*}
\|\widetilde{\UM} - \UM \|_F^2 &= \sum_{k=1}^{n}\| \UM_{\cdot,k} - \widetilde{\UM}_{\cdot,k} \|_2^2 = \sum_{k=1}^{n} (2 - 2 \cos{\theta_k}) \\
&\leq \sum_{k=1}^{n} (2 - 2 \cos^2{\theta_k}) = \sum_{k=1}^{n} 2\sin^2{\theta_k} = 2 \|\sin{\Theta(\widetilde{\UM}, \UM)}\|_F^2,
\end{align*}
where \( \theta_1, \dots, \theta_n \) are the principal angles between the column spaces of \( \widetilde{\UM} \) and \( \UM \).
Therefore, we have:
\[
\|\widetilde{\UM} - \UM \|_F = \sqrt{2}\|\sin{\Theta(\widetilde{\UM}, \UM)}\|_F \leq \frac{\sqrt{2}\epsilon_a}{\delta}.
\]
Recall that $\widetilde{\UM}\widetilde{\LABM}\widetilde{\UM}$ is the eigendecomposition of ${\NAM}^\prime$. We have
\begin{equation}\label{eq:ulambda-bound}
\| \widetilde{\UM}_i\widetilde{\LABM}^t \|_2 = \sqrt{\widetilde{\UM}_i\widetilde{\LABM}^t\cdot (\widetilde{\UM}_i\widetilde{\LABM}^t)^{\top}} = \sqrt{\widetilde{\UM}_i\widetilde{\LABM}^{2t}\widetilde{\UM}_i^{\top}} = \sqrt{{\NAM}^{\prime 2t}_{i,i}} \le 1.
\end{equation}

The Hoffman-Wielandt inequality~\cite{hoffman1953variation} provides an upper bound on the differences between the eigenvalues of two matrices in terms of their Frobenius norm. Specifically, it states: \( \sum_{k=1}^{n} |\lambda_k - \widetilde{\lambda}_k|^2 \leq \|{\NAM}^\prime - \NAM\|^2_F \).
Given that both $\UM$ and $\widetilde{\UM}$ are orthogonal matrices
and the eigenvalues of $\NAM$ satisfy $|\lambda_k| \leq 1\ \forall{1\le k\le n}$, $|{\LABM^t_k}| \leq 1$, the eigenvalues of $|\widetilde{\lambda}_k|$ satisfy similar constraints, i.e., $|\widetilde{\lambda}_k| \leq 1$. Consequently, we can derive the following:
\[
\| (\UM_i-\widetilde{\UM}_i) \LABM^t \|_2 = \sqrt{\sum_{k=1}^n{( (\UM_{i,k}-\widetilde{\UM}_{i,k})\cdot \LABM^t_k})^2} \le \sqrt{\sum_{k=1}^n{(\UM_{i,k}-\widetilde{\UM}_{i,k})}^2}\cdot \max_{1\le k\le n}{|{\LABM^t_k}|} \leq \|\UM_i-\widetilde{\UM}_i\|_2 \leq \|\widetilde{\UM} - \UM \|_F,
\]
\[
\| \widetilde{\UM}_i (\LABM^t - \widetilde{\LABM}^t) \|_2 = \sqrt{\sum_{k=1}^n{( \UM_{i,k}\cdot (\LABM^t_k}-\widetilde{\LABM}^t_k))^2} \le \sqrt{\sum_{k=1}^n{ {\UM_{i,k}}^2}}\cdot \max_{1\le k\le n}{|\LABM^t_k-\widetilde{\LABM}^t_k|} \le \epsilon_a \cdot \|\UM_i\|_2.
\]
Then, by the triangle inequality, 
\[
\| \UM_i \LABM^t - \widetilde{\UM}_i \widetilde{\LABM}^t \|_2 \leq \| (\UM_i - \widetilde{\UM}_i) \LABM^t \|_2 + \| \widetilde{\UM}_i (\LABM^t - \widetilde{\LABM}^t) \|_2 \leq \|\widetilde{\UM} - \UM \|_F + \epsilon_a \leq (\frac{\sqrt{2}}{\delta}+1)\epsilon_a.
\]
Plugging the above inequality and Eq.~\eqref{eq:ulambda-bound} into Eq.~\eqref{eq:uidi-ujdi-bound} derives
\[
\left\|\frac{\UM_i\LABM^t}{\sqrt{d_i}}-\frac{\widetilde{\UM}_i\widetilde{\LABM}^t}{\sqrt{{d}^\prime_i}}\right\|_2 \leq \left| \frac{1}{\sqrt{d_i}} - \frac{1}{\sqrt{{d}^\prime_i}} \right| + \frac{(\frac{\sqrt{2}}{\delta}+1)\epsilon_a}{\sqrt{{d}_i}}.
\]
Let a constant \( C_1 = {\frac{\sqrt{2}}{\delta}}+1 \), the above inequality could be written as:
\[
\left\|\frac{\UM_i\LABM^t}{\sqrt{d_i}}-\frac{\widetilde{\UM}_i\widetilde{\LABM}^t}{\sqrt{{d}^\prime_i}}\right\|_2 \leq \left| \frac{1}{\sqrt{d_i}} - \frac{1}{\sqrt{{d}^\prime_i}} \right| + \frac{C_1\epsilon_a}{\sqrt{{d}_i}}.
\]

The difference in the inverse square roots of degrees is bounded as:
\[
\left| \frac{1}{\sqrt{d_i}} - \frac{1}{\sqrt{{d}^\prime_i}} \right| = \frac{|d_i - \tilde{d}_i|}{\sqrt{d_i \tilde{d}_i} \left( \sqrt{d_i} + \sqrt{\tilde{d}_i} \right)} \leq \frac{\epsilon_d}{2 \tilde{d}^{3/2}},
\]
where \( \tilde{d} = \min(d_i, {d}^\prime_i) \).

Finally, we bound the change of VDD between node $i$ and $j$ using reverse triangle inequality and triangle inequality as:
\begin{align*}    
&|\Delta_{\text{vanilla}}(v_i,v_j) - \widetilde{\Delta}_{\text{vanilla}}(v_i,v_j)| \leq  \left\|\frac{\UM_i\LABM^t}{\sqrt{d_i}}-\frac{\widetilde{\UM}_i\widetilde{\LABM}^t}{\sqrt{{d}^\prime_i}}\right\|_2+ \left\|\frac{\UM_j\LABM^t}{\sqrt{d_j}}-\frac{\widetilde{\UM}_j\widetilde{\LABM}^t}{\sqrt{{d}^\prime_j}}\right\|_2 
\leq \frac{\epsilon_d}{\tilde{d}^{3/2}} + \frac{2C_1 \epsilon_a}{\sqrt{\tilde{d}}},
\end{align*}
where \( \tilde{d} = \min(d_i, d_j, {d}^\prime_i, {d}^\prime_j) \).

\eat{
\renchi{may change to prove the distance of spectral embeddings $\|\UM\LABM^t - \widehat{\UM}\widehat{\LABM}^t\|\le ?$. If they are similar, then diffusion distance will be small.}

\renchi{
Try this way.

\begin{align*}    
\Delta_t(v_i,v_j)^2 - \widetilde{\Delta}_t(v_i,v_j)^2 & = \left\|\frac{\UM_i\LABM^t}{\sqrt{d_i}}-\frac{\UM_j\LABM^t}{\sqrt{d_j}}\right\|_2^2 - \left\|\frac{\widetilde{\UM}_i\widetilde{\LABM}^t}{\sqrt{\tilde{d}_i}}-\frac{\widetilde{\UM}_j\widetilde{\LABM}^t}{\sqrt{\tilde{d}_j}}\right\|_2^2\\
& = (\frac{1}{d_i}\NAM^{2t}_{i,i}+\frac{1}{d_j}\NAM^{2t}_{j,j}-\frac{1}{\sqrt{d_i d_j}}\NAM^{2t}_{i,j}) - (\frac{1}{\tilde{d}_i}{\NAM}^\prime^{2t}_{i,i}+\frac{1}{\tilde{d}_j}{\NAM}^\prime^{2t}_{j,j}-\frac{1}{\sqrt{\tilde{d}_i \tilde{d}_j}}{\NAM}^\prime^{2t}_{i,j})\\
& = (\frac{1}{d_i}\NAM^{2t}_{i,i}-\frac{1}{\tilde{d}_i}{\NAM}^\prime^{2t}_{i,i}) + (\frac{1}{d_j}\NAM^{2t}_{j,j}-\frac{1}{\tilde{d}_j}{\NAM}^\prime^{2t}_{j,j}) + \frac{1}{\sqrt{\tilde{d}_i \tilde{d}_j}}\cdot({\NAM}^\prime^{2t}_{i,j}-\NAM^{2t}_{i,j})
\end{align*}

Need to bound $|\NAM^{2t}_{i,j}-{\NAM}^\prime^{2t}_{i,j}|$ given $\|\NAM-{\NAM}^\prime\|_{max} \leq \|\NAM-{\NAM}^\prime\|_2 = \epsilon_a$.
}

This proof extends seamlessly from the other two types of diffusion distances. Yet, for the HKDD, it is necessary to account for the permutation matrix $\NLM$ rather than $\NAM$. \renchi{need more details.}
}

\stitle{HKDD}
Recall that for the HKDD, we perform an eigendecomposition on $\NLM$ and get $\NLM = \UM\LABM\UM^{\top}$, and that $\NLM = \IM - \NAM$, we have \( \| \NLM - \NLM^\prime \|_F \leq \epsilon_a \).
We decompose the difference of node $i$ after a small perturbation into two terms:
\begin{equation}\label{eq:uidi-ujdi-bound-HKDD}
\left\|\frac{\UM_i e^{-\gamma\LABM}}{\sqrt{d_i}}-\frac{\widetilde{\UM}_i e^{-\gamma\widetilde{\LABM}}}{\sqrt{{d}^\prime_i}}\right\|_2 \leq \left| \frac{1}{\sqrt{d_i}} - \frac{1}{\sqrt{{d}^\prime_i}} \right| \cdot \left\| \widetilde{\UM}_i e^{-\gamma\widetilde{\LABM}} \right\|_2 + \frac{1}{\sqrt{{d}_i}} \left\| \UM_i e^{-\gamma\LABM} - \widetilde{\UM}_i e^{-\gamma\widetilde{\LABM}} \right\|_2.
\end{equation}

\eat{
need to bound \( \| e^{-\gamma\LABM} - e^{-\gamma\widetilde{\LABM}} \|_F = \sum_{k=1}^{n} |e^{-\gamma\lambda_k} - e^{-\gamma\widetilde{\lambda}_k}|^2 \leq \sum_{k=1}^{n} |e^{-\lambda_k} - e^{-\widetilde{\lambda}_k}|^2 \leq \sum_{k=1}^{n} |\lambda_k - \widetilde{\lambda}_k|^2 \)

need to bound \( \| e^{-\gamma\LABM} - e^{-\gamma\widetilde{\LABM}} \|_F = \sum_{k=1}^{n} |e^{-\gamma\lambda_k} - e^{-\gamma\widetilde{\lambda}_k}|^2 \leq \sum_{k=1}^{n} |\lambda_k - \widetilde{\lambda}_k|^2 \)
}
The crucial part is to bound \( \| e^{-\gamma\LABM} - e^{-\gamma\widetilde{\LABM}} \|_F \).
The Frobenius norm is given by:
\[
\| e^{-\gamma\LABM} - e^{-\gamma\widetilde{\LABM}} \|_F = \sqrt{\sum_{i=1}^n \left( e^{-\gamma\lambda_i} - e^{-\gamma\tilde{\lambda}_i} \right)^2}.
\]
By the mean value theorem, for each \( k \), there exists \( \xi_k \) between \( \lambda_k \) and \( \tilde{\lambda}_k \) such that:
\[
e^{-\gamma\lambda_k} - e^{-\gamma\tilde{\lambda}_k} = -\gamma e^{-\gamma\xi_k} (\lambda_k - \tilde{\lambda}_k).
\]
Taking the absolute value:
\[
|e^{-\gamma\lambda_k} - e^{-\gamma\tilde{\lambda}_k}| = \gamma e^{-\gamma\xi_k} |\lambda_k - \tilde{\lambda}_k|.
\]
Substituting into the Frobenius norm:
\[
\| e^{-\gamma\LABM} - e^{-\gamma\widetilde{\LABM}} \|_F = \sqrt{\sum_{i=1}^n \left( e^{-\gamma\lambda_i} - e^{-\gamma\tilde{\lambda}_i} \right)^2} = \sqrt{\sum_{i=1}^n \left( \gamma e^{-\gamma\xi_i} |\lambda_i - \tilde{\lambda}_i| \right)^2}.
\]
Assume that \( \xi_i \geq \lambda_{\min} \), where \( \lambda_{\min} \) is a lower bound for the eigenvalues of both \( \LABM \) and \( \widetilde{\LABM} \). Note that as the eigenvalues of \( \LABM \) and \( \widetilde{\LABM} \) is above 0 (after excluding the trivial eigenvalue \( 0 \)). Then:
\[
e^{-\gamma\xi_i} \leq e^{-\gamma\lambda_{\min}},
\]
and:
\[
\| e^{-\gamma\LABM} - e^{-\gamma\widetilde{\LABM}} \|_F \leq \gamma e^{-\gamma\lambda_{\min}} \sqrt{\sum_{i=1}^n (\lambda_i - \tilde{\lambda}_i)^2} = \gamma e^{-\gamma\lambda_{\min}} \| \LABM - \widetilde{\LABM} \|_F.
\]

We now analyze the behavior of the function \( f(\gamma) = \gamma e^{-\gamma \lambda_{\min}} \). The derivative of \( f(\gamma) \) with respect to \( \gamma \) is given by:
\[
f'(\gamma) = e^{-\gamma \lambda_{\min}} - \gamma \lambda_{\min} e^{-\gamma \lambda_{\min}} = e^{-\gamma \lambda_{\min}} (1 - \gamma \lambda_{\min}).
\]
Setting \( f'(\gamma) = 0 \), we find that:
\[
1 - \gamma \lambda_{\min} = 0 \implies \gamma = \frac{1}{\lambda_{\min}}.
\]
At \( \gamma = \frac{1}{\lambda_{\min}} \), the function \( f(\gamma) \) attains its maximum value:
\[
f\left( \frac{1}{\lambda_{\min}} \right) = \frac{1}{\lambda_{\min}} e^{-1}.
\]
For all \( \gamma > 0 \), we observe that:
\[
f(\gamma) = \gamma e^{-\gamma \lambda_{\min}} \leq \frac{1}{e\lambda_{\min}}.
\]
Therefore, we have:
\[
\| e^{-\gamma\LABM} - e^{-\gamma\widetilde{\LABM}} \|_F \leq \gamma e^{-\gamma\lambda_{\min}} \sqrt{\sum_{i=1}^n (\lambda_i - \tilde{\lambda}_i)^2} \leq \frac{1}{e\lambda_{\min}} \sqrt{\sum_{i=1}^n (\lambda_i - \tilde{\lambda}_i)^2}.
\]
Utilizing the Hoffman-Wielandt inequality, we have:
\[
\| e^{-\gamma\LABM} - e^{-\gamma\widetilde{\LABM}} \|_F \leq \frac{1}{e\lambda_{\min}} \sqrt{\sum_{i=1}^n (\lambda_i - \tilde{\lambda}_i)^2} \leq \frac{1}{e\lambda_{\min}}\|{\NAM}^\prime - \NAM\|_F.
\]
Plugging the above inequalities into Eq.~\eqref{eq:uidi-ujdi-bound-HKDD} derives
\[
\left\|\frac{\UM_i e^{-\gamma\LABM}}{\sqrt{d_i}}-\frac{\widetilde{\UM}_i e^{-\gamma\widetilde{\LABM}}}{\sqrt{{d}^\prime_i}}\right\|_2 \leq \left| \frac{1}{\sqrt{d_i}} - \frac{1}{\sqrt{{d}^\prime_i}} \right| + \frac{(\frac{\sqrt{2}}{\delta}+\frac{1}{e\lambda_{\min}})\epsilon_a}{\sqrt{{d}_i}}.
\]
Let a constant \( C_2 = \frac{\sqrt{2}}{\delta}+\frac{1}{e\lambda_{\min}} \), the above inequality could be written as:
\[
\left\|\frac{\UM_i e^{-\gamma\LABM}}{\sqrt{d_i}}-\frac{\widetilde{\UM}_i e^{-\gamma\widetilde{\LABM}}}{\sqrt{{d}^\prime_i}}\right\|_2 \leq \left| \frac{1}{\sqrt{d_i}} - \frac{1}{\sqrt{{d}^\prime_i}} \right| + \frac{C_2\epsilon_a}{\sqrt{{d}_i}}.
\]
Finally, we bound the change of HKDD between node $i$ and $j$ as:
\begin{align*}   
|\Delta_{\text{HK}}(v_i,v_j) - \widetilde{\Delta}_{\text{HK}}(v_i,v_j)|
&\leq  \left\|\frac{\UM_i e^{-\gamma\LABM}}{\sqrt{d_i}}-\frac{\widetilde{\UM}_i e^{-\gamma\widetilde{\LABM}}}{\sqrt{{d}^\prime_j}}\right\|_2 + \left\|\frac{\UM_j e^{-\gamma\LABM}}{\sqrt{d_j}}-\frac{\widetilde{\UM}_j e^{-\gamma\widetilde{\LABM}}}{\sqrt{{d}^\prime_j}}\right\|_2 \\
&\leq \frac{\epsilon_d}{\tilde{d}^{3/2}} + \frac{2C_2 \epsilon_a}{\sqrt{\tilde{d}}}.
\end{align*}

\stitle{PRDD}
We decompose the difference of node $i$ after a small perturbation into two terms:
\begin{equation}\label{eq:uidi-ujdi-bound-PRDD}
\left\|\frac{\UM_i}{\sqrt{d_i}(1-\gamma\LABM)}-\frac{\widetilde{\UM}_i}{\sqrt{{d}^\prime_i}(1-\gamma\widetilde{\LABM})}\right\|_2 \leq \left| \frac{1}{\sqrt{d_i}} - \frac{1}{\sqrt{{d}^\prime_i}} \right| \cdot \left\| \frac{\widetilde{\UM}_i}{1-\gamma\widetilde{\LABM}} \right\|_2 + \frac{1}{\sqrt{{d}_i}} \left\| \frac{\UM_i}{1-\gamma\LABM} - \frac{\widetilde{\UM}_i}{1-\gamma\widetilde{\LABM}} \right\|_2.
\end{equation}
For the first component, the denominator \( 1 - \gamma \widetilde{\lambda}_k \) satisfies: \( |1 - \gamma \widetilde{\lambda}_k| \geq 1 - \gamma|\widetilde{\lambda}_k| \geq 1 - \gamma \) as \( 0< \gamma < 1 \).
Therefore, we have:
\[
\left\| \frac{\widetilde{\UM}_i}{1-\gamma\widetilde{\LABM}} \right\|_2 \leq \left\| \widetilde{\UM}_i \right\|_2 \cdot \left\| \frac{1}{1-\gamma\widetilde{\LABM}} \right\|_2 \leq 1 \cdot \frac{1}{1-\gamma}
= \frac{1}{1-\gamma}.
\]
Next, we bound \( \left\| \frac{1}{1-\gamma\LABM} - \frac{1}{1-\gamma\widetilde{\LABM}} \right\|_2 \).
For each diagonal element, we simplify the difference:
\[
\frac{1}{1 - \gamma \lambda_k} - \frac{1}{1 - \gamma \widetilde{\lambda}_k} = \frac{\gamma (\lambda_k - \widetilde{\lambda}_k)}{(1 - \gamma \lambda_k)(1 - \gamma \widetilde{\lambda}_k)}.
\]
Taking the absolute value:
\[
\left| \frac{1}{1 - \gamma \lambda_k} - \frac{1}{1 - \gamma \widetilde{\lambda}_k} \right| = \left| \frac{\gamma (\lambda_k - \widetilde{\lambda}_k)}{(1 - \gamma \lambda_k)(1 - \gamma \widetilde{\lambda}_k)} \right|.
\]
We have \( |1 - \gamma \lambda_k| \geq 1 - \gamma \) and \( |1 - \gamma \widetilde{\lambda}_k| \geq 1 - \gamma \).
Using the above, the difference can be bounded as:
\[
\left| \frac{1}{1 - \gamma \lambda_k} - \frac{1}{1 - \gamma \widetilde{\lambda}_k} \right| \leq \frac{\gamma |\lambda_k - \widetilde{\lambda}_k|}{(1 - \gamma)^2}.
\]
Therefore, we have:
\[
\left\| \frac{1}{1 - \gamma \LABM} - \frac{1}{1 - \gamma \widetilde{\LABM}} \right\|_2 \leq \frac{\gamma \left\| \LABM - \widetilde{\LABM} \right\|_2}{(1 - \gamma)^2}.
\]
For the second component, using the triangle inequality:
\begin{align*}
\left\| \frac{\UM_i}{1-\gamma\LABM} - \frac{\widetilde{\UM}_i}{1-\gamma\widetilde{\LABM}} \right\|_2 
&\leq \left\| \frac{\UM_i - \widetilde{\UM}_i}{1-\gamma\LABM} \right\|_2 + \left\| \widetilde{\UM}_i (\frac{1}{1-\gamma\LABM} - \frac{1}{1-\gamma\widetilde{\LABM}}) \right\|_2 \\
&\leq \frac{\| \UM_i - \widetilde{\UM}_i \|_2}{1-\gamma} + \frac{\gamma \left\| \LABM - \widetilde{\LABM} \right\|_2}{(1 - \gamma)^2}\\
&\leq \left(\frac{\sqrt{2}}{(1 - \gamma)\delta}+\frac{\gamma}{(1 - \gamma)^2}\right)\epsilon_a=\frac{1}{1-\gamma} \cdot \left(\frac{\sqrt{2}}{\delta}+\frac{\gamma}{(1 - \gamma)}\right)\epsilon_a.
\end{align*}
Plugging the above inequalities into Eq.~\eqref{eq:uidi-ujdi-bound-PRDD} derives
\[
\left\|\frac{\UM_i}{\sqrt{d_i}(1-\gamma\LABM)}-\frac{\widetilde{\UM}_i}{\sqrt{{d}^\prime_i}(1-\gamma\widetilde{\LABM})}\right\|_2
\leq \left| \frac{1}{\sqrt{d_i}} - \frac{1}{\sqrt{{d}^\prime_i}} \right| \cdot \frac{1}{1-\gamma} + \frac{1}{\sqrt{{d}_i}} \cdot \frac{1}{1-\gamma} \cdot \left(\frac{\sqrt{2}}{\delta}+\frac{\gamma}{(1 - \gamma)}\right)\epsilon_a.
\]
Let a constant \( C_3 = {\frac{\sqrt{2}}{\delta}}+1 \), the above inequality could be written as:
\[
\left\|\frac{\UM_i}{\sqrt{d_i}(1-\gamma\LABM)}-\frac{\widetilde{\UM}_i}{\sqrt{{d}^\prime_i}(1-\gamma\widetilde{\LABM})}\right\|_2
\leq \left| \frac{1}{\sqrt{d_i}} - \frac{1}{\sqrt{{d}^\prime_i}} \right| \cdot \frac{1}{1-\gamma} + \frac{C_3\epsilon_a}{\sqrt{{d}_i}} \cdot \frac{1}{1-\gamma}.
\]
Finally, we bound the PRDD between node $i$ and $j$ as:
\begin{align*}
|\Delta_{\text{PR}}(v_i,v_j) - \widetilde{\Delta}_{\text{PR}}(v_i,v_j)|
&\leq  \left\|\frac{\UM_i}{\sqrt{d_i}(1-\gamma\LABM)}-\frac{\widetilde{\UM}_i}{\sqrt{{d}^\prime_i}(1-\gamma\widetilde{\LABM})}\right\|_2 + \left\|\frac{\UM_j}{\sqrt{d_j}(1-\gamma\LABM)}-\frac{\widetilde{\UM}_j}{\sqrt{{d}^\prime_j}(1-\gamma\widetilde{\LABM})}\right\|_2 \\
&\leq \frac{\epsilon_d}{\tilde{d}^{3/2}(1-\gamma)} + \frac{2C_3 \epsilon_a}{\sqrt{\tilde{d}}(1-\gamma)}.
\end{align*}
\end{proof}

\begin{proof}[\bf Proof of Theorem~\ref{thm:dd-computation}]
We will prove three parts of the theorem separately.

\stitle{VDD}
We perform an eigendecomposition on $\NAM$ and as $\NAM$ is symmetric, we have $\NAM = \UM\LABM\UM^{\top}$.
Then, we have $\NAM^t = (\UM\LABM\UM^{\top})^t = \UM\LABM^t\UM^{\top}$.
Therefore, recall Eq. (\ref{eq:DD}), we have $\Delta_{\text{vanilla}}(v_i, v_j) = \left\|\frac{(\UM\LABM^t\UM^{\top})_i}{\sqrt{d_i}}-\frac{(\UM\LABM^t\UM^{\top})_j}{\sqrt{d_j}}\right\|_2 = \left\|\frac{\UM_i\LABM^t\UM^{\top}}{\sqrt{d_i}}-\frac{\UM_j\LABM^t\UM^{\top}}{\sqrt{d_j}}\right\|_2$.
Expanding this expression, we obtain:
\[
\frac{\UM_i\LABM^t\UM^{\top} \cdot (\UM_i\LABM^t\UM^{\top})^{\top}}{d_i} + \frac{\UM_j\LABM^t\UM^{\top} \cdot (\UM_j\LABM^t\UM^{\top})^{\top}}{d_j} 
- \frac{2\UM_i\LABM^t\UM^{\top} \cdot (\UM_j\LABM^t\UM^{\top})^{\top}}{\sqrt{d_id_j}}.
\]
Since $\UM$ is an orthogonal matrix and $\LABM$ is a diagonal matrix, we can simplify the expression further:
\[
\frac{\UM_i\LABM^{2t}\UM_i^{\top}}{d_i} + \frac{\UM_j\LABM^{2t}\UM_j^{\top}}{d_j} - \frac{2\UM_i\LABM^{2t}\UM_j^{\top}}{\sqrt{d_id_j}}.
\]

The expression $\left\|\frac{\UM_i\LABM^t}{\sqrt{d_i}}-\frac{\UM_j\LABM^t}{\sqrt{d_j}}\right\|_2$ can be expanded and simplified to yield an equivalent result. 

\stitle{HKDD}
We perform an eigendecomposition on $\NLM$ and get $\NLM = \UM\LABM\UM^{\top}$, and therefore, $\NLM = \DM^{-1/2}(\UM\LABM\UM^{\top})\DM^{1/2}$.
To get $e^{-\gamma\NLM}$, using the diagonalization of the symmetrically normalized Laplacian $\NLM$, we firstly have $e^{-\NLM} = e^{-\UM\LABM\UM^{\top}} = \UM e^{-\LABM}\UM^{\top}$. After that, we have
\[
e^{-\gamma\NLM} = (\DM^{-1/2}\UM)e^{-\gamma\LABM}(\UM^{\top}\DM^{1/2}).
\]
Similar to the above proof, $(\DM^{-1/2}\UM)e^{-\gamma\LABM}(\UM^{\top}\DM^{1/2})$ can be simplified to $\DM^{-1/2}\UM e^{-\gamma\LABM}$, therefore, this leads to the form of \( \left\|\frac{\UM_i e^{-\gamma\LABM}}{\sqrt{d_i}}-\frac{\UM_j e^{-\gamma\LABM}}{\sqrt{d_j}}\right\|_2 \).

\stitle{PRDD}
We have $\PM^t = \DM^{-1/2}\NAM^t\DM^{1/2}$, and after an eigendecomposition on $\NAM$, 
we have $\Delta_{\text{PR}}(v_i,v_j) = \left\|(\sum_{t=0}^{\infty}\gamma^t\DM^{-1/2}\UM\LABM^t)_i-(\sum_{t=0}^{\infty}\gamma^t\DM^{-1/2}\UM\LABM^t)_j \right\|_2$.
Applying the properties of the Neumann series, we can have 
\[
\Delta_{\text{PR}}(v_i,v_j) = \left\|\left(\frac{\DM^{-1/2}\UM}{1-\gamma\LABM}\right)_i-\left(\frac{\DM^{-1/2}\UM}{1-\gamma\LABM}\right)_j \right\|_2 
= \left\|\frac{\UM_i}{\sqrt{d_i}(1-\gamma\LABM)}-\frac{\UM_j}{\sqrt{d_j}(1-\gamma\LABM)} \right\|_2.
\]
This finishes the proof.
\end{proof}

\begin{proof}[\bf Proof of Theorem~\ref{prop:dd-unified}]
For the VDD and PRDD, we perform an eigendecompsition on $\NAM$, while for the HKDD, we perform an eigendecompsition on $\NLM$, yielding the decomposition $\UM\LABM\UM^{\top}$.
Suppose $\Phi = \DM^{-1/2} \UM$, where $\UM$ is an orthonormal matrix, i.e., $\UM\UM^{\top} = \UM^{\top}\UM = \IM$, it follows that
\[\Phi\Phi^{\top} = \DM^{-1/2}\UM\UM^{\top}\DM^{-1/2} = \DM^{-1}.\]
Therefore,
\[\sum_{\ell=1}^n\phi_\ell(i)\phi_\ell(j) = (\Phi\Phi^{\top})_{ij} = \frac{\mathbb{1}_{i=j}}{d_i}.\]
Thus, we have:
\[\sum_{\ell=1}^n (\phi_\ell(i) - \phi_\ell(j))^2 = \frac{1}{d_i} + \frac{1}{d_j} - 2 \cdot \frac{\mathbb{1}_{i=j}}{d_i}.\]
Clearly,
\[\sum_{\ell=1}^n (\phi_\ell(i) - \phi_\ell(j))^2 \leq \frac{2}{\min(d_i,d_j)} (1 - \mathbb{1}_{i=j}),\]
for all $i, j = 1, 2, \ldots, n$.
Accordingly, for the three categories of diffusion distances, the results are as follows:

1) for the VDD, \[\Delta^{\prime 2}(v_i, v_j) = \Delta^2(v_i, v_j) - \sum_{\ell=\kappa+1}^n \lambda_\ell^{2t} (\phi_\ell(i) - \phi_\ell(j))^2;\]

2) for the HKDD, \[\Delta^{\prime 2}(v_i, v_j) = \Delta^2(v_i, v_j) - \sum_{\ell=\kappa+1}^n e^{-2\gamma\lambda_\ell} (\phi_\ell(i) - \phi_\ell(j))^2;\]

3) for the PRDD, \[\Delta^{\prime 2}(v_i, v_j) = \Delta^2(v_i, v_j) - \sum_{\ell=\kappa+1}^n \frac{(\phi_\ell(i) - \phi_\ell(j))^2}{(1-\gamma\lambda_\ell)^2}.\]

Taking the approximated VDD as an example, and applying the derived bound:
\begin{align*}
\sum_{\ell=\kappa+1}^n \lambda_\ell^{2t} (\phi_\ell(i) - \phi_\ell(j))^2 &\leq \lambda_\kappa^{2t} \sum_{\ell=\kappa+1}^n (\phi_\ell(i) - \phi_\ell(j))^2\\
&\leq \lambda_\kappa^{2t} \sum_{\ell=1}^n (\phi_\ell(i) - \phi_\ell(j))^2\\
&\leq \frac{2\lambda_\kappa^{2t}}{\min(d_i,d_j)} (1 - \mathbb{1}_{i=j}),
\end{align*}
we obtain:
\[\Delta^{\prime 2}(v_i, v_j) \geq \Delta^2(v_i, v_j) - \frac{2\lambda_\kappa^{2t}}{\min(d_i,d_j)} (1 - \mathbb{1}_{i=j}).\]
Analogous bounds can be proven for the other two approximated distances.
On the other hand, it is apparent that:
\[\Delta^{\prime 2}(v_i, v_j)\leq \Delta^2(v_i, v_j).\]
This ends the proof.
\end{proof}

\end{document}